\documentclass[a4paper,preprint,sort&compress]{cas-sc}

\usepackage[authoryear,longnamesfirst]{natbib}
\def\bng{\bngx}

\font\bngx=bang10

\def\*#1*#2{o\null{#2}{#1}}

\def\sh#1{\setbox0=\hbox{#1}%
     \kern-.02em\copy0\kern-\wd0
     \kern.04em\copy0\kern-\wd0
     \kern-.02em\raise.0433em\box0 }
\usepackage[T1]{fontenc}
\usepackage{amsmath,amssymb,amsfonts}
\usepackage{graphicx}
\usepackage{caption}
\usepackage{subcaption}
\usepackage{enumitem}
\usepackage{algorithm}
\usepackage{url}
\usepackage{xurl}
\usepackage[autostyle]{csquotes}
\PassOptionsToPackage{noend}{algpseudocode}
\usepackage{algpseudocode}
\usepackage{multirow}
\usepackage{booktabs}
\usepackage{float}
\usepackage{tablefootnote}

\errorcontextlines\maxdimen
\makeatletter
\newcommand*{\algrule}[1][\algorithmicindent]{\makebox[#1][l]{\hspace*{.5em}\vrule height .75\baselineskip depth .25\baselineskip}}%

\newcount\ALG@printindent@tempcnta
\def\ALG@printindent{%
    \ifnum \theALG@nested>0% is there anything to print
        \ifx\ALG@text\ALG@x@notext% is this an end group without any text?
            \addvspace{-3pt}% FUDGE for cases where no text is shown, to make the rules line up
        \else
            \unskip
            \ALG@printindent@tempcnta=1
            \loop
                \algrule[\csname ALG@ind@\the\ALG@printindent@tempcnta\endcsname]%
                \advance \ALG@printindent@tempcnta 1
            \ifnum \ALG@printindent@tempcnta<\numexpr\theALG@nested+1\relax% can't do <=, so add one to RHS and use < instead
            \repeat
        \fi
    \fi
    }%
\usepackage{etoolbox}
\patchcmd{\ALG@doentity}{\noindent\hskip\ALG@tlm}{\ALG@printindent}{}{\errmessage{failed to patch}}
\makeatother
\def\tsc#1{\csdef{#1}{\textsc{\lowercase{#1}}\xspace}}
\tsc{WGM}
\tsc{QE}
\tsc{EP}
\tsc{PMS}
\tsc{BEC}
\tsc{DE}
\usepackage{lipsum}

\begin{document}
\let\WriteBookmarks\relax
\def\floatpagepagefraction{1}
\def\textpagefraction{.001}

\shorttitle{Bengali Fake Reviews: A Benchmark Dataset and Detection System}

\shortauthors{Shahariar et~al.}

\title [mode = title]{Bengali Fake Reviews: A Benchmark Dataset and Detection System}                      
% \tnotemark[1,2]

% % Title footnote 1.
% % eg: \tnotetext[1]{Title footnote text}
% % \tnotetext[<tnote number>]{<tnote text>} 
% \tnotetext[1]{This document is the results of the research
%    project funded by the National Science Foundation.}

% \tnotetext[2]{The second title footnote which is a longer text matter
%    to fill through the whole text width and overflow into
%    another line in the footnotes area of the first page.}
\author[]{G M Shahariar}

\cormark[1]

\fnmark[1]

\ead{sshibli745@gmail.com}

% % URL of the first author
% \ead[url]{www.cvr.cc, cvr@sayahna.org}

% %  Credit authorship
% \credit{Conceptualization of this study, Methodology, Software}

\affiliation[]{organization={Ahsanullah University of Science and Technology},
    % addressline={Radarweg 29}, 
    city={Dhaka},
    % citysep={}, % Uncomment if no comma needed between city and postcode
    % postcode={1043 NX}, 
    country={Bangladesh}}

\author[]{Md. Tanvir Rouf Shawon}
% style=chinese

\fnmark[1]
\ead{shawontanvir.cse@aust.edu}

\author[]{Faisal Muhammad Shah}
   % role=Co-ordinator,
   % suffix=Jr,

\ead{faisal.cse@aust.edu}

% \ead[URL]{www.sayahna.org}

% \credit{Data curation, Writing - Original draft preparation}

% \affiliation[2]{organization={Sayahna Foundation},
%     % addressline={}, 
%     city={Jagathy},
%     % citysep={}, % Uncomment if no comma needed between city and postcode
%     postcode={695014}, 
%     state={Trivandrum},
%     country={India}}

\author[] {Mohammad Shafiul Alam}
% \cormark[2]
% \fnmark[1,3]
\ead{shafiul.cse@aust.edu}
% \ead[URL]{www.stmdocs.in}

% \affiliation[3]{organization={STM Document Engineering Pvt Ltd.},
%     addressline={Mepukada}, 
%     city={Malayinkil},
%     % citysep={}, % Uncomment if no comma needed between city and postcode
%     postcode={695571}, 
%     state={Trivandrum},
%     country={India}}

% Fifth author
\author[] {Md. Shahriar Mahbub}
% \cormark[2]
% \fnmark[1,3]
\ead{shahriar.cse@aust.edu}
% \ead[URL]{www.stmdocs.in}

% \affiliation[3]{organization={STM Document Engineering Pvt Ltd.},
%     addressline={Mepukada}, 
%     city={Malayinkil},
%     % citysep={}, % Uncomment if no comma needed between city and postcode
%     postcode={695571}, 
%     state={Trivandrum},
%     country={India}}

\cortext[cor1]{Corresponding author}
%\cortext[cor2]{Principal corresponding author}

\fntext[fn1]{These authors contributed equally to this work.}
% \fntext[fn2]{Another author footnote, this is a very long footnote and
%   it should be a really long footnote. But this footnote is not yet
%   sufficiently long enough to make two lines of footnote text.}

% % For a title note without a number/mark
% \nonumnote{This note has no numbers. In this work we demonstrate $a_b$
%   the formation Y\_1 of a new type of polariton on the interface
%   between a cuprous oxide slab and a polystyrene micro-sphere placed

\begin{abstract}
The proliferation of fake reviews on various online platforms has created a major concern for both consumers and businesses. Such reviews can deceive customers and cause damage to the reputation of products or services, making it crucial to identify them. Although the detection of fake reviews has been extensively studied in English language, detecting fake reviews in non-English languages such as Bengali is still a relatively unexplored research area. The novelty of the study unfolds on three fronts: i) a new publicly available dataset called Bengali Fake Review Detection (BFRD) dataset is introduced, ii) a unique pipeline has been proposed that translates English words to their corresponding Bengali meaning and also back transliterates Romanized Bengali to Bengali, iii) a weighted ensemble model that combines four pre-trained transformers model is proposed. The developed dataset consists of 7710 non-fake and 1339 fake food-related reviews collected from social media posts. Rigorous experiments have been conducted to compare multiple deep learning and pre-trained transformer language models and our proposed model to identify the best-performing model. According to the experimental results, the proposed ensemble model attained a weighted F1-score of 0.9843 on a dataset of 13,390 reviews, comprising 1,339 actual fake reviews, 5,356 augmented fake reviews, and 6,695 reviews randomly selected from the 7,710 non-fake instances.

\end{abstract}

% \includegraphics{figs/grabs.pdf}

% \begin{highlights}
% \item Research highlights item 1
% \item Research highlights item 2
% \item Research highlights item 3
% \end{highlights}

% Keywords

\maketitle
 
\section{Introduction \label{sec1}}
Online reviews are written comments or assessments posted on websites, social media, or other digital platforms by individuals to convey their views about a product, service, or experience \citep{duan2008online,luca2015user}. Consumers depend extensively on reviews when making purchasing decisions, arranging vacation, picking restaurants, and selecting service providers \citep{ha2015impact}. Online reviews can also have a significant impact on businesses \citep{businessinsiderFakeReviews,luca2016reviews}. Positive reviews can boost a company's reputation, increase their visibility, and lead to increased sales. Negative reviews, on the other hand, can damage a company's reputation and result in lost business. For these reasons, many companies actively monitor and respond to reviews in an effort to manage their online reputation. However, with the rise of online reviews, there has also been an increase in fake reviews.

Reviews that do not come from actual consumers and are written with the intention of influencing how a product or service is perceived online are generally known as fake/spam reviews \citep{jindal2007review}. Fake reviews can be created by employees or associates of a business to artificially boost the ratings or reputation of their services or products, by rival businesses or dissatisfied clients to damage the reputation of a company's products or services, or by people who have never used the goods or services but have been paid to write favorable reviews \citep{mukherjee2012spotting}. Consumers might be misled by fraudulent reviews, which lowers the credibility of all online reviews which is why fake review detection is important. While some nations have passed legislation to control online reviews and prevent the spread of false reviews, many platforms have implemented procedures to identify and eliminate fake reviews in an effort to solve this issue \citep{mukherjee2013yelp,luca2016fake}. 

Fake review detection is a complex task due to several reasons including the sophisticated techniques used by those who create fake reviews, the large volume of reviews posted on popular platforms, the diverse types of reviews, and the challenge of distinguishing genuine reviews that may sound similar to one another \citep{salminen2022creating}. There are also linguistic challenges \citep{rao2021review} such as language dependence, slang and colloquial language, grammatical errors, and the need for accurate sentiment and contextual analysis. Machine learning is a reliable approach to identify fake reviews as it can examine various factors including the length of the review, its sentiment, and the structure of sentences, patterns and keywords. Over time, many research studies have been conducted on detecting fake reviews. These include traditional machine learning methods \citep{jindal2008opinion,ott-etal-2011-finding,mukherjee2013fake,banerjee2015using},
% ,etaiwi2017impact,shan2021conflicts,mohawesh2021analysis,sedighi2017rlosd,khurshid2017recital,hernandez2017cross,yao2021ensemble,khurshid2019enactment,mani2018spam,hammad2013approach}, 
deep learning techniques \citep{li2015learning,zhao2018towards,wang2018detecting},
% ren2016deceptive,wang2018lstm,liu2020incorporating,zeng2019review}, 
hybrid and transformer models \citep{zhang2018dri,dhamani2019using,shiblispam,gupta2021leveraging}
% mir2023online}. 
% Although previous machine learning models have demonstrated success, there are still certain challenges and limitations to address. Some of these challenges include computational complexity, the need for large labeled data sets, and difficulty detecting more advanced forms of fake reviews.

Bengali is spoken by approximately 272.7 million individuals and ranks seventh on the list of most spoken languages globally \citep{wikipediaListLanguages}, but it is still viewed as a language with limited resources. More than forty five million people use Bengali for textual communication every day on social media platforms like Facebook and YouTube \citep{sharif2022tackling}. There might be various reasons why individuals utilize social media (i.e. Facebook) posts or groups to write both authentic and false reviews in Bengali. One explanation is that Facebook is a prominent site with a huge user base in Bangladesh, thus it gives a quick and accessible option to exchange reviews with others. Also, people may feel more at ease writing and sharing evaluations in Bengali, which may not be supported by other review services. 

However, the development of fake review detection methods in languages other than English, such as Bengali, is still limited due to the challenges posed by the language's low resource status. There is a shortage of data sets and linguistic resources for Bengali, making it difficult to create effective language processing tools and models. Furthermore, the lack of standardization in the language and variation in language use across different dialects make the development of such tools more complicated. 

In this study, we introduce the Bengali Fake Review Detection (BFRD) dataset that focuses on food-related reviews in Bengali language. To create this dataset, we collected food-related reviews written in Bengali from social media groups and posts and carefully annotated them as fake or non-fake using expert annotators with some predefined criteria. To the best of our knowledge, we are presenting the first publicly accessible dataset for identifying fake reviews in Bengali comprising 7710 non-fake and 1339 fake reviews. To convert non-Bengali words (English, Romanized Bengali) to Bengali in a review, we have created a unique text conversion pipeline consisting of translation and back transliteration. In addition to creating the dataset, we explored several deep learning and pre-trained transformer models including CNN \citep{lecun1998gradient}, BiLSTM \citep{graves2005framewise}, CNN-BiLSTM \citep{rhanoui2019cnn}, CNN-BiLSTM with attention mechanism \citep{lu2021cnn} as well as five available pre-trained BERT \citep{devlin-etal-2019-bert}, ELECTRA \citep{clark2020electra} and ALBERT \citep{lan2019albert} variant language models. After conducting extensive experimentation, we propose a weighted ensemble model that combines four pre-trained transformers i.e. \textit{BanglaBERT Base} \citep{Sagor_2020}, \textit{BanglaBERT} \citep{bhattacharjee-etal-2022-banglabert}, \textit{BanglaBERT Large} \citep{bhattacharjee-etal-2022-banglabert}, and \textit{BanglaBERT Generator} \citep{bhattacharjee-etal-2022-banglabert} to detect fake reviews. To overcome the problem of class imbalance, we utilized text augmentation techniques to increase the number of fake reviews. Specifically, we employed two available augmentation libraries: \textit{nlpaug} \citep{ma2019nlpaug} and \textit{bnaug}\footnote{\url{https://github.com/sagorbrur/bnaug}}. Using the proposed weighted ensemble model, we achieved a 0.9843 weighted F1-score on 13390 reviews, of which 6695 were fake (1339 were actual fake and the rest were augmented using \textit{nlpaug}) and 6695 were non-fake (randomly selected from 7710 instances). Similarly, using the same ensemble model, we achieved a 0.9558 weighted F1-score when the fake reviews were augmented using \textit{bnaug}. In summary, we have made the following contributions in this paper:
\begin{itemize}
\item We have developed a Bengali Fake Review Detection (BFRD)\footnote{\url{https://github.com/shahariar-shibli/Bengali-Fake-Reviews-A-Benchmark-Dataset-and-Detection-System}} dataset that contains $9,049$ food-related reviews among which $1339$ are fake and $7710$ are non-fake. The reviews are collected from social media and annotated with the help of expert annotators.

\item We have developed a unique text conversion pipeline that translate English words to in a text to their corresponding Bengali meaning and back transliterates Romanized Bengali to Bengali.

\item We have utilized text augmentation techniques such as token replacement (using random masking, GloVe, Word2Vec), back translation, paraphrasing to handle class imbalance problem by increasing the fake review instances. 

\item We have proposed a weighted ensemble model consisting of four pre-trained Bengali language models: \textit{BanglaBERT Base} \citep{Sagor_2020}, \textit{BanglaBERT} \citep{bhattacharjee-etal-2022-banglabert}, \textit{BanglaBERT Large} \citep{bhattacharjee-etal-2022-banglabert}, and \textit{BanglaBERT Generator} \citep{bhattacharjee-etal-2022-banglabert} that outperforms average ensemble approach and other deep learning and transformer models.

\item We have conducted extensive experimentation and presented both quantitative and qualitative analysis of the results. Moreover, we have employed the Local Interpretable Model-Agnostic Explanations (LIME) \citep{ribeiro-etal-2016-trust} text explainer framework to provide explanations for the model's predictions. We have also analyzed misclassification categories and compared the proposed ensemble model with other existing ensemble techniques. 
\end{itemize}

The rest of the paper is structured as follows: in section \ref{sec2}, some of the related previous works are discussed. Section \ref{sec3} defines the problem, while section \ref{sec4} provides an overview of the process of creating the corpus. The proposed methodology is outlined in section \ref{sec5}, followed by an explanation and analysis of the experimental results in section \ref{sec6}. Lastly, in sections \ref{sec7} and \ref{sec8}, limitations and a concluding remark are provided respectively. 

\section{Related Works \label{sec2}}
In this section, we discuss some works related to our research. During our exploration, we found no previous work on Bengali fake review detection. There are a number of works on detecting fake reviews in English and some other languages, which we have studied and classified into three main categories: traditional machine learning approaches, deep learning approaches, and hybrid along with transformer approaches. We discuss each of these categories below.
\subsection{Traditional Approaches}
Fake review detection using traditional machine learning involves a range of methods such as Support Vector Machines (SVM), Logistic Regression (LR), Naïve Bayes (NB), Random Forests (RF), and Decision Trees (DT). 

\cite{jindal2008opinion} classified fake reviews into three categories: non-reviews, brand-only reviews, and untruthful reviews. They trained a Logistic Regression classifier using duplicate or near-duplicate reviews as fake reviews, and the remaining reviews as truthful reviews. \cite{ott-etal-2011-finding} created the first gold-standard deceptive opinion spam dataset, gathering data through crowd-sourcing on Amazon Mechanical Turk. They discovered that although part-of-speech n-gram features provided reasonable accuracy in detecting fake reviews, the SVM classifier performed slightly better when psycho-linguistic features are incorporated. 

\cite{mukherjee2013fake} conducted a set of classification experiments utilizing SVM and NB algorithms with n-gram features. They examined both fake and non-fake reviews from Yelp.com and their findings indicated that the accuracy of fake review detection on Yelp's real-life data was only 67.8\%. \cite{banerjee2015using} employed RF, SVM and NB on the 15 Asia Hotel Reviews dataset, taking into account features such as writing style, word structure, level of details and cognition indicators. \cite{etaiwi2017impact} used the same set of classifiers on the Chicago Hotel Review dataset along with Decision Tree classifier and they also considered n-gram features in all experiments. 

\cite{shan2021conflicts} utilized lexicon-based features (SentiWordNet) with RF, NB, SVM, and multi-layer perceptrons (MLP). The RF algorithm achieved the highest accuracy of 92.9\%. \cite{mohawesh2021analysis} analyzed concept drift using SVM, LR, and pruning neural networks (PNN) on four different datasets: Yelp Chi, Yelp NYC, Yelp ZIP, and Yelp Consumer Electronic. There are also several works based on Decision Tree \citep{sedighi2017rlosd}, AdaBoost, JRip \citep{khurshid2017recital} and Support Vector Network \citep{hernandez2017cross}. 

Many researchers have found that using an ensemble approach with traditional machine learning models can be effective in detecting fake reviews. For instance, \cite{yao2021ensemble} utilized a combination of RF, XGBoost, CatBoost, Light Gradient-Boosting Machine (LightGBM), and Gradient-Boosting Decision Tree (GBDT) on the Yelp Chi dataset. They took into account both review and reviewer centric features and calculated F1-scores for hotel and restaurant domains separately using stacking and majority voting. \cite{khurshid2019enactment} and \cite{mani2018spam} also used ensemble models on the Yelp Chi and AMT dataset respectively, considering features such as TF-IDF and uni-gram, bi-grams. 

\cite{hammad2013approach} extended the methods used in previous papers to detect spam in Arabic opinion reviews. The dataset was created by crawling Arabic reviews from tripadvisor.com, booking.com, and agoda.ae and 26 features including review content, reviewer features, and hotel information features were used. Naïve Bayes was found to perform better than SVM with an F1-score of 99.59\% achieved using random oversampling.

\subsection{Deep Learning Approaches}
In contrast to conventional machine learning, deep learning techniques are capable of identifying significant features and comprehending the semantic context of textual data. Several studies have utilized Recurrent Neural Network (RNN), Convolutional Neural Network (CNN), hybrid approaches and Transformers to detect fake reviews.

\textbf{CNN based approaches:}
\cite{li2015learning} applied Convolutional Neural Networks (CNN) to identify spam opinions by learning document representations. The model used word vectors as input for both training and testing, and had a two-layer architecture consisting of a sentence layer for sentence composition and a document layer for transforming the sentence vector into a document vector. The study demonstrated the effectiveness of CNNs across different domains.

\cite{zhao2018towards} presented Convolutional Neural Networks for text analysis that preserves word order. They utilized Word2Vec and a pooling method that preserves word order rather than the typical max pooling to generate word vectors. The model then concatenated the obtained features from the pooling layer as an output layer. Results showed that the proposed method had a higher accuracy of 70.02\% compared to state-of-the-art methods.

\cite{wang2018detecting} applied a CNN with attention mechanism to detect if a review is deceptive due to behavior, language, or both. They employed a multi-layer perceptron to extract behavioral features, and a CNN to extract linguistic features. With the attention method, the model dynamically weighed linguistic and behavioral features. The study demonstrated that the proposed approach outperformed the current state-of-the-art methods, achieving an accuracy of 88.8\% for Hotel domain and 91\% for Restaurant domain.

\textbf{RNN based approaches:}
\cite{ren2016deceptive} employed a gated recurrent neural network model with attention mechanism to generate document representation for detecting fake reviews. Although they achieved improved results in the hotel and restaurant domains compared to the doctor domain, there was a high percentage of unidentified vocabulary. To overcome this issue, they used a logistic regression method with neural features and concatenated them with the discrete feature before the softmax layer. The results showed an accuracy of 81.3\% for Hotel, 87\% for Restaurant domain and 76.3\% for Doctor domain. However, the model suffers from time complexity.

\cite{wang2018lstm} developed a spam detection model using long short-term memory (LSTM) network that focuses on dictionary-based features. The model comprises three layers: input, LSTM hidden, and output layer. The output neuron's value determines whether the review is deceptive (1) or genuine (0). The study found that the LSTM model outperformed SVM with an 89.4\% accuracy in detecting deceptive reviews. However, the model solely focused on text and disregarded other important features such as metadata and behavioral features that could enhance performance.

\cite{liu2020incorporating} presented a bidirectional LSTM model by combining features such as parts of speech, first-person pronoun features, and document representation with word embedding (Glove). The model outperformed state-of-the-art methods in mixed domains with an 83.9\% accuracy and achieved high accuracy in individual domains with 83.9\%, 85.8\%, and 83.8\% on hotel, restaurant, and doctor domains respectively. The study revealed that first-person pronoun features are important in detecting deceptive reviews.

\cite{zeng2019review} showed that fake reviews often start or end with similar sentences, and the first and last sentences typically express stronger emotions than the middle sentence. The proposed ensemble model used four separate bidirectional LSTM models to encode the beginning, middle, and end of the reviews. The representations obtained from the four models were then concatenated and passed through self-attention and attention mechanisms to produce a final representation. The model achieved superior performance compared to SWNN and SAGA in one domain (hotel, doctor, and restaurant) with accuracy scores of 85.7\%, 84.7\% and 85.5\% respectively. In the mixed domain, the model achieved an accuracy score of 83.4\%. However, the model struggled to perform well in cross-domain scenarios, achieving an accuracy score of only 71.6\% on the restaurant domain and 60.5\% on the doctor domain.

\textbf{Hybrid and Transformer approaches:}
\cite{zhang2018dri} proposed DRI-RCNN which combines recurrent neural networks and convolutional neural networks. The model has four layers, including a convolutional layer to create a vector representation for each word, and a recurrent layer to learn context vectors for fake and real reviews. The experiments showed that this model achieved the highest accuracy of 82.9\% on AMT datasets.

\cite{dhamani2019using} proposed a model that combined an ensemble method with long short-term memory and character-level convolutional neural network to detect spam emails, spam reviews, and political statements across three different datasets. The proposed model utilized a simpler feature extraction method, such as n-grams. Interestingly, the model demonstrated the ability to transfer knowledge from labelled data in one domain to another domain.

\cite{shiblispam} applied deep learning techniques for detecting spam reviews using both labeled and unlabeled data. Their methods included Multi-Layer Perceptron (MLP), Convolutional Neural Network (CNN), and Long Short-Term Memory (LSTM). They also implemented an active learning algorithm that gradually labeled all unlabeled data by measuring the probability difference with a threshold value for accurate classification. The experimental results showed that the Convolutional Neural Network achieved the highest accuracy of 91.58\% on the Ott dataset and 95.56\% on the Yelp dataset, while the Long Short-Term Memory achieved the highest accuracy of 94.56\% on the Ott dataset and 96.75\% on the Yelp dataset. The models used Word2Vec for feature representation.

\cite{gupta2021leveraging} employed several pre-trained models such as BERT, RoBERTa, ALBERT, and DistilBERT to classify real and fake reviews. RoBERTa acquired the best results but the classification results need more improvements. \cite{mir2023online} employed BERT to obtain word embeddings from review texts. The word embeddings were then fed into various classifiers including SVM, Random Forests, Naive Bayes, and others. The outcomes revealed that SVM outperformed other classifiers by achieving an accuracy of 87.81\%. Although pre-trained transformer models have been widely used for SMS spam detection \citep{guo2023spam,rojas2021using,liu2021spam,sahmoud2022spam}, there are limited studies that have applied such models for detecting fake reviews. 

\section{Problem Definition \label{sec3}}
\noindent\textbf{Informal definition.} The problem is best framed as a supervised binary classification task where a review text written in Bengali language will be categorized into one of the two classes — \textit{fake, non-fake}.
\\
\textbf{Formal definition.} The objective of this study is to create a system that can identify fake Bengali review texts. The system will be capable of categorizing a given review text $x_{i}$, from a collection of $n$ Bengali review texts $X = $\{$x_{1}, x_{2}, ..., x_{n}$\}, into one of two pre-determined review categories: $C = $\{$c_{1}, c_{2}$\}. The purpose is to assign a review class $C_{i}$ to the text, where $c_{1}$ and $c_{2}$ correspond to genuine and fraudulent reviews respectively.

% \shibli{Pictorial definition\\
% For reference see data mining project}

\section{Corpus Creation \label{sec4}}
\begin{figure*}[h]
    \centering
    \includegraphics[scale=0.68]{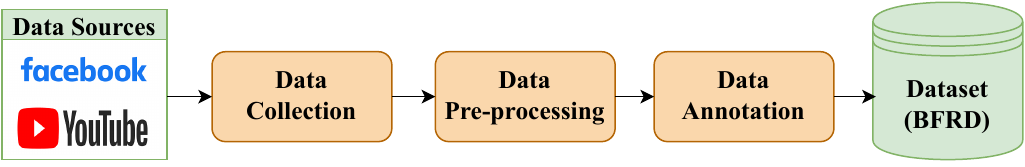}
    \caption{Dataset development pipeline}
    \label{fig:datasetpipeline}
\end{figure*}

%need to refine later
As per our exploration, there are no publicly available datasets on fake review detection task in Bengali language. Therefore, we have developed a Bengali fake review detection dataset
% \footnote{\url{https://github.com/shahariar-shibli/Bengali-Fake-Reviews-A-Benchmark-Dataset-and-Detection-System}} 
(we refer to as `BFRD' dataset) in this work. Figure \ref{fig:datasetpipeline} illustrates the dataset development pipeline. Firstly, we select suitable social media sources to obtain available food reviews. Next, we manually collect data and preprocess it to simplify the data annotation process. Before proceeding with data annotation, we conduct a data conversion task where we perform back transliteration to convert Romanized Bengali to Bengali, and we perform translation to include the meaning of English words in Bengali text as well as to convert digits. Finally, we enlist the help of four annotators to perform data annotation and construct the BFRD dataset. We will delve into the details of each component of the pipeline in the following subsections.

\subsection{Data Collection}
We manually gathered a total of \textbf{12,000} fake and non-fake review texts in Bengali across various social media platforms such as Facebook and YouTube. We manually collected potential texts from 15 public Bengali Facebook groups and two YouTube channels to create our dataset. Table \ref{tab:source} provides a comprehensive breakdown of the statistics for the sources\footnote{\url{https://pastebin.ubuntu.com/p/8gQMnCtRVw/}} from both Facebook and YouTube platforms. For our dataset, we limited our consideration to posts and comments from January 2019 to January 2023. Based on social media statistics\footnote{\url{https://gs.statcounter.com/social-media-stats/all/bangladesh}}, a significant proportion of social media users in Bangladesh, 89.65\% and 7.59\%, utilize Facebook and YouTube respectively. 

\begin{table}[h]
\caption{Statistics of some data collection sources (Facebook group/YouTube Channels). YT represents YouTube}
\label{tab:source}
%\resizebox{\textwidth}{!}{
\begin{tabular}{lccc}
\hline
\multicolumn{1}{c}{\textbf{\begin{tabular}[c]{@{}c@{}} Facebook group/\\ YouTube Channel\end{tabular}}} & \textbf{\begin{tabular}[c]{@{}c@{}}Popularity \\ (No. of group \\ members/ \\ Subscribers)\end{tabular}} & \textbf{\begin{tabular}[c]{@{}c@{}}Reaction \\ per post/ \\ video \\ (Avg)\end{tabular}} & \textbf{\begin{tabular}[c]{@{}c@{}}Engagement \\ (Frequency \\ of Posting)\end{tabular}} \\ \hline
FoodBank         & 2.1M         & 150& 150/day      \\ 
Food Bloggers BD & 942.3K       & 25 & 15/day       \\ 
Food N Foodies   & 244.4K       & 20 & 50/day       \\ 
Sylhet Foodies   & 193.3K       & 15 & 10/day       \\ 
Food Court       & 161.5K       & 15 & 25/week      \\ 
COMILLA FOODIES  & 152.5K       & 70 & 40/week      \\ 
Foodlovers of Narayanganj   & 132.4K       & 35 & 50/day       \\ 
FoodBank Mirpur  & 130.7K       & 25 & 30/day       \\ 
Food Bloggers Shonirakhra   & 111.2K       & 30 & 15/day       \\ 
Food Bank -Faridpur         & 84.3K        & 30 & 50/week      \\ 
Cox's Bazar Hotel \& Food Review & 67.0K        & 15 & 20/week      \\ 
FoodBank Dhaka   & 56.2K        & 10 & 12/week      \\ 
Feni Foodies     & 56.0K        & 30 & 10/week      \\ 
Food Bank Brahmanbaria      & 49.3K        & 45 & 50/week      \\ 
Efood Offer, Help \& Review & 39.8K        & 10 & 10/week      \\ 
Bangladeshi Food Reviewer (YT)  & 1.28M        & 4K & 20/month      \\ 
Rafsan TheChotoBhai (YT)  & 1.46M        & 50K & 5/month      \\ \hline
\end{tabular}
%}
\end{table}

The majority of the data instances in this study were gathered from Facebook since it is the primary platform used by Bengali social media users. Despite the abundance of food reviews available on Facebook, we found a relatively small amount of data available on YouTube. Our observations indicate that many individuals who view Bengali food vloggers on YouTube typically comment on the vlogger's appearance, presentation style, or personality, rather than the food itself. Furthermore, most people leave their reviews in English in the comments section. Recent statistics reveal that a mere 0.55\% of Twitter users in Bangladesh use the platform for social communication, and they predominantly communicate in English. As a result of the limited availability of Bengali texts related to food reviews on Twitter, we did not include Twitter data in our current work. To ensure the authenticity of the reviews collected, we followed specific criteria during the process of collecting both fake and non-fake reviews:
\begin{itemize}
    \item We gathered data by selecting posts that had received at least 200 reactions (likes, comments, and shares). 
    \item We took into account the comments and replies associated with the review posts.
    \item We also collected posts that advertised appealing offers and included digital menu cards.
    \item During data collection, we annotated review posts as fake if they were mentioned as paid in the comments, and we also marked review comments as fake if they received replies stating that the comment was fake.
    \item We identified certain individuals who spammed the same reviews in multiple groups and labeled their posts or comments as fake during data collection.
\end{itemize}

\subsection{Data Pre-processing}
To minimize the annotation workload and enhance consistency, we employed several preprocessing filters on the collected texts. After applying these filters, we were able to eliminate 961 texts, leaving \textbf{11,039} texts that were subsequently passed on to the human annotators for manual annotation. The steps followed in processing the texts are as follows:
\begin{enumerate}
    \item Repeated punctuations were removed.
    \item Texts with a length of fewer than three words were discarded as they do not contain any valuable information.
    \item Duplicate texts were removed.
\end{enumerate}

\subsection{Data Annotation}
In this section, we describe the annotation process of the Bengali fake review detection dataset (BFRD). Data annotation is challenging because of the subjective nature of the task. Based on some pre-defined criteria, annotators have to rely on their own interpretation and judgment to label the data that may lead to inconsistencies and biases in the annotations. Ensuring high quality, finding skilled annotators, and managing time effectively are other challenges that need to be overcome. To address these challenges, we identified annotators with domain knowledge. We established clear guidelines to ensure consistency and accuracy in the annotations. To maintain quality control, we implemented measures to assess the quality of the annotations.

\noindent \textbf{(a) \underline{Annotator Recruitment:}}
Initially, we invited 10 individuals to help us with the annotation process. We then assessed the trustworthiness score \citep{price2020six} of each participant by asking them to label 100 reviews. From the dataset, we randomly selected 80 reviews that were manually annotated by two authors and created 20 control samples. We interspersed one control sample after every four reviews that the participants labeled. The control samples were easy to label and were not previously known by the participants. After the annotation task, we analyzed how many control samples were correctly labeled by each participant.We set the threshold trustworthiness score for this task to be at least 90\%. Based on the evaluation, only four participants achieved a trustworthiness score above 90\%.

\begin{table}[h]
\centering
\caption{Information regarding the experience, expertise, and other pertinent details of the annotators}
\label{tab:annotator-identity}
%\resizebox{\textwidth}{!}{
\begin{tabular}{lcccc} 
\hline
\multicolumn{1}{c}{\textbf{Information}} & \begin{tabular}[c]{@{}c@{}}\textbf{First~}\\\textbf{~ Annotator~~}\end{tabular} & \begin{tabular}[c]{@{}c@{}}\textbf{Second}\\\textbf{~ Annotator~~}\end{tabular} & \begin{tabular}[c]{@{}c@{}}\textbf{Third}\\\textbf{~ Annotator~~}\end{tabular} & \begin{tabular}[c]{@{}c@{}}\textbf{Fourth}\\\textbf{~ Annotator~~}\end{tabular}  \\ 
\hline
Role       & Lecturer                         & Lecturer                         & \begin{tabular}[c]{@{}c@{}}Research\\Assistant\end{tabular}                    & Lecturer                          \\
Research field                           & NLP& NLP& NLP                             & NLP \\
Experience & 2 years                          & 1 year                           & 1 year                          & 3 years                           \\
Read food reviews                        & Yes& Yes& Yes                             & Yes \\
Written food reviews~ ~                  & No & No & Yes                             & Yes \\
\hline
\end{tabular}
%}
\end{table}

\noindent \textbf{(b) \underline{Annotator Identity:}}
We enlisted four individuals (with trustworthiness score above 90\%) to perform the task of manual annotation: a graduate student and three academic specialists. All of them are native Bengali speakers and possess a background in NLP research projects that involve data labeling, with experience ranging from 1 to 3 years. They all are active on social media and have a habit of reading food reviews before visiting a restaurant and writing their own reviews after trying a particular food item. The information regarding their experience, expertise, and other pertinent details is presented in Table \ref{tab:annotator-identity}.

\begin{figure}[h]
    \centering
    \includegraphics[scale=0.8]{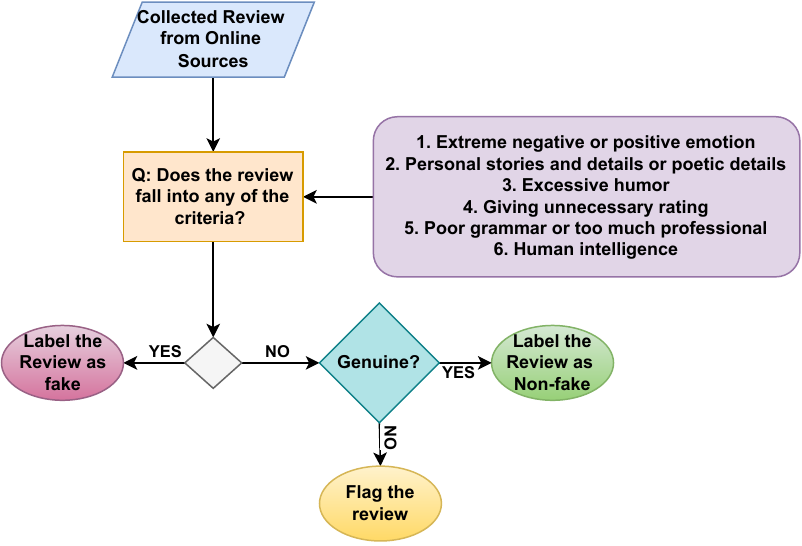}
    \caption{Data annotation procedure along with pre-defined guidelines}
    \label{fig:criteria}
\end{figure}

\noindent \textbf{(c) \underline{Annotation Guidelines:}}
Providing detailed annotation guidelines is essential for ensuring the quality of annotation and gaining a deeper understanding of the dataset, as individual interpretations and perceptions can differ significantly. During the annotation process, we request the annotators to follow the steps illustrated in Figure \ref{fig:criteria}. To identify a review as fake, we present six specific criteria. These criteria include the presence of extreme positive or negative emotions in the text, the inclusion of personal stories, details, or poetic descriptions, an excessive focus on humor rather than qualitative details, the use of unnecessary ratings such as 100 or 200 out of 10, texts that appear professionally written or exhibit poor grammar quality, and, lastly, the intelligence and judgment of the annotator. Prior to the annotation task, we provide the annotators with a few examples for each criterion and explain why a particular example should be categorized as a specific class. In Table \ref{tab:missannot}, we provide few instances of fake reviews from the dataset along with the corresponding criteria they satisfy. Some more instances can be found in Appendix \ref{secA7}.
\begin{table}[h]
\caption{Few annotated fake review instances with corresponding annotation criteria}
\label{tab:missannot}
\centering
% \resizebox{\columnwidth}{!}
\begin{tabular}{lc}
\hline

\multicolumn{1}{c}{\textbf{Fake reviews}}  & \textbf{Criteria} \\ \hline
\begin{tabular}[c]{@{}l@{}}\parbox[t]{5.2in}{\bng Aphar edekh eglam. pRthem ipj/ja edekh{I} sa{I}j Aar icejr Uper kRash ekhey eglam, Jaek bel Eekbaer  labh ET phar/s/T sa{I}T. tarpr AalLaHr nam iney ekhla shuru. tarpr ipj/jar duinyay Aaim Ek pithk. Ja{I} eHak kha{I}ya bYapk mja pa{I}ich ta{I} AaebgapLut Hey egichlam Aarik. Aphaerr Aar matR 4 idn baik. Erkm paint daem shrbt ekhet ca{I}el AepkKa na ker cel eJet paern bha{I}elak/s.} \\\parbox[t]{5.2in}{(I went to after seeing the offer. The first time I saw the pizza, I fell in love with the size and cheese, which is called love at first sight. Then the game started with the name of Allah. Then I was a traveler in the world of pizza. Anyway, I had a lot of fun eating it, so I got emotional. Only 4 days left for the offer. If you want to drink juice at the price of water, you can go without waiting brothers.)} \end{tabular} & 2, 3                 \\ \hline

\begin{tabular}[c]{@{}l@{}}\parbox[t]{5.2in}{Place: Castle {\bng salaemr paesh eriTNNG :10/10} Behavior :10/10 Price:1500-1485=15 tk {\bng brRiSh/Tr smy kha{I}et eglam. saeth ekU ichela na. khabarTa JethSh/T grm ichela. ipyaejr pirmanTa bhaela{I} ichela. smy epel Aabar Jaeba Aapnara{O} Jaebn. ktha Hec/ch Aaim} 10 out of 10 \bng idet eper Aenk dhnY.}\\

\parbox[t]{5.2in}{(Place: Next to Castle Salam Rating: 10/10 Behavior: 10/10 Price: 1500-1485=15 Tk. I went to eat during rain. There was no one with me. The food was hot enough. The amount of onion was good. I will go again if I have time. You can also go. I am very blessed to be able to give 10 out of 10.)}\end{tabular}  & 1, 4                 \\ \hline
\end{tabular}
\end{table}

During the manual annotation process, we exclude instances that have already been labeled as fake during the data collection phase. Each review text is individually annotated by three annotators. If a review fulfills any of the six criteria, the annotator classifies it as fake. On the other hand, if a review does not fit into any of the six criteria, the annotator marks it as genuine or flags it for further examination. We evaluate each text based on three labels. If all three labels match, we consider the associated label as final. However, if all three labels contain the flag, we discard the text. In cases where there is disagreement among the annotators, an expert annotator resolves the issue through discussion, determining whether to keep or remove the text. The final label for such texts is decided during the discussion. Within our dataset, we identified 3,433 texts where both flag marks and disagreements among annotators were present. Out of this subset, 1,443 texts were successfully resolved through discussions by an additional expert annotator. The remaining 1,990 texts were subsequently discarded. As a result, our final ``BFRD'' dataset consists of \textbf{9,049} processed and annotated texts. Among these, \textbf{1,339} texts were labeled as \textit{fake}, while the remaining \textbf{7,710} instances were labeled as \textit{non-fake}. For reference, Table \ref{tab:excluded} provides a few samples along with the reasoning behind their exclusion due to disagreements and confusion in assigning a class label (flag).

\begin{table}[h]
\caption{Some excluded review instances along with labels and the reason behind exclusion}
\label{tab:excluded}
\begin{tabular}{ccc}
\hline
\textbf{Reviews}& \textbf{Label}              & \textbf{Remarks}               \\ \hline
\begin{tabular}[c]{@{}c@{}}\parbox[t]{3.7in}{Buy one get one offer. {\bng Et/tgula Aa{I}eTm Aamar eshSh kret kSh/T Heyichl. es/TkTa Aar masrumTa Aamar khub bhaela elegech. Jara km baejeT ebshii ekhet can em{I}nil taedr jnY. mema, salad, ebhijeTbl, icekn es/Tk, mashrum phRa{I}D ra{I}s ikchu{I} baik ichlna. (saeth du{I}Ta} sweet \& sour soup  {\bng {{O} ichl}} Appetizer {\bng iHeseb) Aar ik ca{I} \bng beln? ruphTp ers/Tuern/T echaTkhaeTar medhY sun/dr EkTa jayga.} @only 500/= {\bng TRa{I} kret paern.} Taste: 9/10}\end{tabular} & \begin{tabular}[c]{@{}c@{}}fake, \\ Non-fake\end{tabular} & \begin{tabular}[c]{@{}c@{}}promotion of offers but \\ seems an honest review\end{tabular}   \\ \hline

\begin{tabular}[c]{@{}c@{}}\parbox[t]{3.7in}{{\bng sba{I}} Surprise gift {\bng paec/ch edkhet edkhet Aaim{O} Aaj epey eglam. Ekhn{O} ibshWas Hec/ch na.} Chillux Always {\bng Aamar} fvrt Restaurant {\bng ta{I} phuD iribhU Aalada ker edbar ikchu en{I}.} Thank you so much The Chillux for giving me a surprise gift.}\end{tabular}          & flag                        & \begin{tabular}[c]{@{}c@{}}Does not fall into the \\ criteria of being a review\end{tabular}
\\ \hline
\end{tabular}
\end{table}

\noindent \textbf{(d) \underline{Annotation Quality:}}
In order to measure the quality of the annotations, we assessed the level of agreement among the annotators using Fleiss' kappa score \citep{fleiss1971measuring}. This statistical measure is specifically designed to determine agreement among multiple annotators, extending Cohen's kappa \citep{cohen1960coefficient} which is typically used for two annotators. Fleiss' kappa compares the actual agreement observed among the annotators to the expected agreement based on chance allocation. The resulting score ranges from 0 to 1, where a score of 0 suggests no agreement beyond chance, and a score of 1 indicates perfect agreement. Intermediate values indicate different levels of agreement, varying based on the specific score obtained. Fleiss' kappa score can be calculated as follows:

\begin{equation}\label{kappa_eq}
    \kappa = \frac{P - P_{e}}{1 - P_{e}}
\end{equation}

In Eq.~\ref{kappa_eq}, $\kappa$ denotes the Fleiss' kappa score, where $P$ represents the observed proportion of agreement among the annotators, and $P_{e}$ represents the expected proportion of agreement due to chance. To compute the value of $P$, we add up the number of annotations for each category across all annotators and divide it by the product of the total number of instances and the total number of annotators. On the other hand, to calculate $P_{e}$, we sum up the squared proportion of annotations for each category and divide it by the square of the total number of instances.
\begin{table}[h]
\centering
\caption{Fleiss' kappa score on \textit{fake} and \textit{non-fake} class}
\label{tab:kappa-score}
\begin{tabular}{ccc} 
\hline
\textbf{Class} & \textbf{~ Kappa score~~} & \textbf{~Average~}     \\ 
\hline
Fake           & 0.83                     & \multirow{2}{*}{0.81}  \\
~ Non-fake~~   & 0.79                     &                        \\
\hline
\end{tabular}
\end{table}
In Table \ref{tab:kappa-score}, we provide the kappa score for each class. The \textit{fake} class demonstrates the highest agreement with a score of 0.83. The average kappa score of 0.81 reflects an almost perfect level of agreement among the annotators.

\begin{table}[h]
\caption{Class wise statistics of \textbf{BFRD} dataset}
\label{tab:stati}
\centering
\begin{tabular}{crr}
\hline
\textbf{Statistics} & \textbf{Fake} & \textbf{Non-fake} \\
\hline
Total words & 1,55,789      & 9,27,902          \\ \hline
Total unique words & 17,739        & 51,200            \\ \hline
Max Review length & 693           & 1,614             \\ \hline
\begin{tabular}[c]{@{}c@{}}Avg number of \\ words\end{tabular} & 116.35        & 120.35            \\ \hline
\begin{tabular}[c]{@{}c@{}}Avg number of\\  unique words\end{tabular} & 84.99         & 88.42            
\\ \hline
\end{tabular}
\end{table}

\subsection{Dataset Statistics}
\begin{figure}[h]
    \centering
    \includegraphics[scale=0.5]{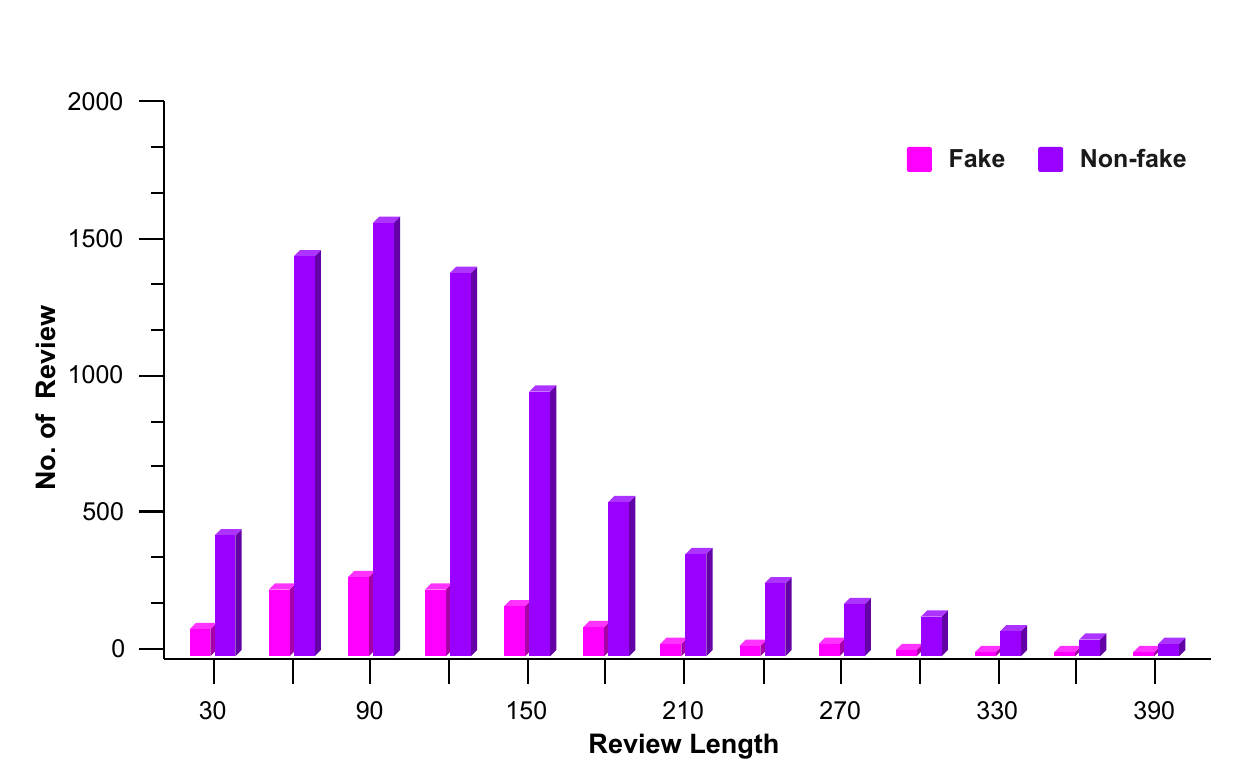}
    \caption{Class wise ratio of number of reviews with respect to the review length}
    \label{fig:reviewlength}
\end{figure}

From a total of 9,049 annotated reviews in the BFRD dataset, 1339 reviews are fake while 7,710 are non-fake. We provide some statistics regarding each class in Table \ref{tab:stati}. The maximum review length for fake reviews is 693, whereas the maximum review length for non-fake reviews is 1614, indicating that fake reviews tend to be shorter. Nonetheless, despite their length, the average number of unique words in fake reviews, around 85, is very close to the average number of unique words in non-fake reviews. This suggests that fake reviews use captivating or inventive word choices to catch the reader's attention. Figure \ref{fig:reviewlength} depicts the relationship between review length and the number of reviews for each class, indicating that the majority of reviews have a length between 60 and 120 words for both fake and non-fake classes.
Table \ref{tab:datasplit} presents the statistics of our data split along with our augmentation approach. The augmentation was only performed on the fake review class, with four augmentations progressively applied to each review. This means that we first created one augmentation for each review and performed experimentation before creating two, then three, and so on. We ensured that the dataset remained balanced at each augmentation level by taking an equal number of instances from both the fake and non-fake classes. The `No. of Augmentation' column indicates the level of augmentation, where a value of 3 in this column means we created three augmentations for each fake review, resulting in a total of 5,356 fake reviews (including the original 1,339). To maintain balance in the dataset during training, we randomly selected 5,356 non-fake reviews (from the total of 7,710). The dataset was then split into three sets for training, validation, and testing, with 80\%, 10\%, and 10\% of the data, respectively. We shuffled the data instances and split them into the three sets randomly. 
\begin{table}[h]
\caption{Summary of data split (Train, Validation and Test) of \textbf{BFRD} dataset. \textit{Aug} indicates the number of augmentations for a single fake review}
\label{tab:datasplit}
\centering
\begin{tabular}{c|ccccc|ccccc}
\hline
\textbf{Class}               & \multicolumn{5}{c|}{\textbf{Fake}}                                                                                   & \multicolumn{5}{c}{\textbf{Non-fake}}                                                                               \\ \hline
\textbf{No. of Augmentation} & \multicolumn{1}{c}{0}    & \multicolumn{1}{c}{1}    & \multicolumn{1}{c}{2}    & \multicolumn{1}{c}{3}    & 4    & \multicolumn{1}{c}{0}    & \multicolumn{1}{c}{1}    & \multicolumn{1}{c}{2}    & \multicolumn{1}{c}{3}    & 4    \\ \hline
\textbf{Train}               & \multicolumn{1}{c}{1071} & \multicolumn{1}{c}{2142} & \multicolumn{1}{c}{3213} & \multicolumn{1}{c}{4285} & 5356 & \multicolumn{1}{c}{1071} & \multicolumn{1}{c}{2142} & \multicolumn{1}{c}{3214} & \multicolumn{1}{c}{4285} & 5356 \\ 
\textbf{Validation}          & \multicolumn{1}{c}{134}  & \multicolumn{1}{c}{268}  & \multicolumn{1}{c}{401}  & \multicolumn{1}{c}{535}  & 670  & \multicolumn{1}{c}{134}  & \multicolumn{1}{c}{268}  & \multicolumn{1}{c}{402}  & \multicolumn{1}{c}{535}  & 669  \\ 
\textbf{Test}                & \multicolumn{1}{c}{134}  & \multicolumn{1}{c}{268}  & \multicolumn{1}{c}{402}  & \multicolumn{1}{c}{536}  & 669  & \multicolumn{1}{c}{134}  & \multicolumn{1}{c}{268}  & \multicolumn{1}{c}{402}  & \multicolumn{1}{c}{536}  & 670  \\ \hline
\textbf{Total}               & \multicolumn{1}{c}{1339} & \multicolumn{1}{c}{2678} & \multicolumn{1}{c}{4016} & \multicolumn{1}{c}{5356} & 6695 & \multicolumn{1}{c}{1339} & \multicolumn{1}{c}{2678} & \multicolumn{1}{c}{4018} & \multicolumn{1}{c}{5356} & 6695 \\ \hline
\end{tabular}
\end{table}
\section{Methodology \label{sec5}}
\begin{figure}[h]
    \centering
        \includegraphics[scale=0.8]{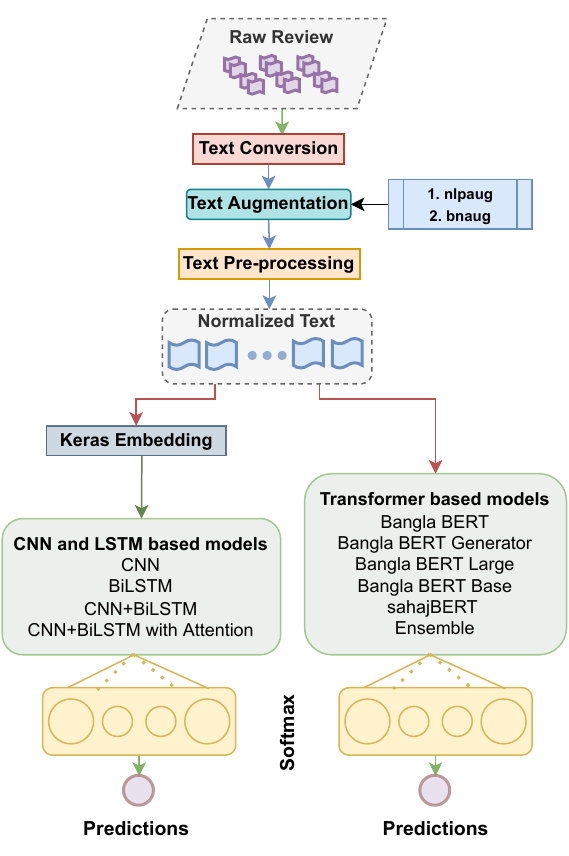}
    \caption{Schematic diagram of Bengali fake review detection system}
    \label{fig:methodology}
\end{figure}

Figure \ref{fig:methodology} illustrates the schematic diagram of our proposed approach for Bengali fake review detection task. Our proposed approach consists of four key steps: conversion, augmentation, pre-processing, and detection. In the first step, we convert English words in the text to their corresponding Bengali meanings, and also transliterate Romanized Bengali words to Bengali. The second step involves text augmentation, which helps to create more fake instances. After that, we pre-process the texts to prepare them for detection. Finally, we apply a range of deep learning techniques, as well as pre-trained transformers and ensemble methods, to detect fake reviews. We discuss each of these steps in the proposed methodology below.

While collecting and initially processing the data, we made two noteworthy observations. First, a considerable number of reviews contained Bengali words written using English letters, commonly referred to as Romanized Bengali. In fact, some reviews even consisted of one or two complete sentences entirely in Romanized Bengali. Second, we found that many reviews included English words and digits. These observations highlight the linguistic diversity present within the reviews and emphasize the need for appropriate handling of both Bengali and English elements during further analysis. If we exclude these texts that contain Romanized Bengali or English words, we risk losing a considerable amount of data given the scarcity of data available to us. Furthermore, our proposed detection model is designed to work best with Bengali words. Therefore, we developed an algorithm for text conversion purpose that can back-transliterate the Romanized Bengali words into Bengali words and at the same time translate the English words into their Bengali equivalent meanings in a text. We also convert English digits to their Bengali counterparts. We present our text conversion procedure that involves transliteration and translation in Algorithm \ref{algo:conversion}.

The algorithm starts by taking raw review sentences as input. For each sentence, it performs some initial processing, such as allocating spaces before and after punctuation marks, replacing newlines and emojis with spaces. Next, the individual words are extracted from the sentence. 
\begin{algorithm}
\caption{Algorithm of the proposed text conversion pipeline}
\label{algo:conversion}
\begin{algorithmic}[1]

\Function{Conversion}{$word, method, src, des$}
    \State temp = {`` '' }
    \If {$method$ is {`gt'}}\Comment{gt = google translate}
        \State temp = Translate $word$ with google translate ($src$ - $des$)
    \Else
        \State temp = Transliterate $word$ with bnbphonetic parser ($src$ - $des$)
    \EndIf
    \State \Return temp
\EndFunction\\
% \newline
\Function{text\_conversion}{Raw Sentences}       
    \For{each sentence}
        \State $converted\_sent$ = {`` '' }
        \State Allocate space before and after punctuations
        \State Replace \textbf{newline} and \textbf{emojis} with a space
        % \State Replace any type of emoji with a space
        \State Split words from sentence
        \State Split mixed English-Bengali words using regex
        \For{each $word$}
            \If{$word$  is not {``$  \vert  $''}} \Comment{End of Sentence}
                \If{$word$ is in English dictionary}
                    \If{$word$ is not a digit}
                        \State $converted\_sent$+=Conversion($word$,en,bn,`gt')

                    \Else
                        \State $converted\_sent$+=Conversion($word$,bn,bn,`gt')
                    \EndIf

                \Else
                    \If{$word$ is a digit}
                        \State $converted\_sent$+=Conversion($word$,bn,bn,`gt')
                    \Else
                        \If{$word$ is not Bengali}
                            \If{$word$ is English}
 \State $converted\_sent$+=Conversion($word$,en,bn,`gt')  
                            \Else \Comment{Romanized Bengali}
 \State $converted\_sent$+=Conversion($word$,bn,bn,`bnb')
                            \EndIf
                        \Else
                            \State $converted\_sent$$+=$ $word$
                        \EndIf
                    \EndIf
                \EndIf
            \Else
                \State $converted\_sent$$ += $$word$
            \EndIf
        \EndFor
        \State \Return $converted\_sent$
    \EndFor
\EndFunction

\end{algorithmic}
\end{algorithm}
Since the reviews are written by humans, we observed instances where English and Bengali words in a sentence were concatenated. This occurred due to typing errors, such as missing appropriate spaces between words. For instance, we came across a sentence like ``{\bng khabar pirebshn Aar{O}better Het pareta}'' \textit{[Food service could have been muchbetter]} where the word ``{\bng Aar{O}better}'' \textit{[muchbetter]} was concatenated due to a missing space. The algorithm identifies and separates such concatenated English-Bengali words using regular expressions. The algorithm then proceeds to process each word individually. If a word is not an end-of-sentence punctuation mark ({\bng .}), we check whether it is an English word by utilizing the dictionary from the Enchant library\footnote{\url{https://abiword.github.io/enchant/}}. If the algorithm detects an English word or digits, it simply translates them into their corresponding Bengali words or digits using the Googletrans library\footnote{\url{https://py-googletrans.readthedocs.io/en/latest/}}. Otherwise, there are two possibilities: the word is either Bengali or Romanized Bengali. If the word is Bengali, the algorithm directly concatenates it back into the sentence. If it is Romanized Bengali, the algorithm employs a Bengali phonetic parser library\footnote{\url{https://github.com/porimol/bnbphoneticparser}} to back-transliterate the word into Bengali. When the algorithm encounters an end-of-sentence punctuation mark, it reconstructs the original sentence by concatenating all the processed and initial words. The algorithm uses a function called \textit{CONVERSION} to perform these translation and back-transliteration processes. The argument \textit{'word'} is the raw token, \textit{'method'} is the name of the specific library, and \textit{'src'} and \textit{'des'} are the source and destination languages to convert. We present some of the converted text instances through the algorithm in Table \ref{tab:converted-texts}. A popular LLM model, Llama 2 (70B)\footnote{\url{https://www.llama2.ai/}}, was also explored to convert the text given in Table \ref{tab:converted-texts}, but it was unable to convert the majority of the texts in a meaningful way. Table \ref{tab:converted-textsllama} displays the converted texts with Llama 2. Appendix \ref{secA8} provides an explanation for the performance of the LLM model on our data. The reviews posted on social media platforms are code-mixed. Dealing with such content is difficult for any model, particularly those developed predominantly in English. Our proposed pipeline is created using a variety of open-source tools that are publicly available online. Fine-tuning and widespread use of LLMs for large-scale data conversion can be costly. LLMs can be a great future alternative to our proposed pipeline if some post-processing of the texts can be done on the outputs of the models.

\begin{table}[h]
\caption{Few example instances before and after applying the text conversion pipeline}
\label{tab:converted-texts}
%\resizebox{\textwidth}{!}{
\begin{tabular}{c|c}
\hline
\textbf{Original Text}& \textbf{Converted Text}         \\ \hline
\begin{tabular}[c]{@{}c@{}}\parbox[t]{3in}{Best pizza offer for Mirpur peoples Buy 1 Get 1 free 1.Mashroom lovers pizza.2.Meat lovers pizza. Choto bhaiyer jsc xam sesh kore gelm n e offer ta kheye ashlam onk tasty chilo n tader bebohar o valo chilo.. mane ak kothay paisa osul Taste:9/10 Price: 330tk Location: The Hub Restaurant (60 feet) — at The Hub Restaurant}\end{tabular}            & \begin{tabular}[c]{@{}c@{}}\parbox[t]{3in}{\bng esra ipj/ja Aphar jnY imrpur elaekra ekna 1 pa{O}Ja 1 ibnamuuelY 1 mashrum epRmiira ipj/ja 2 maNNGs epRmiira ipj/ja echaT bha{I}Jar ejEsis jYam eshSh ekaer eglm En {I} Aphar Ta ekhey Aaslam {O}NG/k susWadu ichela En  taedr ebebaHar {O} bhaela ichela mYan Eek ekathay pa{I}sa {O}sul sWad 9 10 dam 330 elaekshn dY Hab erNNes/tara 60 iphT ET dY Hab erNNes/tara}\end{tabular}                            
\\ \hline
\begin{tabular}[c]{@{}c@{}}\parbox[t]{3in}{
Cafe Famous Wari {\bng pRay ebsh keykidn Aaeg igeyichlam.... Eedr bar/gar gula bhaela laeg Aamar. ekan jaygar saeth tulna krb na  teb bhaela laeg Aamar sbsmy icekn bar/gar TRa{I} kra H{I}es sb khaen ephmaes egel{O} E{I}Ta{I} TRa{I} kra Hy Ebar Jas/T EkTu ecNJ/j krlam ibph Ta iney ibph ebkn-350 Taka iDRNNGk/s-40 Taka ker ephRNJ/c phRa{I}s-100 Taka \#elaekshn: {O}yarii} (yellow {\bng Er Uperr tlay) \#Aa{I}eTm:} Beef n bacon \bng \#dam: 350 Taka}\\ \end{tabular} & \begin{tabular}[c]{@{}c@{}}\parbox[t]{3in}{\bng kYaeph ibkhYat {O}Jarii pRay ebsh keykidn Aaeg igeyichlam. Eedr bar/gar gula bhaela laeg Aamar. ekan jaygar saeth tulna krb na teb bhaela laeg Aamar sbsmy icekn bar/gar TRa{I} kra H{I}es sb khaen ephmaes egel{O} E{I}Ta{I} TRa{I} kra Hy Ebar Jas/T EkTu ecNJ/j krlam ibph Ta iney ibph ebkn 350 Taka iDRNNGk/s 40 Taka ker ephRNJ/c phRa{I}s 100 Taka elaekshn {O}yarii Hlud Er Uperr tlay Aa{I}eTm grur maNNGs En ebkn dam 350 Taka}\end{tabular} \\ \hline
\begin{tabular}[c]{@{}c@{}}\parbox[t]{3in}{Berger Xpress \bng E bn/dhuedr saeth bar/gar ekhet egichlam. per inejr bar/gar inej{I} pa{I}lam na}\end{tabular}            & \begin{tabular}[c]{@{}c@{}}\parbox[t]{3in}{\bng bar/gar Ek/sepRs E bn/dhuedr saeth bar/gar ekhet egichlam per inejr bar/gar inej{I} pa{I}lam na}\end{tabular}           \\ \hline
\end{tabular}
%}
\end{table}

\subsection{Text Augmentation}
\begin{figure}[ht]
		\centering
            \includegraphics[width=0.8\linewidth]{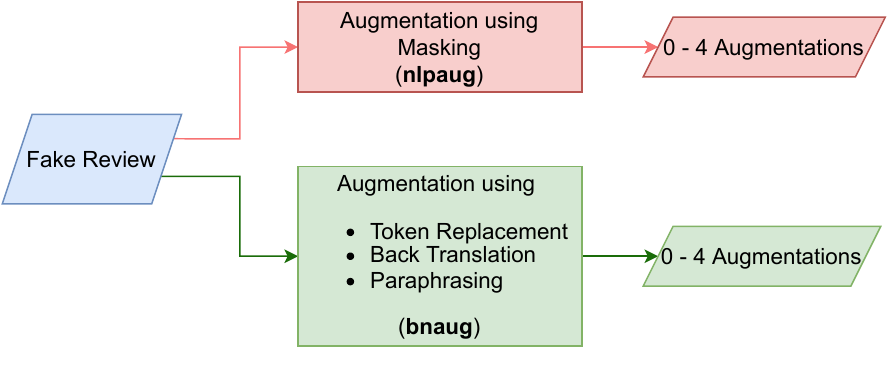}
		\caption{Augmentation pipeline using \textit{nlpaug} and \textit{bnaug}}
		\label{fig:augmentation}
\end{figure}

Since we have a limited number of fake reviews (only 1339) in our dataset, we decided to only augment the fake reviews in order to improve the quality of fake reviews. To achieve this, we utilized two available python libraries for text augmentation: \textit{nlpaug} \citep{ma2019nlpaug} and \textit{bnaug}\footnote{\url{https://github.com/sagorbrur/bnaug}}. The augmentation pipeline is shown in Figure \ref{fig:augmentation}, where we pass a single fake review to either nlpaug or bnaug library. We create augmented instances progressively for experimentation purposes, starting with one augmentation for each review, then two, and so on.
Specifically, we created a maximum of four augmentations using the nlpaug library, which applies random 15\% masking on the input sequence and employs various BERT variant models pre-trained for Bengali language. We used four different models, including \textit{BanglaBERT Base} \citep{Sagor_2020}, \textit{BanglaBERT} \citep{bhattacharjee-etal-2022-banglabert}, \textit{BanglaBERT Generator} \citep{bhattacharjee-etal-2022-banglabert}, and \textit{sahajBERT}\footnote{\url{https://huggingface.co/neuropark/sahajBERT}}, to leverage the linguistic generalization capability of the language-specific models. We conducted several experiments with varied numbers of random masking and discovered that the percentage of random masking had little effect on the model outcomes. It should be emphasized that the replacement of any word in the nlpaug augmentation approach takes into account the embedding of the replaced word. In this augmentation process, the new word is chosen ensuring that it has a similar embedding to the previous one. So the replaced embedding has little effect on the model's output. As shown in Table \ref{tab:random_sampling}, the results for the Bangla BERT model's weighted f1 score range from 0.954 to 0.981 for varied percentages of random replacement, with a variance of approximately 0.027\%.

% \usepackage{multirow}
\begin{table}[]
\caption{Performance comparison of five individual experiments using varying percentages of random masking during augmentation for the \textit{Bangla BERT} model utilizing \textit{nlpaug} technique}
\label{tab:random_sampling}
\begin{tabular}{c|ccc|ccc|c|c|c}
\hline
\multirow{2}{*}{\textbf{\begin{tabular}[c]{@{}c@{}}Percentage of \\ Augmentation\end{tabular}}} & \multicolumn{3}{c|}{\textbf{Fake}}                                              & \multicolumn{3}{c|}{\textbf{Non-Fake}}                                          & \multirow{2}{*}{\textbf{WF1}} & \multirow{2}{*}{\textbf{{\begin{tabular}[c]{@{}c@{}}ROC- \\ AUC\end{tabular}}}} & \multirow{2}{*}{\textbf{MCC}} \\ \cline{2-7} & \multicolumn{1}{c}{\textbf{P}} & \multicolumn{1}{c}{\textbf{R}} & \textbf{F1} & \multicolumn{1}{c}{\textbf{P}} & \multicolumn{1}{c}{\textbf{R}} & \textbf{F1} &                               &                                   &                               \\ \hline
8                                                                                               & \multicolumn{1}{c}{0.964}      & \multicolumn{1}{c}{0.958}      & 0.961       & \multicolumn{1}{c}{0.958}      & \multicolumn{1}{c}{0.964}      & 0.961       & 0.961                         & 0.961                             & 0.922                         \\ 
10                                                                                              & \multicolumn{1}{c}{0.960}      & \multicolumn{1}{c}{0.961}      & 0.960       & \multicolumn{1}{c}{0.961}      & \multicolumn{1}{c}{0.960}      & 0.960       & 0.960                         & 0.960                             & 0.921                         \\ 
12                                                                                              & \multicolumn{1}{c}{0.971}      & \multicolumn{1}{c}{0.937}      & 0.954       & \multicolumn{1}{c}{0.939}      & \multicolumn{1}{c}{0.971}      & 0.955       & 0.954                         & 0.954                             & 0.909                         \\ 
15                                                                                              & \multicolumn{1}{c}{0.976}      & \multicolumn{1}{c}{0.985}      & 0.981       & \multicolumn{1}{c}{0.985}      & \multicolumn{1}{c}{0.976}      & 0.980       & 0.981                         & 0.981                             & 0.961                         \\ 
18                                                                                               & \multicolumn{1}{c}{0.974}      & \multicolumn{1}{c}{0.954}      & 0.964       & \multicolumn{1}{c}{0.955}      & \multicolumn{1}{c}{0.975}      & 0.964       & 0.964                         & 0.964                             & 0.928                         \\ \hline
\end{tabular}
\end{table}

\begin{table}[h]
\caption{Some augmented fake reviews generated using \textit{nlpaug} and \textit{bnaug}}
\label{tab:aug-example}
\begin{tabular}{cc}
\hline
\textbf{Type}           & \textbf{Review} \\
\hline
Converted Review        & \begin{tabular}[c]{@{}c@{}}\parbox[t]{3.4in}{{\bng jiibn Aamar dhnY Hela   ibph Aacarii ikhcuir ekhey sitY Asadharn eTs/T 10 10 dam 150 Taka icekn 130 Taka   ekalaHl ers/Tuern/T Dhaka kr/mas Er ibpriiet imrpur 2 Dhaka.}\\ (My life is blessed after  eating beef achari khichuri. Really awesome. Test 10/10 price 150 taka   chicken 130 taka noisy restaurant. Opposite Dhaka Commerce Mirpur 2 Dhaka.)}\end{tabular} \\ \hline
nlpaug (1)              & \begin{tabular}[c]{@{}c@{}}\parbox[t]{3.4in}{\bng jiibn Aamar dhnY Har/T ibph en{I} Aar ekhey sitY Ek eTs/T 10 10 iminT 150 Taka icekn 130 Taka ekalaHl ers/Tuern/T Dhaka Aamar pirbaerr ibpriiet cYaenl Tu Dhaka 1}\end{tabular}\\ \hline
nlpaug (2)              & \begin{tabular}[c]{@{}c@{}}\parbox[t]{3.4in}{\bng jiibn Aamar dhnY Hela ibph Aacarii ekk ekhey sitY ik khabar 10 10 dam 150 Ta icekn 130 Taka nrmal ers/Tuern/T mainkgNJ/j 35 bhaetr ibpriiet mainkgNJ/j 2 Dhaka 1}\end{tabular}  \\ \hline
nlpaug (3)              & \begin{tabular}[c]{@{}c@{}}\parbox[t]{3.4in}{\bng cakrii Aamar dhnY Hela bhat Aar ikhcuir ekhey Aar Asadharn kaj krlam 10 dam 150 Taka maer/kiTNNG Tu jn ekalaHl ers/Tuern/T Dhaka kr/mas Er ibpriiet imrpur ker/paerT Dhaka.}\end{tabular}                          \\ \hline
nlpaug (4)              & \begin{tabular}[c]{@{}c@{}}\parbox[t]{3.4in}{\bng jiibn Aanen/d dhnY Hela ibph ers/Tueren/T EkTu ekhey eJet Hy eTs/T laNJ/c 10 Dlar 150 Taka icekn 600 Taka ekalaHl ibic/chn/n Dhaka kr/mas Aamar ibpriiet imrpur 2 Dhaka.}\end{tabular}                        \\ \hline
bnaug (word2vec)        & \begin{tabular}[c]{@{}c@{}}\parbox[t]{3.4in}{\bng jiibnek Aapnar dhnY Hela ibph Aacarii ikhcuir ekhey sitY Asadharn {O}yaneD 10 10 dam 150 pysa icekn 160 pysa ekalaHl ers/Tuern/T Dhaka kr/mas -Er ibpriiet imrpur 2 Dhaka.}\end{tabular}                       \\ \hline
bnaug (glove)            & \begin{tabular}[c]{@{}c@{}}\parbox[t]{3.4in}{\bng jiibn Aaim dhnY Hela ibph Aacarii ikhcuir ekhey sitY Asadharn {O}iDAa{I} 10 10 dam 150 bYaey icekn 140 bYaey ekalaHl ers/Tuern/T rajshaHii {I}n/DaiSh/TR Er ibpriiet imrpur 2 rajshaHii.}\end{tabular}                \\ \hline
bnaug (back translation) & \begin{tabular}[c]{@{}c@{}}\parbox[t]{3.4in}{\bng jiibn Aamar dhnY Hela ibph trkair ekhey sitY AsadharN. Dhaka kr/mas Er ibpriiet imrpur 2.}\end{tabular}      \\ \hline
bnaug (paraphrasing)    & \begin{tabular}[c]{@{}c@{}}\parbox[t]{3.4in}{\bng Aamar jiibn sukhii grur maNNGesr ikhcuir ekhey, sitYkaerr AsadharN priikKa Hec/ch 10 10 Taka murgii 130 Taka, ekalaHl ers/Tuern/T Dhaka kr/mas Er ibpriiet imrpur 2 Dhaka.}\end{tabular}     \\ \hline                    
\end{tabular}
\end{table}
Table \ref{tab:aug-example} shows examples of augmentations generated by the four models for a single review instance, indicating that nlpaug randomly masks tokens and replaces them using the corresponding pre-trained model's contextual information. To explore other augmentation techniques, we also utilized the bnaug library for random mask-based generation using pre-trained Bengali GloVe\footnote{\url{https://huggingface.co/sagorsarker/bangla-glove-vectors}} and Word2Vec\footnote{\url{https://huggingface.co/sagorsarker/bangla_word2vec}} embeddings, as well as for back translation and paraphrasing. For back translation, we employed pre-trained Bengali T5 neural machine translation model \citep{bhattacharjee2022banglanlg} that translates Bengali text to English and back to Bengali. For paraphrase generation, we used pre-trained Bengali T5 paraphrase model \citep{akil2022banglaparaphrase}. As per the example instances shown in Table \ref{tab:aug-example}, we can observe that the
Bengali word {\bng eTs/T} \textit{[taste]} has been replaced by \textit{sahajBERT} (nlpaug), word2vec (bnaug), glove (bnaug), and paraphrase (bnaug) with the Bengali words {\bng kaj, {O}yaneD, {O}iDAa{I}, priikKa} \textit{[work, one day, ODI, test]} respectively. The variations in meaning of the word ``taste'' could be due to a number of factors such as the context in which the word appears, the specific pre-trained models used for augmentation, and the quality and quantity of the training data used to train these models. Moreover, the word embeddings (GloVe, Word2Vec) are based on the distributional hypothesis that words that occur in similar contexts have similar meanings. This means that the context of the word ``taste'' is ambiguous or not well represented in the training data, which is why the resulting embeddings might not be able to accurately capture the actual meaning.

\subsection{Pre-processing}
\textbf{\underline{Text Pre-processing}:} To clean all the Bengali texts, we employed a Python module for text normalization as outlined in \citep{hasan-etal-2020-low}. The process of text cleaning involved a number of steps, including managing multiple white spaces, handling URLs, replacing emojis, fixing double or single quotes, and replacing Unicode characters.

\textbf{\underline{Feature Extraction}:} To obtain the embedding features, we utilized the default \textit{Keras embedding layer} which required creating a vocabulary of size 25000 to map each word in a text to its corresponding index in the vocabulary. To convert variable length sequences into fixed length vectors of size 512, we employed the \textit{Keras pad\_sequences} method which removed extra values from long sequences and padded short ones with zeros. The embedding dimension was set to 300, and as a result, the embedding layer converted a text of length 512 into a matrix of size $512\times300$.

\subsection{Detection Methods}
To detect fake reviews in Bengali, we have utilized three different methods: CNN and LSTM-based, transformer-based and ensemble-based. In the CNN and LSTM-based approach, we have employed four models: CNN \citep{lecun1998gradient}, BiLSTM \citep{graves2005framewise}, CNN-BiLSTM \citep{rhanoui2019cnn}, and CNN-BiLSTM with attention mechanism \citep{lu2021cnn}.
In the transformer-based approach, we fine-tuned three
ELECTRA \citep{clark2020electra} variants, two BERT \citep{devlin-etal-2019-bert} variants, and one ALBERT \citep{lan2019albert} variant pre-trained Bengali language models. For the ensemble-based approach, we proposed a method of combining the results of multiple pre-trained transformer models for better accuracy. We discuss all the approaches below in detail.

\subsubsection{CNN and LSTM based models}
We present the architecture details of all the CNN and LSTM-based models in Table \ref{tab:dlHyper}.

\noindent \textbf{(a) \underline{CNN}}: Convolutional Neural Network \citep{lecun1998gradient} is a deep learning architecture that applies filters over word embeddings or character sequences to capture local patterns and features such as n-grams or word combinations. They are particularly useful for tasks where local context plays a crucial role, such fake review detection. The convolutional layers slide filters across the input, extracting local information. They can capture local patterns regardless of their position in the text, making them suitable for tasks with significant local dependencies. The extracted features are then passed through fully connected layers for classification. CNNs leverage shared weights to learn hierarchical representations of the text data.\\
\noindent \textbf{(b) \underline{BiLSTM}}: Bidirectional Long Short-Term Memory \citep{graves2005framewise} is an extension of the LSTM architecture, which is capable of modeling long-term dependencies in sequential data which is valuable for text classification tasks where the meaning of a word can depend on the surrounding words. It processes the input in both forward and backward directions simultaneously. BiLSTMs utilize memory cells and gates to capture and update information across different time steps. By processing the input sequence in both directions, the network can learn from past and future contexts, enhancing the model's ability to understand the text.\\
\noindent \textbf{(c) \underline{CNN-BiLSTM}}: CNN-BiLSTM \citep{rhanoui2019cnn} combines the strengths of both CNN and BiLSTM architectures for text classification tasks. It leverages the local feature extraction capabilities of CNNs and the contextual understanding of BiLSTMs. The model typically starts with a CNN layer to extract local features, followed by a BiLSTM layer to capture sequential dependencies. The output of the BiLSTM is then passed through fully connected layers for classification. By combining both, the model can learn both local and global representations of the text, leading to enhanced representation learning and improved classification accuracy.\\
\noindent \textbf{(d) \underline{CNN-BiLSTM with Attention mechanism}}: CNN-BiLSTM with attention mechanism \citep{lu2021cnn} extends the CNN-BiLSTM model by incorporating an attention mechanism. Attention allows the model to focus on relevant parts of the input sequence. The model first applies a CNN layer to extract local features and then uses a BiLSTM layer to capture sequential dependencies. The attention mechanism assigns weights to different parts of the input sequence which enables the model to dynamically attend to different parts of the text, focusing on the most relevant words or phrases contributing to the classification decision. Moreover, the combination of CNN and BiLSTM layers provides a comprehensive representation learning framework that leverages both local and global dependencies, leading to improved performance in text classification.

\begin{table}[H]
\centering
\caption{Architecture details of the deep learning models}
\label{tab:dlHyper}
\begin{tabular}{ccccc}
\hline
\textbf{Architecture}                            & \textbf{CNN}        & \textbf{BiLSTM}   & \begin{tabular}[c]{@{}c@{}}\textbf{CNN} \\ \textbf{BiLSTM}\end{tabular} & \begin{tabular}[c]{@{}c@{}}\textbf{CNN-BiLSTM}\\\textbf{with}\\ \textbf{Attention}\end{tabular} \\ \hline
Input   length              & \multicolumn{4}{c}{512}           \\ \hline
Embedding Dimension         & \multicolumn{4}{c}{300}           \\ \hline
Filters (Layer-1)           & 200        & -         & 200                      & 512         \\
Kernel Size                 & 3          & -         & 3                        & 4           \\
Filters (Layer-2)           & 100        & -         & 200                      & 256         \\
Kernel Size                 & 3          & -         & 3                        & 3           \\
Filters (Layer-3)           & -          & -         & -                        & 128         \\
Kernel Size                 & -          & -         & -                        & 2           \\
Pooling Type                & max        & -         & max                      & max         \\
BiLSTM Cell (Layer-1)      &  -          & 128       & 128                      & 200         \\
dropout                     &  -          & 0.5       & 0.5                      & -           \\
BiLSTM Cell (Layer-2)      &    -        & 128       & 128                      & -           \\
Attention Vector            & -          & -         & -                        & yes         \\ 
Total Parameter             & 10,288,202 & 7,799,998 & 8,049,198                & 8,777,751   \\ \hline
Activation                  & \multicolumn{4}{c}{ReLU}          \\ \hline
Activation   (Output layer) & \multicolumn{4}{c}{softmax}       \\ \hline
Loss                        & \multicolumn{4}{c}{categorical\_crossentropy}                   \\ \hline
Optimizer                   & \multicolumn{4}{c}{Adam}  \\ \hline       
\end{tabular}
\end{table}

\subsubsection{Transformer based Models}
Table \ref{tab:transhyper} contains the architecture configurations of the pre-trained transformer models.

\noindent \textbf{(a) \underline{BERT based Transformer Models}}: The Bidirectional Encoder Representations from Transformers (BERT) \citep{devlin-etal-2019-bert} is constructed upon a deep learning framework where the connections between input and output elements are established, and the weights are adaptively determined based on their relationship. What sets BERT apart is its ability to perform bidirectional training, allowing the language model to grasp the context of a word by considering its surrounding words rather than solely focusing on the preceding or succeeding word. \textit{BanglaBERT Base}\footnote{\url{https://github.com/sagorbrur/bangla-bert}} follows the same architecture as the original BERT model.\\ 
\noindent \textbf{(b) \underline{ELECTRA based Transformer Models}}: Efficiently Learning an Encoder that Classifies Token Replacements Accurately (ELECTRA) \citep{clark2020electra} utilizes a pre-training task that revolves around identifying replaced tokens within the input sequence. This involves training a discriminator model to recognize the tokens that have been replaced in a corrupted sequence, while a generator model is simultaneously trained to predict the original tokens for the masked out ones. This setup resembles a generative adversarial network training system but without the adversarial aspect, as the generator is not trained to deceive the discriminator. \textit{BanglaBERT} \citep{bhattacharjee-etal-2022-banglabert} serves as the ELECTRA discriminator model, while \textit{BanglaBERT Generator } \citep{bhattacharjee-etal-2022-banglabert} is an ELECTRA generator model that has been pre-trained using the masked language modeling (MLM) objective on substantial amounts of Bengali texts. \textit{BanglaBERT Large} \citep{bhattacharjee-etal-2022-banglabert} provides improved performance compared to the base model due to its larger training dataset. It captures more linguistic variations, improves generalization, and enhances the accuracy of text classification in Bengali.\\
\noindent \textbf{(c) \underline{ALBERT based Transformer Models}}: ALBERT (A Lite BERT) \citep{lan2019albert} has shown that superior language models do not necessarily require larger models. It achieves this by utilizing the same encoder segment architecture as the original Transformer but with three crucial modifications: factorized embedding parameters, cross-layer parameter sharing, and employing Sentence-order prediction (SOP) instead of Next Sentence Prediction (NSP). In the context of Bengali language, \textit{sahajBERT}\footnote{\url{https://huggingface.co/neuropark/sahajBERT}} is a collaborative pre-trained ALBERT model that utilizes masked language modeling (MLM) and Sentence Order Prediction (SOP) objectives.

\begin{table}[]
\centering
\caption{Architecture details of the pre-trained transformer models}
\label{tab:transhyper}
\begin{tabular}{ccccc}
\hline
\textbf{Architecture}    & \textbf{Used Model} & \textbf{L} & \textbf{H} & \textbf{P} \\ \hline
\multirow{3}{*}{ELECTRA} & BanglaBERT        & 12         & 12         & 110M       \\
                         & BanglaBERT Generator & 12         & 4          & 34M        \\
                         & BanglaBERT Large      & 24         & 16         & 335M       \\ \hline
BERT                     & BanglaBERT Base      & 12         & 12         & 110M       \\ \hline
ALBERT                   & sahajBERT          & 24         & 16         & 18M       \\ \hline
\end{tabular}
\end{table}

\subsection{Proposed Ensemble Model} \label{sec:ensemble}
Several studies \citep{gutierrez2020ensemble,javed2021fake} have employed ensemble methods to identify fake reviews and explored the usefulness of these techniques in classification tasks. In classification, combining the power of several models in an ensemble highlights the strengths of each model. If one model fails to accurately predict a testing sample, the other models work collectively to compensate for the shortcomings of that model. 
\begin{figure}[ht]
		\centering
            \includegraphics[width=0.8\linewidth]{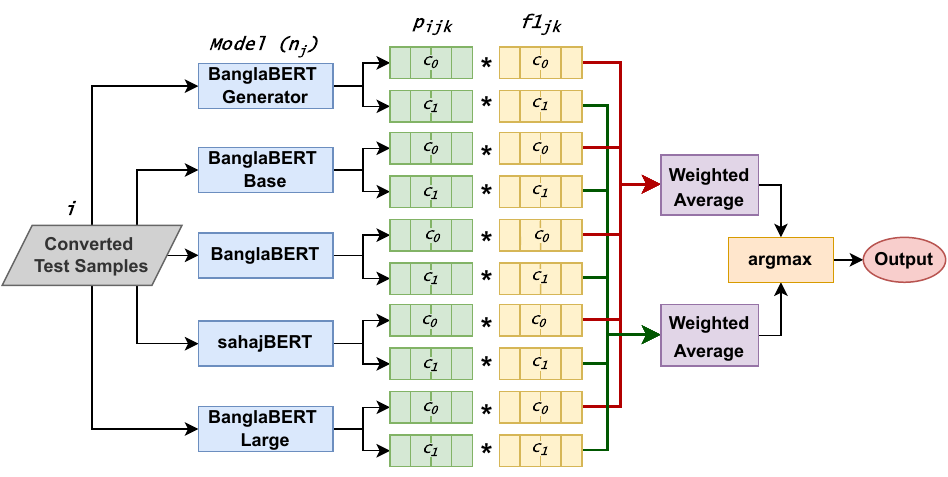}
		\caption{Architecture of the proposed weighted ensemble model for two classes. \textit{$C_{k=0}$} denotes the \textit{fake} class and \textit{$C_{k=1}$} denotes the \textit{Non-fake} class}
		\label{fig:methodensamble}
\end{figure}
In our study, we have considered five pre-trained models that were fine-tuned on the Bengali corpus for the ensemble i.e. \textit{BanglaBERT} \citep{bhattacharjee-etal-2022-banglabert}, \textit{BanglaBERT Generator} \citep{bhattacharjee-etal-2022-banglabert}, \textit{BanglaBERT Large} \citep{bhattacharjee-etal-2022-banglabert}, \textit{BanglaBERT Base} \citep{Sagor_2020}, and \textit{sahajBERT}\footnote{\url{https://huggingface.co/neuropark/sahajBERT}}. By working together, they form an efficient ensemble of transformer models. We have utilized two ensemble methods: average and weighted ensemble. The average ensemble technique \citep{shifath2021transformer, gundapu2021transformer} considers the average of softmax probabilities of all the models and selects the predicted class with the highest score, making the individual predictions relatively less important. On the other hand, the weighted ensemble technique considers the prior predictions of the individual models. Our approach for the weighted ensemble technique differs slightly from the one presented in \citep{sharif2022tackling}. We compared our technique with theirs in Section \ref{compare-ensemble}. In our method, we use the prior f1-scores of each class, and multiply them by the class probabilities (softmax), which serves as an additional weight for the softmax probabilities. Figure \ref{fig:methodensamble} shows a schematic diagram of our proposed ensemble technique for two classes. To give more importance to the prediction of each individual class, we use the individual f1 scores for each class, assuming that the corresponding class is a positive one.

Suppose we have $m$ testing samples and $n$ models in our ensemble approach. Each of the test samples is classified in $c$ classes. The model $n_j$ generates a probability (softmax) for class $c_k$ for each $m_i$ sample which is denoted by $p_{ijk}$. The prior f1-scores of the $c$ classes of $n$ models evaluated on the testing set are $f1_{11}$, $f1_{12}$,...,$f1_{1c}$,$f1_{21}$, $f1_{22}$,...,$f1_{2c}$,... ...,$f1_{n1}$, $f1_{n2}$,...,$f1_{nc}$ (assuming each class as positive). Taking into account all the values shown earlier, Eq. \ref{eqn:ensemble} generates the final result by the proposed ensemble technique for multiple classes.

\begin{equation}
\label{eqn:ensemble}
\resizebox{.9\hsize}{!}{$ O=\operatorname{argmax}\left(\forall_{\mathrm{i} \in(1, \mathrm{~m})}\left( \frac{\sum_{\mathrm{j}=1}^{\mathrm{n}} \mathrm{p}_{\mathrm{ij0}} * \mathrm{f1}_{\mathrm{j0}}}{\sum_{\mathrm{j}=1}^{\mathrm{n}} \mathrm{f1}_{\mathrm{j0}}}, \frac{\sum_{\mathrm{j}=1}^{\mathrm{n}} \mathrm{p}_{\mathrm{ij1}} * \mathrm{f1}_{\mathrm{j1}}}{\sum_{\mathrm{j}=1}^{\mathrm{n}} \mathrm{f1}_{\mathrm{j1}}},... , \frac{\sum_{\mathrm{j}=1}^{\mathrm{n}} \mathrm{p}_{\mathrm{ijk}} * \mathrm{f1}_{\mathrm{jk}}}{\sum_{\mathrm{j}=1}^{\mathrm{n}} \mathrm{f1}_{\mathrm{jk}}}\right)\right)$}
\end{equation}
% \begin{equation}
% \resizebox{.9\hsize}{!}{$ O=\operatorname{argmax}\left(\forall_{\mathrm{i} \in(1, \mathrm{~m})}\left( \frac{\sum_{\mathrm{j}=1}^{\mathrm{n}} \mathrm{p}_{\mathrm{ij}}[\mathrm{0}] * \mathrm{f1}_{\mathrm{j0}}}{\sum_{\mathrm{j}=1}^{\mathrm{n}} \mathrm{f1}_{\mathrm{j0}}}, \frac{\sum_{\mathrm{j}=1}^{\mathrm{n}} \mathrm{p}_{\mathrm{ij}}[\mathrm{1}] * \mathrm{f1}_{\mathrm{j1}}}{\sum_{\mathrm{j}=1}^{\mathrm{n}} \mathrm{f1}_{\mathrm{j1}}},..., \frac{\sum_{\mathrm{j}=1}^{\mathrm{n}} \mathrm{p}_{\mathrm{ij}}[\mathrm{k}] * \mathrm{f1}_{\mathrm{jk}}}{\sum_{\mathrm{j}=1}^{\mathrm{n}} \mathrm{f1}_{\mathrm{jk}}}\right)\right)$}
% \end{equation}
% \begin{equation}
% O=\operatorname{argmax}\left(\forall_{\mathrm{i} \in(1, \mathrm{~m}), \mathrm{k} \in(1,c)} \frac{\sum_{\mathrm{j}=1}^{\mathrm{n}} \mathrm{p}_{\mathrm{ij}}[\mathrm{c}] * \mathrm{f1}_{\mathrm{jk}}}{\sum_{\mathrm{j}=1}^{\mathrm{n}} \mathrm{f1}_{\mathrm{jk}}}\right)
% \end{equation}
Here, $O$ stands for the output vector that contains the prediction of our suggested ensemble technique for $m$ individual samples.

The process of utilizing the weighted ensemble technique to make predictions is outlined in Appendix \ref{secA4}. At first, we multiply the probabilities (softmax) of each class by their corresponding f1 scores and then add them together. We then normalize the individual scores for each class by dividing the sum of the prior f1 scores of the models for that class. Finally, we identify the highest estimated scores across all classes and determine the final prediction.

\section{Experiments and Result Analysis \label{sec6}}
In this section, we present the details of our experiments including the hyper-parameter settings for the detection models. We provide a comprehensive analysis of the results obtained from both independent models and ensemble techniques, including both quantitative and qualitative aspects. Furthermore, we investigate misclassified reviews and categorize the types of errors encountered. Finally, we compare our approach with an existing weighted ensemble technique. Details of the evaluation metrics used in our study can be found in Appendix \ref{secA3}.
\subsection{Experimental Setup}
For experimentation purpose, we utilized CNN, BiLSTM, CNN-BiLSTM, and CNN-BiLSTM with an attention mechanism as deep learning models, along with \textit{BanglaBERT} (BB), \textit{BanglaBERT Generator } (BBG), \textit{BanglaBERT Base} (BBB), \textit{sahajBERT} (SB), and \textit{BanglaBERT Large} (BBL) transformer models. We proposed a weighted ensemble approach which involves combining different combinations of transformer models. We conducted three experiments for both the individual models and ensemble techniques: one with no augmentation, one with augmented fake reviews by \textit{nlpaug}, and one with augmented fake reviews by \textit{bnaug}. We referred to the experiment without any augmentation as \textbf{Approach-1}, while we used the labels \textbf{Approach-2} and \textbf{Approach-3} to describe the experiments that utilized the \textit{nlpaug} and \textit{bnaug} augmentation techniques respectively.

\subsection{Hyper-parameter Settings}
We used Jupyter Notebook as an integrated development environment (IDE) for most of our experiments. Our experiments were carried out using the Asus Dual GeForce RTX 3060 V2 12GB GDDR6 GPU, which was equipped with built-in 16GB RAM and 1TB of storage space. For some of our experiments, we also employed Google Collaboratory with a 13GB back-end GPU and 12GB of local GDDR5 RAM. The Python version we used was 3.6.13 along with numpy (1.19.2) and pandas (1.1.5). Deep learning models were implemented using Tensorflow 2.12.0 and Keras 2.12.0. Scikit-learn 1.2.2 was used to evaluate performance metrics. For the transformer models, we used version 4.28.1 of transformers and implemented them using the PyTorch library. In our experiments, we ensured that each of the train, validation, and test sets were completely distinct. We present the data split in Table \ref{tab:datasplit}. After properly validating the models on the validation set and training on the training data, we evaluated them using samples from the test sets. The hyper-parameters such as batch size, learning rate, and number of epochs utilized in the experiments are listed in Appendix \ref{secA2} through a table. We obtained the optimal hyper-parameters by experimenting within a wide hyper-parameter space. It is worth noting that due to memory constraints, the input sequence length for all models, except \textit{sahajBERT} and \textit{BanglaBERT Large} was limited to 512. However, the input sequence length was 256 for these two models.

\subsection{Results}
\begin{figure}[t]
		\centering
		\begin{subfigure}[b]{0.49\columnwidth}
		    \centering
			\includegraphics[width=1\linewidth]{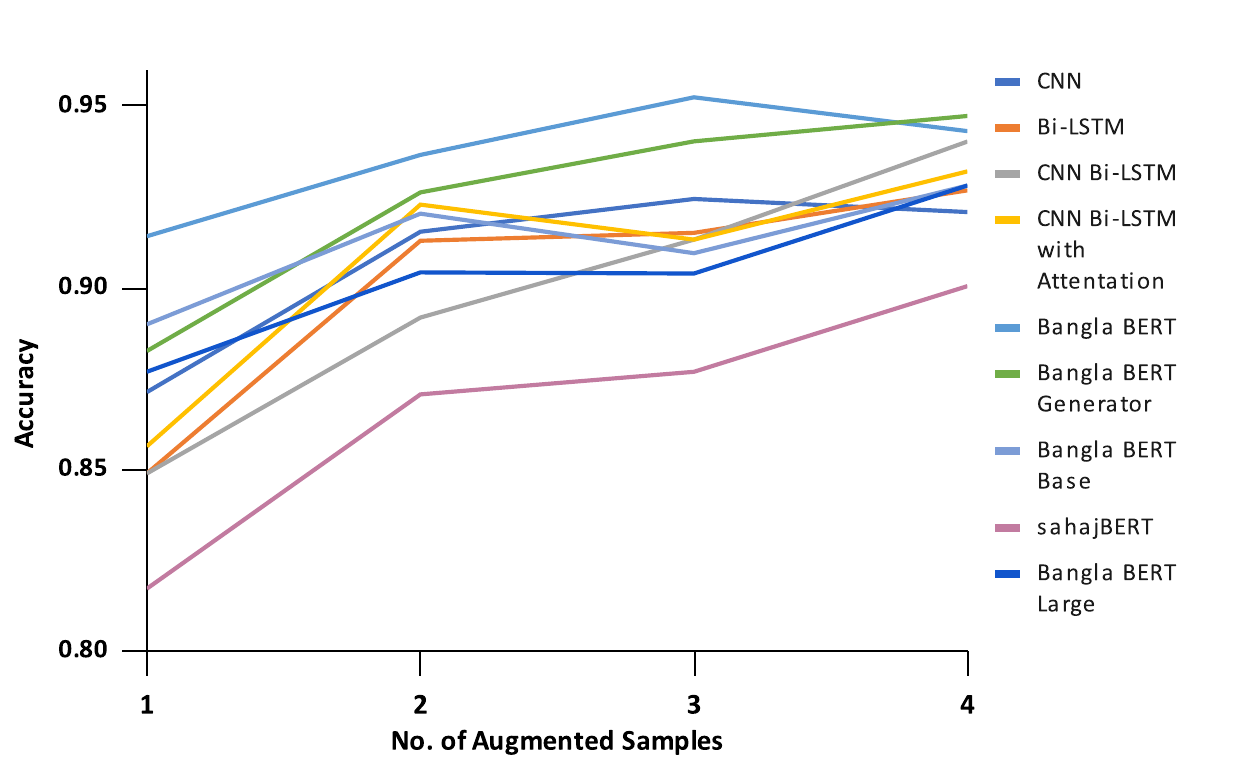}
                \subcaption{Approach - 2 (nlpaug)}
			\label{fig:chart_our_approach}
		\end{subfigure}
		\begin{subfigure}[b]{0.49\columnwidth}
		    \centering
			\includegraphics[width=1\linewidth]{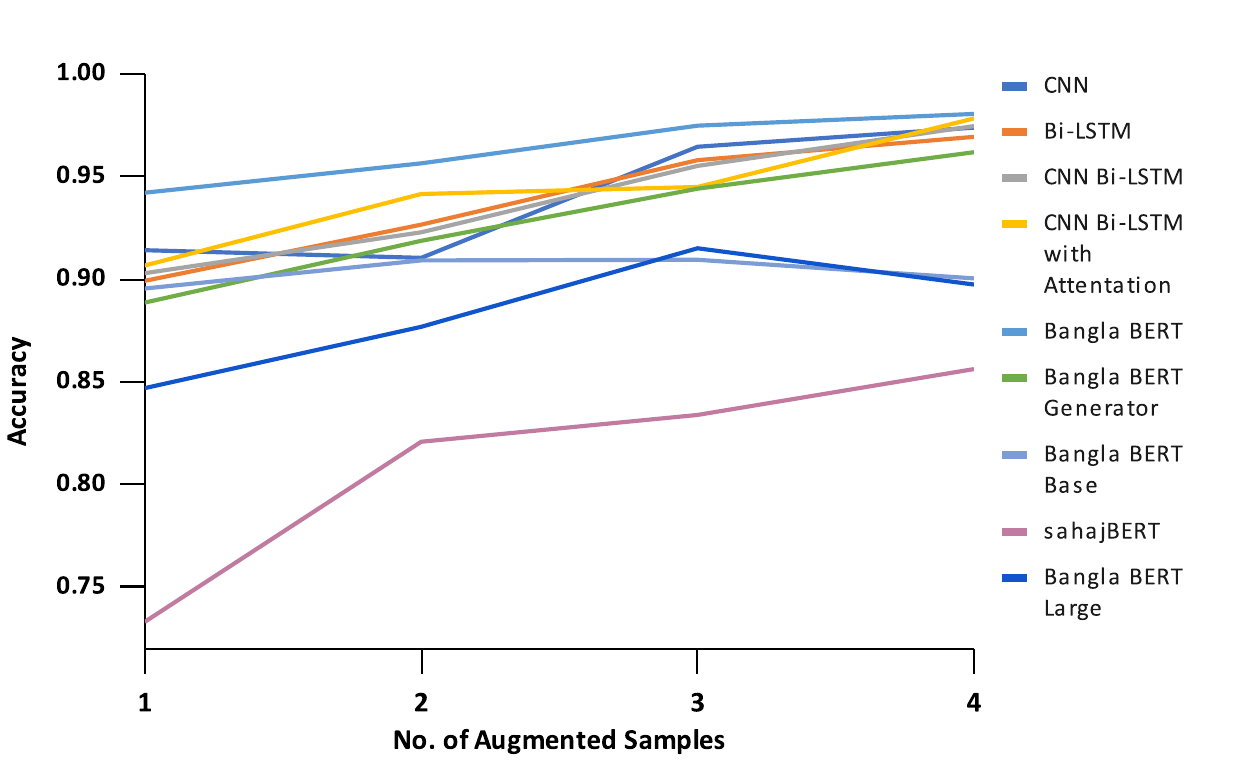}
                \subcaption{Approach - 3 (bnaug)}
			\label{fig:chart_nlpaug}
		\end{subfigure}
		\caption{Performance visualization of different models with increasing number of augmented fake review instances}
		\label{fig:chart_acc_vs_aug}
\end{figure}

To start our analysis, we first present the experimental results of the deep learning and transformer models. The results of Approach-1 for the individual models are shown in Tables \ref{tab:tradnoaug} and a comparative analysis of  Approach-2, and Approach-3 for four augmented samples along with the original fake data through a bar chart can be seen in Fig. \ref{fig:compchart}. The detailed result of Approach-2 and Approach-3 is shown in Appendix \ref{secA1} through Tables \ref{tab:tradnlpaug} and \ref{tab:tradsagar}. For ensemble techniques, four augmented fake data per review along with the actual one are considered for Approach-2 and Approach-3, and their results are shown in Tables \ref{tab:ensemblenlpaug} and \ref{tab:ensemblesagar}. On the other hand, Table \ref{tab:ensemblenoaug} shows the results of all combinations of Approach-1. Appendix \ref{secA1} provides the results using $0$ to $3$ augmented fake data per review in Tables \ref{tab:appensemblenlpaug} and \ref{tab:appensemblesagar} for Approach-2 and Approach-3 respectively. The two ensemble approaches (average and weighted) comprise of six combinations (5 for four individual and 1 for five individual transformer models) out of all possible combinations of the five distinct transformer models. Gradual performance improvement can be observed for individual models on both Approach-2 and Approach-3 as shown in Figure \ref{fig:chart_acc_vs_aug}. These charts demonstrate the importance of augmentation in our imbalanced dataset.

\subsection{Quantitative Analysis}
This section aims to empirically justify the performance of the models and is divided into two parts. The first part discusses the deep learning models and transformers, while the second part covers the ensemble techniques.

\subsubsection{Deep Learning and Transformer Models}
This section is divided into three parts taking into account the use of data augmentation. Firstly, we present the results without any data augmentation. Then, we discuss the results obtained with the addition of two data augmentation techniques aiming to improve the performance of the models.

\noindent\textbf{• \underline{Approach-1:}}
% \subsubsection{Performance (Without any augmentation)}
An equal number (1339) of data from each label are utilized when there is no augmentation. Due to the unequal distribution of data between fake and non-fake classes, only 1339 of the 7710 non-fake data are randomly taken into account while preserving a 1:1 ratio with the fake data. Table \ref{tab:trial} illustrates five individual trials that randomly selected 1339 samples from the 7710 non-fake data along with 1339 fake samples for the Bangla BERT model. Random sampling of non-fake data yields identical results in all the cases, varying only by 0.026\% in terms of weighted f1 score with a mean of \textbf{0.791} and a standard deviation of \textbf{0.01157} which is quite acceptable.

% \usepackage{multirow}
% \usepackage{multirow}
\begin{table}[H]
\caption{Performance comparison of five individual trials with a random sampling of non-fake reviews for \textit{Bangla BERT} model without any augmentation (Approach - 1)}
\label{tab:trial}
\begin{tabular}{c|ccc|ccc|c|c|c}
\hline
\multirow{2}{*}{\textbf{\begin{tabular}[c]{@{}c@{}}Trial \\ No.\end{tabular}}} & \multicolumn{3}{c|}{\textbf{Fake}}                                              & \multicolumn{3}{c|}{\textbf{Non-Fake}}                                          & \multirow{2}{*}{\textbf{WF1}} & \multirow{2}{*}{\textbf{\begin{tabular}[c]{@{}c@{}}ROC-\\ AUC\end{tabular}}} & \multirow{2}{*}{\textbf{MCC}} \\ \cline{2-7} & \multicolumn{1}{c}{\textbf{P}} & \multicolumn{1}{c}{\textbf{R}} & \textbf{F1} & \multicolumn{1}{c}{\textbf{P}} & \multicolumn{1}{c}{\textbf{R}} & \textbf{F1} &                               &                                                                              &                               \\ \hline
1                                                                              & \multicolumn{1}{c}{0.886}      & \multicolumn{1}{c}{0.694}      & 0.778       & \multicolumn{1}{c}{0.748}      & \multicolumn{1}{c}{0.910}      & 0.822       & 0.800                         & 0.802                                                                        & 0.619                         \\
2                                                                              & \multicolumn{1}{c}{0.841}      & \multicolumn{1}{c}{0.709}      & 0.769       & \multicolumn{1}{c}{0.748}      & \multicolumn{1}{c}{0.866}      & 0.803       & 0.786                         & 0.787                                                                        & 0.582                         \\
3                                                                              & \multicolumn{1}{c}{0.811}      & \multicolumn{1}{c}{0.799}      & 0.805       & \multicolumn{1}{c}{0.801}      & \multicolumn{1}{c}{0.813}      & 0.807       & \textbf{0.806}                         & 0.806                                                                        & 0.612                         \\
4                                                                              & \multicolumn{1}{c}{0.838}      & \multicolumn{1}{c}{0.694}      & 0.759       & \multicolumn{1}{c}{0.739}      & \multicolumn{1}{c}{0.866}      & 0.797       & 0.778                         & 0.780                                                                        & 0.568                         \\ 
5                                                                              & \multicolumn{1}{c}{0.867}      & \multicolumn{1}{c}{0.679}      & 0.762       & \multicolumn{1}{c}{0.736}      & \multicolumn{1}{c}{0.896}      & 0.808       & 0.785                         & 0.787                                                                        & 0.589                         \\ \hline
\end{tabular}
\end{table}

In this experiment, deep learning techniques like CNN, BiLSTM, and their combinations performed remarkably well. The performance of the models on the small quantity of data is almost identical for CNN, BiLSTM, and CNN BiLSTM with attention layer, each with a weighted-f1 (WF1) of 0.757, 0.769, and 0.761 which can be seen at Table \ref{tab:tradnoaug}. The model's ability to predict the fake review with a very little number of training data is further supported by the ROC-AUC and MCC scores. With a low WF1 of 0.694 and MCC of 0.388, the hybrid CNN-BiLSTM model is marginally under performing in the absence of augmentation. Transformers play a better role in this case. Among all the models, \textit{BanglaBERT} (BB) has produced the best results, with a promising WF1 and MCC of 0.809 and 0.621 respectively. \textit{BanglaBERT Generator } (BBG) and \textit{BanglaBERT Base} (BBB) exhibit comparable performances, with WF1 scores of 0.809 and 0.806 respectively. \textit{BanglaBERT Large} (BBL), which is trained with approximately half as many tokens than other models, achieves excellent results with WF1 of 0.787 and MCC of 0.577. As can be observed in Table \ref{tab:tradnoaug} where the recall value of the fake class is lower than the precision value and the converse for the non-fake class, it should be noted that BBL is slightly biased to the fake class as it predicts the non-fake classes as fake in more situations than it predicts the fake classes as non-fake. With an MCC score of just 0.412, \textit{sahajBERT} (SB) is the transformer that performs the worst. 
\begin{table}[]
\caption{Performance comparison among individual models without any augmentation (Approach - 1)}
\label{tab:tradnoaug}
\begin{tabular}{c|ccc|ccc|ccc}

\hline
                        & \multicolumn{3}{c|}{\textbf{Fake}} & \multicolumn{3}{c|}{\textbf{Non-Fake}}   &       &          &       \\ \hline
                        
\textbf{Model}   & \multicolumn{1}{c}{\textbf{P}} & \multicolumn{1}{c}{\textbf{R}} & \textbf{F1} & \multicolumn{1}{c}{\textbf{P}} & \multicolumn{1}{c}{\textbf{R}} & \textbf{F1} & \multicolumn{1}{c}{\begin{tabular}[c]{@{}c@{}}{\rotatebox[origin=c]{90}{\textbf{ WF1 }}}\end{tabular}}  & \multicolumn{1}{c}{\begin{tabular}[c]{@{}c@{}}{\rotatebox[origin=c]{90}{\textbf{ ROC-AUC }}}\end{tabular}} & \multicolumn{1}{c}{\begin{tabular}[c]{@{}c@{}}{\rotatebox[origin=c]{90}{\textbf{ MCC }}}\end{tabular}} \\ \hline

CNN                     & \multicolumn{1}{c}{0.759} & \multicolumn{1}{c}{0.754} & 0.757 & \multicolumn{1}{c}{0.756} & \multicolumn{1}{c}{0.761} & 0.758 & \textbf{0.757}  & 0.823   & 0.515 \\ \hline
BiLSTM                 & \multicolumn{1}{c}{0.765} & \multicolumn{1}{c}{0.776} & 0.770 & \multicolumn{1}{c}{0.773} & \multicolumn{1}{c}{0.761} & 0.767 & \textbf{0.769}  & 0.833   & 0.537 \\ \hline
{\begin{tabular}[c]{@{}c@{}}CNN   \\ BiLSTM\end{tabular}}             & \multicolumn{1}{c}{0.694} & \multicolumn{1}{c}{0.694} & 0.694 & \multicolumn{1}{c}{0.694} & \multicolumn{1}{c}{0.694} & 0.694 & \textbf{0.694} & 0.794   & \textbf{0.388} \\ \hline
{\begin{tabular}[c]{@{}c@{}}CNN\\ BiLSTM\\ with\\ Attention\end{tabular}} & \multicolumn{1}{c}{0.769} & \multicolumn{1}{c}{0.746} & 0.758 & \multicolumn{1}{c}{0.754} & \multicolumn{1}{c}{0.776} & 0.765 & \textbf{0.761}  & 0.819   & 0.523 \\ \hline
{\begin{tabular}[c]{@{}c@{}}Bangla\\BERT (BB)\end{tabular}}                    & \multicolumn{1}{c}{0.832} & \multicolumn{1}{c}{0.776} & 0.803 & \multicolumn{1}{c}{0.790} & \multicolumn{1}{c}{0.843} & 0.816 & \textbf{0.809}  & 0.810   & \textbf{0.621} \\ \hline
{\begin{tabular}[c]{@{}c@{}}Bangla\\BERT\\Generator\\(BBG)\end{tabular}}          & \multicolumn{1}{c}{0.806} & \multicolumn{1}{c}{0.813} & 0.809 & \multicolumn{1}{c}{0.811} & \multicolumn{1}{c}{0.805} & 0.808 & \textbf{0.809} & 0.809   & 0.617 \\ \hline
{\begin{tabular}[c]{@{}c@{}}Bangla\\BERT\\Base (BBB)\end{tabular}}                   & \multicolumn{1}{c}{0.811} & \multicolumn{1}{c}{0.799} & 0.805 & \multicolumn{1}{c}{0.801} & \multicolumn{1}{c}{0.813} & 0.807 & \textbf{0.806} & 0.806   & 0.612 \\ \hline
{\begin{tabular}[c]{@{}c@{}}sahaj\\BERT (SB)\end{tabular}}               & \multicolumn{1}{c}{0.755} & \multicolumn{1}{c}{0.597} & 0.667 & \multicolumn{1}{c}{0.667} & \multicolumn{1}{c}{0.806} & 0.730 & 0.698 & 0.701   & \textbf{0.412} \\ \hline
{\begin{tabular}[c]{@{}c@{}}Bangla\\BERT\\Large (BBL)\end{tabular}}              & \multicolumn{1}{c}{0.813} & \multicolumn{1}{c}{0.746} & 0.778 & \multicolumn{1}{c}{0.766} & \multicolumn{1}{c}{0.828} & 0.796 & \textbf{0.787} & 0.787   & \textbf{0.57}7 \\ \hline
\end{tabular}
\end{table}

\begin{figure}[ht]
		\centering
            \includegraphics[width=1\linewidth]{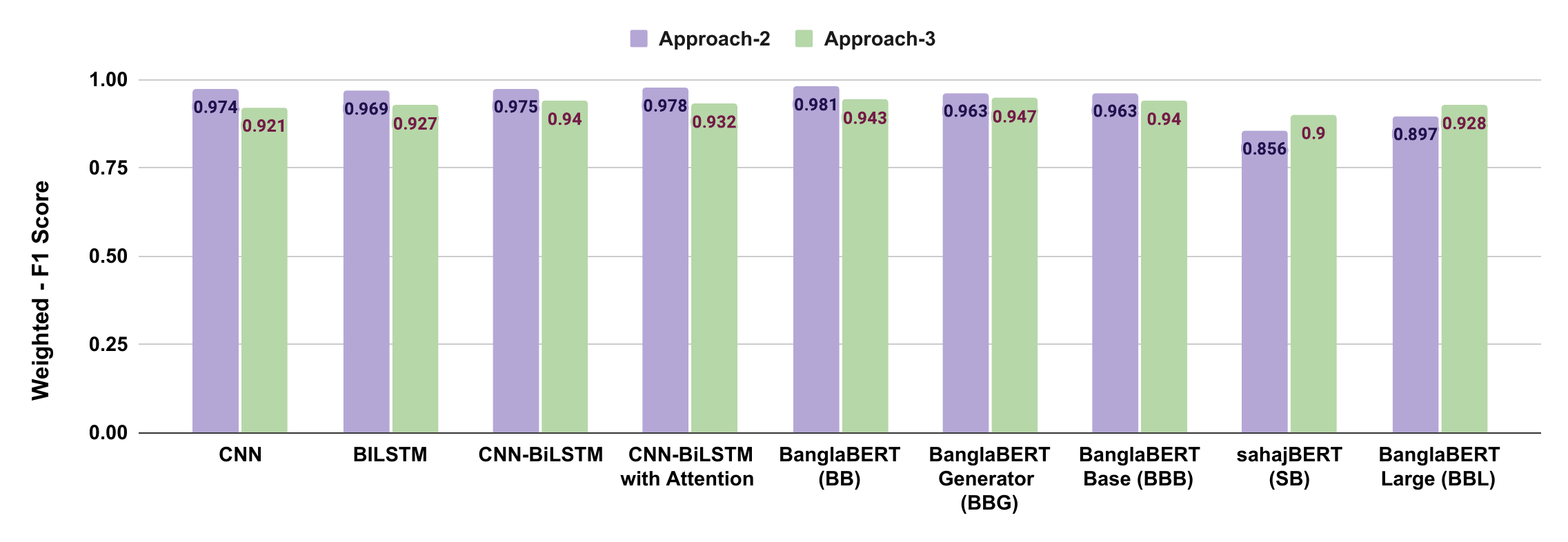}
		\caption{Comparison of weighted-F1 scores of different models for four augmented samples along with the original fake data using \textit{nlpaug} (Approach-2) and \textit{bnaug} (Approach-3) augmentation techniques}
		\label{fig:compchart}
\end{figure}

\noindent\textbf{• \underline{Approach-2 \& Approach-3:}} The weighted-F1 scores of distinct individual models are shown in Fig.\ref{fig:compchart} for approaches 2 and 3 in terms of four augmented samples along with the original fake review. The chart clearly shows that models using the \textit{bnaug} augmentation strategy performed lower than those using \textit{nlpaug}. The overall performance of approach-3 has decreased by 2\% to 4\% compared to models of approach-2. Despite the short sequence length, only SB and BBL perform well; their respective WF1 values are 4.4\% and 3.1\% higher than those of previous studies using \textit{nlpaug}. In approach-2, the CNN and BiLSTM hybrid models outperform the standalone models. The model developed with CNN-BiLSTM and an attention layer achieved the best WF1 (0.978) for four separate augmentations. In the presence of augmentation, transformers produce results that are similar to those of DL techniques. BanglaBERT (BB) performs best in terms of WF1 score (0.981) for four distinct augmentations. In approach-3, the hybrid model with CNN and BiLSTM outperformed the separate models somewhat. Similar to other experiments using nlpaug, BB and BBG here yield the best results. BBG and BB achieved the highest WF1 of 0.947 and 0.943. The detailed results of all the individual models for approach-2 and 3 can be seen in Appendix \ref{secA1},

\subsubsection{Ensemble Approaches}
The ensemble technique is utilized to overcome the limitations of individual models, and in this paper, we employ two ensemble approaches: average and weighted. To address data augmentation, this section is also divided into three segments. Each instance is categorized into two separate groups using the two ensemble techniques mentioned earlier. In the weighted approach, the f1 scores of each class for each model are considered to maintain the significance of each class separately, as described in section \ref{sec:ensemble}. The two ensemble approaches (average and weighted) comprise of six combinations (5 for four individual and 1 for five individual transformer models) out of all possible combinations of the five distinct transformer models. We refer the combinations as: (a) \textbf{EN1}: BBG+BB+BBL+BBB, (b) \textbf{EN2}: BB+BBL+BBB+SB, (c) \textbf{EN3}: BBG+BB+BBL+SB, (d) \textbf{EN4}: BBG+BB+BBB+SB, (e) \textbf{EN5}: BBG+BBL+BBB+SB and (f) \textbf{EN6}: BBG+BB+SB+BBL+BBB.

\noindent \textbf{• \underline{Approach-1:}}
% \subsubsection{Performance (Without any augmentation)}
In case of ensemble approaches (average and weighted), all the underlying ensemble models performed better than the individual model which can be seen in Table \ref{tab:tradnoaug}. The lowest WF1 for the ensemble is 0.8208 which is achieved by EN2 in both techniques. The best performance is shown by the combination of EN1 in the average case which is 0.8394 but dropped by almost 4\% in the weighted case. The result of EN6 which is the combination of all the transformers shows the highest performance in the weighted case without augmentation.

\begin{table}[!ht]
\caption{Performance comparison between two ensemble approaches (average and weighted) without any augmentation (Approach - 1)}
\label{tab:ensemblenoaug}
%\resizebox{\textwidth}{!}{
\begin{tabular}{cc|ccc|ccc|cccc}
\hline
       & \multicolumn{1}{c|}{}   & \multicolumn{3}{c|}{\textbf{Fake}}                   & \multicolumn{3}{c|}{\textbf{Non-Fake}}                & \multicolumn{4}{c}{}                           \\ \hline
     & \multicolumn{1}{c|}{\textbf{Methods}}  & \multicolumn{1}{c}{\textbf{P}}    & \multicolumn{1}{l}{\textbf{R}}    & \multicolumn{1}{l|}{\textbf{F1}}   & \multicolumn{1}{l}{\textbf{P}}    & \multicolumn{1}{l}{\textbf{R}}    & \multicolumn{1}{l|}{\textbf{F1}}   & \multicolumn{1}{l}{\textbf{WF1}} & \multicolumn{1}{l}{\textbf{ACC}}  & \multicolumn{1}{l}{\begin{tabular}[c]{@{}c@{}}\textbf{ROC}\\ \textbf{-AUC}\end{tabular}} & \textbf{MCC}  \\ \hline
\multirow{6}{*}{\rotatebox[origin=c]{90}{AVERAGE}} & EN1  & 0.86       & 0.81       & 0.84        & 0.82       & 0.87       & 0.84        &  \textbf{0.8394 }             &0.84    & 0.84                  & 0.68\\  
        & EN2     & 0.84       & 0.80       & 0.82        & 0.81       & 0.84       & 0.82        & \textbf{0.8208}       &0.82        & 0.82                  & 0.64\\  
        & EN3     & 0.83       & 0.81       & 0.82        & 0.82       & 0.84       & 0.83        & 0.8246      &0.82      & 0.82                  & 0.65\\  
        & EN4     & 0.85       & 0.81       & 0.83        & 0.82       & 0.86       & 0.84        & 0.8357    &0.84      & 0.84                  & 0.67\\  
        & EN5    & 0.84       & 0.81       & 0.82        & 0.81       & 0.85       & 0.83        & 0.8283     &0.83     & 0.83                  & 0.66\\  
        & EN6 & 0.85       & 0.81       & 0.83        & 0.82       & 0.86       & 0.84        & 0.8357        &0.84          & 0.84                  & 0.67\\ \hline 
% \textbf{}    & \multicolumn{10}{c}{}                      \\ \hline
\multirow{6}{*}{\rotatebox[origin=c]{90}{ WEIGHTED}}   & EN1  & 0.82       & 0.77       & 0.79        & 0.78       & 0.83       & 0.80        & 0.7983           &0.80      & 0.80                  & 0.60         \\  
        & EN2          & 0.83       & 0.80       & 0.81        & 0.81       & 0.84       & 0.82        & 0.8171    &0.82      & 0.82                  & 0.63\\  
        & EN3         & 0.83       & 0.81       & 0.82        & 0.82       & 0.83       & 0.82        & 0.8209     &0.82           & 0.82                  & 0.64\\  
        & EN4          & 0.85       & 0.81       & 0.83        & 0.82       & 0.86       & 0.84        & 0.8357        &0.84      & 0.84                  & 0.67\\  
        & EN5         & 0.84       & 0.81       & 0.82        & 0.81       & 0.84       & 0.83        & 0.8246    &0.82        & 0.82                  & 0.65\\  
        & EN6 & 0.85       & 0.82       & 0.83        & 0.83       & 0.85       & 0.84        & 0.8358     &0.84     & 0.84                  & 0.67\\ \hline
\end{tabular}
%}
\end{table}

\begin{figure}
     \centering
     \begin{subfigure}[b]{0.45\textwidth}
         \centering
         \includegraphics[width=1\textwidth]{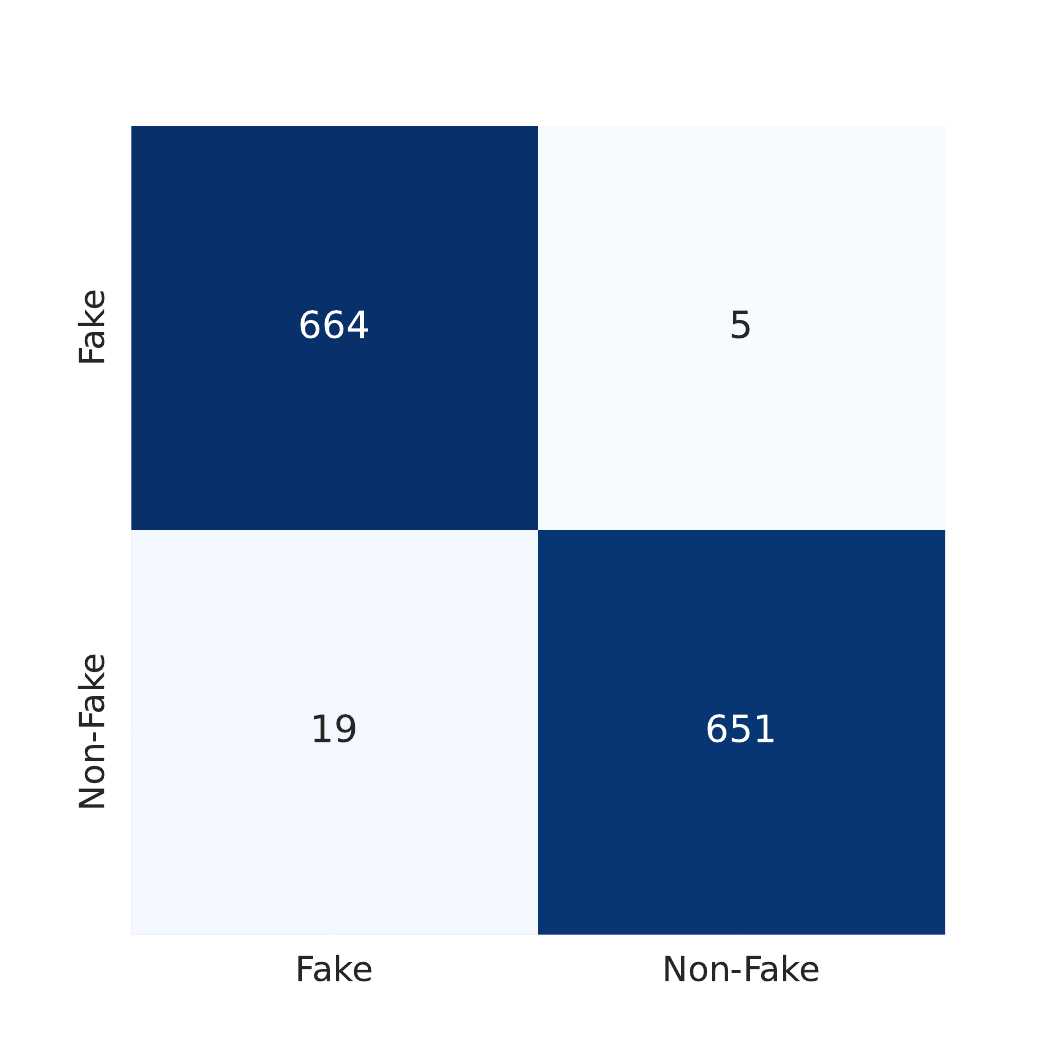}
         \caption{Approach - 2 (EN1)}
         \label{fig:app2confusion}
     \end{subfigure}
     \hfill
     \begin{subfigure}[b]{0.45\textwidth}
         \centering
         \includegraphics[width=1\textwidth]{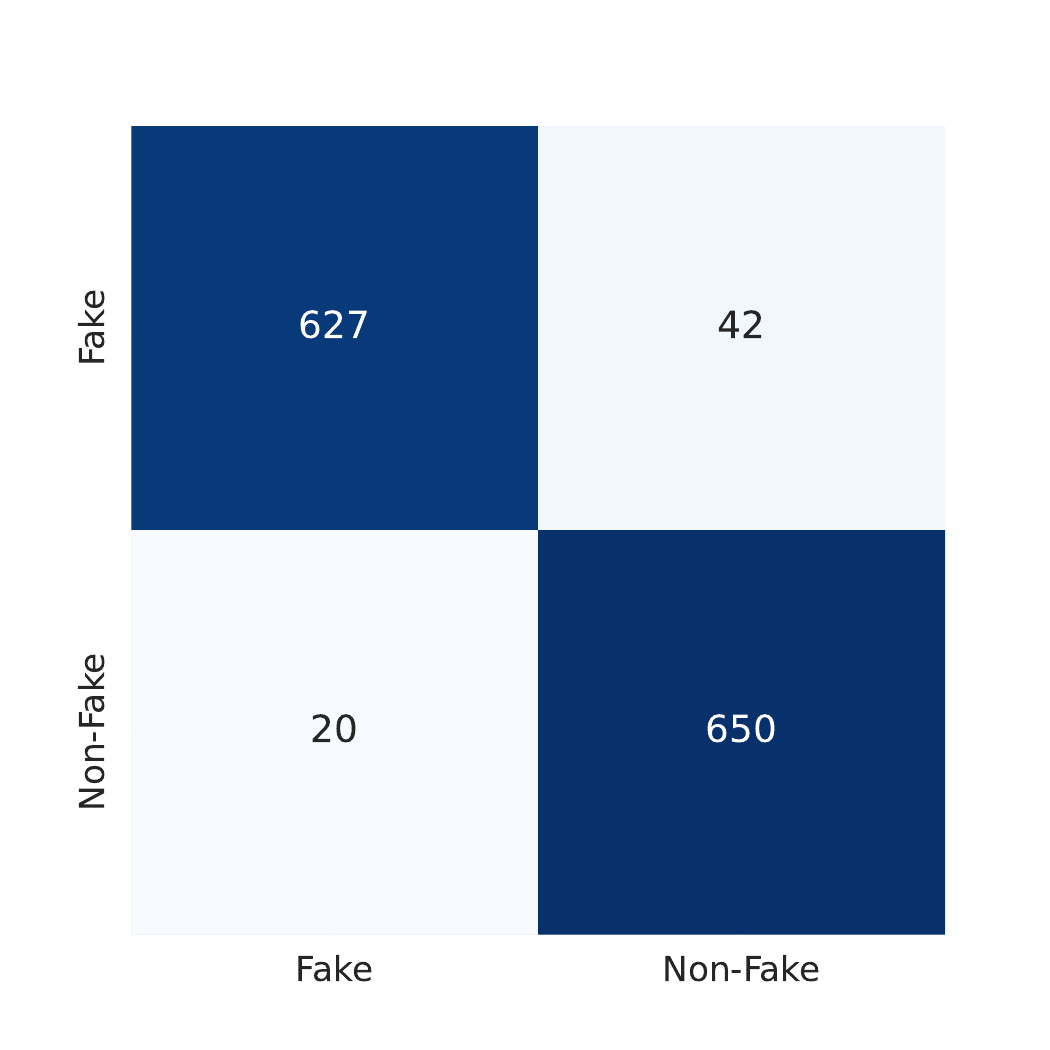}
         \caption{Approach - 3 (EN1)}
         \label{fig:app3confusion}
     \end{subfigure}
        \caption{Confusion matrix of the proposed weighted ensemble model for approach - 2 and approach - 3}
        \label{fig:confusion}
\end{figure}

\noindent \textbf{• \underline{Approach-2:}}
% \subsubsection{Performance (Augmentation with nlpaug)}
% As the lowest WF1 with one augmentation per review is 0.91, the precision attained by the ensemble methods is extremely apparent. 
With the increase of the augmented samples in each of the combinations, Table \ref{tab:ensemblenlpaug} clearly shows the constant escalation of the performance. Since only combinations of four and five models have been considered, there is a little difference between the average and weighted ensemble models. Therefore, if one model can not accurately predict some scenarios, another model can correct the incorrect predictions and offer a strong performance when combined. Since \textit{BanglaBERT} is the best performing individual model, as seen in Table \ref{tab:tradnlpaug}, its dominance is clearly visible. EN5, a combination without BB, does not produce the same outcomes as other combinations with BB, which are roughly 1\% less effective in each case of augmentation. On the other hand, \textit{sahajBERT} (SB) is the lowest-performing individual model which also affects the result here. We can see that EN1 which is an ensemble of transformers without SB is the highest-performing model in both the average and weighted cases. Figure \ref{fig:app2confusion} shows that EN1 ensemble model classifies 19 non-fake classes as fake due to high similarities between fake and non-fake instances. 

\begin{table}[!h]
\caption{Performance comparison between two ensemble approaches (average and weighted) on four augmentations per actual fake review generated using \textit{nlpaug} (Approach - 2)}
\label{tab:ensemblenlpaug}
%\resizebox{\textwidth}{!}{
\begin{tabular}{cc|ccc|ccc|cccc}
\hline
     & \multicolumn{1}{c|}{}   & \multicolumn{3}{c|}{\textbf{Fake}}                   & \multicolumn{3}{c|}{\textbf{Non-Fake}}                & \multicolumn{4}{c}{}                           \\ \hline
     & \multicolumn{1}{c|}{\textbf{Methods}}  & \multicolumn{1}{c}{\textbf{P}}    & \multicolumn{1}{l}{\textbf{R}}    & \multicolumn{1}{l|}{\textbf{F1}}   & \multicolumn{1}{l}{\textbf{P}}    & \multicolumn{1}{l}{\textbf{R}}    & \multicolumn{1}{l|}{\textbf{F1}}   & \multicolumn{1}{l}{\textbf{WF1}} & \multicolumn{1}{l}{\textbf{ACC}} & \multicolumn{1}{l}{\begin{tabular}[c]{@{}c@{}}\textbf{ROC}\\ \textbf{-AUC}\end{tabular}} & \textbf{MCC}  \\ \hline
 \multirow{6}{*}{\rotatebox[origin=c]{90}{AVERAGE}}      &  EN1         & 0.98 & 0.99 & 0.98 & 0.98 & 0.98 & 0.98 & 0.9808  & 0.98 & 0.98         & 0.97 \\ 

    & EN2 & 0.98 & 0.98 & 0.98 & 0.98 & 0.98 & 0.98 & 0.9805 & 0.98 & 0.98         & 0.96 \\ 

   & EN3 & 0.98 & 0.97 & 0.97 & 0.97 & 0.98 & 0.97 & 0.9731 & 0.97  & 0.97         & 0.95 \\  

    & EN4 & 0.97 & 0.99 & 0.98 & 0.99 & 0.97 & 0.98 & 0.9768 & 0.98 & 0.98         & 0.95 \\ 
  
   & EN5 & 0.97 & 0.97 & 0.97 & 0.97 & 0.97 & 0.97 & \textbf{0.9686} & 0.97 & 0.97         & 0.94 \\ 
  
   & EN6 & 0.97 & 0.98 & 0.98 & 0.98 & 0.97 & 0.98 & 0.9768 & 0.98  & 0.98         & 0.95 \\  \hline

\multirow{6}{*}{\rotatebox[origin=c]{90}{WEIGHTED}} & EN1 &  0.97 & 0.99 & 0.98 & 0.99 & 0.97 & 0.98 & \textbf{0.9843} & 0.98 & 0.98         & 0.96 \\
 & EN2 &  0.98 & 0.98 & 0.98 & 0.98 & 0.98 & 0.98 & 0.9805 & 0.98  & 0.98         & 0.96 \\ 
 & EN3 &  0.98 & 0.99 & 0.98 & 0.98 & 0.98 & 0.98 & 0.9813 & 0.98 & 0.98         & 0.96 \\ 
  & EN4 &  0.97 & 0.99 & 0.98 & 0.99 & 0.97 & 0.98 & 0.9798 & 0.98 & 0.98         & 0.96 \\ 
  & EN5 &  0.97 & 0.98 & 0.98 & 0.98 & 0.97 & 0.98 & 0.9775 & 0.98 & 0.98         & 0.96 \\ 
  & EN6 &  0.97 & 0.99 & 0.98 & 0.99 & 0.97 & 0.98 & 0.9805 & 0.98 & 0.98         & 0.96 \\ \hline
\end{tabular}
%}
\end{table}

\noindent \textbf{• \underline{Approach-3:}}
% \subsubsection{Performance (Augmentation with bnaug)}
Due to the variance in generating augmented samples, ensemble approaches exhibit almost 3\% lower performance when augmented samples by \textit{bnaug} are used instead of \textit{nlpaug}. The highest performing model is again EN1 for both the average and weighted ensemble having the same WF1 score of 0.9558 illustrated in Table \ref{tab:ensemblesagar}. Figure \ref{fig:app3confusion} shows EN1 ensemble model classifies 42 fake classes as non-fake due to high similarities between the fake and non-fake instances. That is why the recall is lower than the precision considering the fake class as positive. The maximum MCC score of 0.91 obtained by these two cases supports the claim. 

\subsubsection{Summary of Quantitative Analysis}
Summarizing the findings, we observe that among the stand-alone deep learning models, CNN BiLSTM with attention layer has the highest WF1 score for approach-2 and CNN BiLSTM has the highest WF1 score for approach-3. Again \textit{Bangla BERT} and \textit{Bangla BERT Generator} are the best-performing transformer model for approach-2 and approach-3 respectively in terms of WF1 score. The proposed weighted ensemble model produced a 0.9843 weighted F1-score on 13390 reviews, of which 6695 were fake (1339 were genuine fakes, while the remaining 5356 were augmented using \textit{nlpaug} and 6695 were non-fake (randomly chosen from 7710 cases). The same ensemble model was also used to generate a 0.9558 weighted F1-score when \textit{bnaug} was used to augment fake reviews.

\begin{table}[!ht]
\caption{Performance comparison between two ensemble approaches (average and weighted) on four augmentations per actual fake review generated using \textit{bnaug} (Approach - 3)}
\label{tab:ensemblesagar}
%\resizebox{\textwidth}{!}{
\begin{tabular}{cc|ccc|ccc|cccc}
\hline
     & \multicolumn{1}{c|}{}   & \multicolumn{3}{c|}{\textbf{Fake}}                   & \multicolumn{3}{c|}{\textbf{Non-Fake}}                & \multicolumn{4}{c}{}                           \\ \hline
     & \multicolumn{1}{c|}{\textbf{Methods}}  & \multicolumn{1}{c}{\textbf{P}}    & \multicolumn{1}{l}{\textbf{R}}    & \multicolumn{1}{l|}{\textbf{F1}}   & \multicolumn{1}{l}{\textbf{P}}    & \multicolumn{1}{l}{\textbf{R}}    & \multicolumn{1}{l|}{\textbf{F1}}   & \multicolumn{1}{l}{\textbf{WF1}} & \multicolumn{1}{l}{\textbf{ACC}}  & \multicolumn{1}{l}{\begin{tabular}[c]{@{}c@{}}\textbf{ROC}\\ \textbf{-AUC}\end{tabular}} & \textbf{MCC}  \\ \hline
\multirow{6}{*}{\rotatebox[origin=c]{90}{AVERAGE}}      & EN1 &  0.97 & 0.94 & 0.95 & 0.94 & 0.97 & 0.95 & \textbf{0.9558}  & 0.95 & 0.95         & \textbf{0.91} \\  
   & EN2 & 0.97 & 0.93 & 0.95 & 0.93 & 0.97 & 0.95 & 0.9491 & 0.95 & 0.95         & 0.90 \\ 
   & EN3 &  0.97 & 0.93 & 0.95 & 0.93 & 0.97 & 0.95 & 0.9483 & 0.95 & 0.95         & 0.90 \\  
   & EN4 &  0.96 & 0.95 & 0.96 & 0.95 & 0.97 & 0.96 & 0.9528 & 0.96 & 0.96         & 0.91 \\ 
   & EN5 &  0.97 & 0.93 & 0.95 & 0.93 & 0.97 & 0.95 & 0.9491 & 0.95 & 0.95         & 0.90 \\  
   & EN6 &  0.96 & 0.95 & 0.96 & 0.95 & 0.97 & 0.96 & 0.9528 & 0.96 & 0.96         & 0.91 \\ \hline

\multirow{6}{*}{\rotatebox[origin=c]{90}{WEIGHTED}} & EN1 &  0.97 & 0.94 & 0.95 & 0.94 & 0.97 & 0.95 & \textbf{0.9558} & 0.95 & 0.95         & \textbf{0.91} \\ 
   & EN2 &  0.97 & 0.93 & 0.95 & 0.93 & 0.97 & 0.95 & 0.9483 & 0.95 & 0.95         & 0.90 \\ 
   & EN3 &  0.97 & 0.93 & 0.95 & 0.93 & 0.97 & 0.95 & 0.9498 & 0.95 & 0.95         & 0.90 \\  
   & EN4 &  0.96 & 0.94 & 0.95 & 0.94 & 0.97 & 0.95 & 0.9543 & 0.95 & 0.95         & 0.91 \\ 
   & EN5 &  0.98 & 0.92 & 0.95 & 0.93 & 0.98 & 0.95 & 0.9528 & 0.95  & 0.95         & 0.91 \\  
   & EN6 &  0.97 & 0.93 & 0.95 & 0.94 & 0.97 & 0.95 & 0.9528 & 0.95 & 0.95         & 0.91 \\ \hline
\end{tabular}
%}
\end{table}

\subsection{Qualitative Analysis}
In this section, the quality of the models is evaluated in light of generalization performance on actual test samples. For this purpose, we used our proposed model which is the combination of all the transformers except \textit{sahajBERT} (EN1) with both augmentation techniques. In addition to the initial samples for the fake class, four augmentation samples are used to train the models that are being examined here. Table \ref{tab:ensemblenlpaug} and \ref{tab:ensemblesagar} both demonstrate that EN1 is the best-performing model. Table \ref{tab:samples1} contains two samples from the test dataset; Two more samples\footnote{\url{https://www.facebook.com/groups/foodbankbd/posts/6375361032571128/}}'\footnote{\url{https://www.facebook.com/groups/foodbankbd/posts/6382876588486239/}} can be found in Appendix \ref{secA5} which come from recent posts on social media. The first one from Table \ref{tab:samples1} is a negative review which is asking questions and is somewhat deceptive for the model trained with \textit{bnaug} (approach-3), Approach-2 correctly determined it to be non-fake, but approach-3 could not. The reverse situation occurs in the second example, where approach-2 fails to accurately identify it as fake since it did not contain the commonly used word in fraudulent reviews. 
We have provided some reasons for the behavior of the models, but they are based on assumptions and may not fully explain why the models behave in a certain way with respect to a particular testing sample. To better understand the behavior of the models, we have turned to explainable AI techniques. Specifically, we utilized the Local Interpretable Model Agnostic Explanation (LIME) \citep{ribeiro-etal-2016-trust} approach in our analysis.
Table \ref{tab:LIME} displays some samples with important features highlighted by LIME based on the predictions of the BanglaBERT models. We provide side by side comparison as BanglaBERT was trained using both \textit{nlpaug} and \textit{bnaug} augmentation techniques. The table utilizes different shades of orange and blue to indicate the words responsible for predicting a review as non-fake or fake respectively. The words with the most dominant impact are colored in deep blue or deep orange while the less dominant words are colored in lighter shades of the same colors. The samples in the table are accurately classified. Some more examples of misclassified data are can be found in Appendix \ref{secA5}.
\begin{table}[]
\centering
\caption{Generalization performance on some unseen test reviews by the proposed weighted ensemble model. (0 stands for \textit{fake} and 1 stands for \textit{non-fake})}
      \label{tab:samples1}
\begin{tabular}{cccc}
\hline
\textbf{Review}& \begin{tabular}[c]{@{}c@{}}\textbf{True}\\ \textbf{Label} \end{tabular} &  \begin{tabular}[c]{@{}c@{}} \textbf{Approach}\\ \textbf{-2} \end{tabular} & \begin{tabular}[c]{@{}c@{}} \textbf{Approach}\\ \textbf{-3} \end{tabular} \\ \hline
\begin{tabular}[c]{@{}c@{}}\parbox[t]{3.8in}{{\bng Taka idey E{I} ibiryain ek khay bha{I} Aar ETaek ibiryain bel? kar kaech ekmn laeg E{I} ebabar  ibiryain?}\\ (Who eats this biryani with money and calls it biryani? Who likes this Boba's biryani?)} \end{tabular}& 1             & 1               & 0              \\ \hline
\begin{tabular}[c]{@{}c@{}}\parbox[t]{3.8in}{{\bng Aapin ik eDsar/T pagl epis/TR ekk edkhel inejek thamaet paern na Aar thamaet Heb na. sada Hl khabar bhir/t eTibl Er bueph et paec/chn AanilimeTD epis/TR ekk ta{O} Aabar 4 dhrenr. Aaech ibibhn/n ephLbhaerr mus eDanaT EbNNG Aaera Aenk Aenk Aa{I}eTm. E{I}sb eDsar/T Aa{I}eTemr pashapaish bueph et reyech Aaera Aenk Aa{I}eTm eJmn phRa{I}D ra{I}s phRa{I}D icekn 3 dhrenr kabab baTar EbNNG garilk ebRD 3 dhrenr sYup EbNNG Aaera Aenk Aa{I}eTm. 9 10 s/TueDn/T Aphaer 750 Takar buephet paec/chn 10\%  chaD. shudhu matR lan/c Er jnY}\\ (Are you crazy about desserts? Can't stop yourself when you see a pastry cake? You don't have to stop anymore. You can get unlimited pastry cakes in the buffet of the White hall with a full table, there are also 4 types of mousse donuts of different flavors and many more items. In addition to these dessert items, the buffet has many other items such as fried rice, fried chicken, 3 types of kebabs, butter and garlic bread, 3 types of soup and many more items. 9/10 student offer 750 taka buffet get 10\% discount. Just for lunch)}  \end{tabular} & 0             & 1               & 0              \\ \hline 
\end{tabular}
\end{table}


\begin{table}[]
\caption{Feature importance explanation generated by LIME based on the best performing individual model (\textit{BanglaBERT}). Here, the true level of the data is given on top of each review}
      \label{tab:LIME}
\begin{tabular}{ccc} %#Solution : Take images of data with English annotations
\hline
Reviews  & LIME Outputs & Predictions \\ \hline
{\begin{tabular}[c]{@{}c@{}}\textbf{non-fake}\\\parbox[t]{3in}{{\bng E{I} grem kilja ThaN/Da krar jnY Erkm Ek gLas laic/ch JethSh/T. na ekhel Aasel bujhaena Jaebna EI laic/chr eTs/T ekmn. dam matR 30 Taka. eriTNNG 9}.{\bng 5/10}\\ (A glass of lassi is enough to cool down the heart in this heat. If you  don't drink it, you can't really understand how the test of this lassi is. The price is only 30 taka. Rating 9.5/10)}\end{tabular}}     &        \begin{tabular}[c]{@{}c@{}}Approach-2\\\raisebox{-0.5\totalheight}{\includegraphics[width=0.35\textwidth, height=15mm]{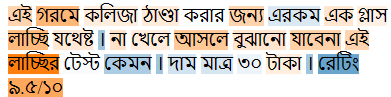}} \\Approach-3\\\raisebox{-0.5\totalheight}{\includegraphics[width=0.35\textwidth, height=15mm]{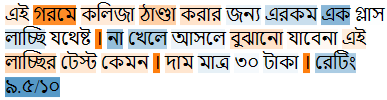}} \end{tabular}   &    {\begin{tabular}[l]{@{}c@{}} \textbf{non-fake}\\ \\ \\ \\\textbf{non-fake} \end{tabular}}       \\ \hline
{\begin{tabular}[c]{@{}c@{}} \textbf{fake}\\\parbox[t]{3in}{{\bng 100 Takar maTn basmit kaic/c jas/T eceT pueT ekheyich jas/T eceT pueT ekhlam E{I} Aphar lueph na inel bYapk ims Heta bYapk ims. dhanmin/D 27 ejeniTk pLajar ApijeT}\\ (I have eaten the 100 taka mutton basmati kacchi to the fullest. if you don't take advantage of this offer, it would have been a big miss. Dhanmondi 27. Opposite Genetic Plaza)}\end{tabular}}     &        \begin{tabular}[c]{@{}c@{}}Approach-2\\\raisebox{-0.5\totalheight}{\includegraphics[width=0.35\textwidth, height=15mm]{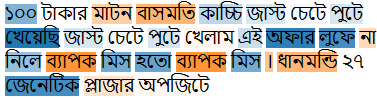}} \\Approach-3\\\raisebox{-0.5\totalheight}{\includegraphics[width=0.35\textwidth, height=15mm]{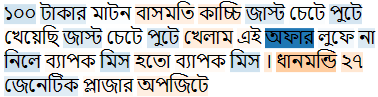}} \end{tabular}   &    {\begin{tabular}[l]{@{}c@{}} \textbf{fake}\\ \\ \\ \\\textbf{fake} \end{tabular}}       \\ \hline
\end{tabular}
\end{table}

The first two samples show that both models are concentrating on nearly identical terms, which explains why they predicted the example as non-fake and fake separately. The two most prevalent terms in the first sample for both models are {\bng grem} and {\bng laic/chr} which are the primary factors for predicting the sample as not fake while the word {\bng eriTNNG} does the opposite. The prediction to classify as fake in the second example is caused by the terms {\bng Aphar}, {\bng ims} which are frequently used in fake reviews. 

\subsection{Misclassification Analysis}
We gathered all the reviews that were incorrectly classified by our ensemble approach. We then conducted a manual analysis of these instances to identify the reasons for misclassifications. Through this analysis, we were able to identify five distinct categories of errors which are discussed in detail below:\\
\textbf{(a) \underline{Inability to handle negation}}: The possible reason behind misclassification could be the inability of the proposed model to comprehend negation. For example, the review instance {\bng sWad Aaegr meta reyech bleba naH. Aaegr ethek men Hela Aaera bhaela elegech.} \textit{[I will not say that the taste is the same as before. I think I like it more than before.]} might have been misclassified because the model was unable to comprehend that the taste of the food had improved, as it failed to take into account the negation in the statement.\\
\textbf{(b) \underline{Inability to deal with contexts}}:  During the translation and transliteration process of reviews, semantic information can be lost leading to a change in the meaning of the sentence. For instance, in this example, {\bng na kha{I}el Aapnar ls gbhiir smudR kl/pna} \textit{[Your loss if you do not try deep sea fantasy]} a pizza named ``deep sea fantasy'' was mistranslated causing the meaning of the sentence to be altered. Some reviews involve irrelevant or extraneous information not focused on the food which is difficult to process for the model. For instance, the review {\bng raja rajY dkhl kreb ETa{I} sabhaibk ikn/tu ibgt ikchu idn dher ranii rajY dkhelr kaj Haet ineyech. ik idn Aa{I}ela buijh naH bar/gar raNii emaHamMdpur imrpuer Eeta idn rajtW ker Ekhn emaHamMdpuer. ta{I} bhablam Ebar EkTu tar rajY edikhya Aais. pRthem khuej epet EkTu kSh/T Hel{O} per iThkmeta epey eglam. emaHamMdpur ekn/dRiiy kelejr paesh{I}. {I}en/Tiryr iney iney matha bYatha EkTu ebish Aamar. ta{I} Aaeg Aais {I}en/Tiryerr kthay. km jaygar medhY Aenk bhaela {I}en/Tiryr ichela Aamar met. kala mainekr edesh Jaeba Aar kaela mainekr sWad ineba naH ta ik ker Hy ineyichlam EkiT bLak bar/gar 95Taka EbNNG EkiT ebNG/gabeyj bar/gar 70Taka.} \textit{[It is normal that the king will take over the kingdom, but for the past few days, the queen has taken up the task of taking over the kingdom. I don't understand what days have come. Burger Queen Mohammadpur has been reigning in Mirpur for so long now in Mohammadpur. So I thought let's see his kingdom. At first it was a little difficult to find, but later I got it right. Next to Mohammadpur Central College. My headache is a bit more with the interior. So let's talk about the interior first. A much better interior in less space in my opinion. How can I go to the land of black manik and not taste black manik? Took a black burger for 95 taka and a bengboyz burger for 70 taka.]} discusses the location, interior and reputation of a restaurant rather than the food.\\
\textbf{(c) \underline{Repetition of words}}: We found some misclassified reviews where we noticed that some words are frequently repeated such as {\bng phRii phRii phRii phRii phRii phRii phRii phRii kabab{O}yalar phuD Aa{I}eTm du{I}Ta iknel EkTa iphR Er epas/T edekh kha{O}yar jnY igeyichlam. ikn/tu Aphar kalek ethek E{I} eDTTa Aaim ekhyal kirin. Aaim taedr ergular kas/Tmar Aaim eshanar per eJ Htashar chap Ta muekh pRkash epeyech esTa tara buejhech EbNNG Aamaek AphariT ideyech. Aenk Aenk dhnYbad Er jnY kabab{O}yalaek. elaekshn Hajiipara Eepk/s esharum Er Ul/Tapaesh. rampura. kabab{O}yala rampura} \textit{[Free Free Free Free Free Free Free Free Kababwala's food items if you buy two, I saw the post of one free and went to eat. But I didn't notice this date from the offer tomorrow. After hearing that I am their regular customer, they understood the look of disappointment on my face and gave me the offer. Many many thanks to Kebabwala for this. Location Hajipara opposite Apex Showroom. Rampura. Kebabwala Rampura]}. It seems that the text is discussing some misclassified reviews where the word ``free'' is repeatedly used in order to draw attention to the reviewer's positive experience of receiving a free food item from the restaurant. We believe that the model becomes confused during prediction because both genuine and fake misclassified reviews are affected by the issue of frequently repeating words.\\ 
\textbf{(d) \underline{Usage of common words}}: Sometimes those who fabricate reviews alter certain words and sentence structures of an authentic review to make it appear genuine. We conducted an analysis on the misclassified reviews and discovered that the top 100 most common words were present in both fake and non-fake reviews. Among the highly recurring words are {\bng Afar, phRii, bhalo, mojar, dhonoYbad, esra, Aenk, eTs/T, vablam, eglam}. \\
\textbf{(e) \underline{Fewer tokens}}: Some reviews might be wrongly classified as there is insufficient information due to a lack of tokens. For example, the review {\bng ipj/ja Oey eTs/T Aar dam Er jonYo s/BnamdhonYo nam. ipj/ja Afar ekhlam mat/ro 300 Takay. bar/gar mat/ro 99 Takay.} \textit{[Pizza way is famous for taste and price. Tried pizza offer only for 300 taka. Burger is only 99 taka.]} simply advertises the offers and does not provide sufficient information about the food. One more such example is {\bng esra muHuur/t labhDggar imrpur Er seNG/go. E{I} bhaelabasar {O}eyejs matR 120 Taka.} \textit{[Best Moments with Lovedogger Mirpur. Lovely wedges are only 120 taka.]} 

\subsection{Comparison with an Existing Ensemble Approach}\label{compare-ensemble}
As there was no previous work done in detecting fake reviews in the Bengali language, we were unable to directly compare our results with those of other studies. To address this limitation, we decided to compare our findings with those of a similar study conducted on a Bengali text dataset. \citep{sharif2022tackling} proposed a weighted ensemble approach on the Bengali aggressive text dataset (BAD) to detect aggressive texts in Bengali. The ensemble method they proposed uses weighted f1 scores as an additional weight for the softmax probabilities for both classes. They computed the individual f1 scores of each model and multiplied the scores as an additional weight to the softmax probability. We applied their proposed method on our BFRD dataset and compared the results using Approach-2 and Approach-3. The comparative results presented in Table \ref{tab:comparison} indicate that the outcomes from our proposed ensemble model on both approaches are very close. Though our proposed ensemble model marginally outperform the method presented in \citep{sharif2022tackling} in terms of weighted F1-scores, the difference is negligible. This implies that our proposed ensemble approach can be a viable solution for other text classification tasks in the Bengali language.

\begin{table}[]
\centering
\caption{Performance comparison by applying another existing ensemble approach on the BFRD dataset}
\label{tab:comparison}
\begin{tabular}{ccc}
\hline
\textbf{Ensemble Model} & \textbf{Approach - 2} & \textbf{Approach - 3} \\ \hline
\citep{sharif2022tackling}                 & 0.982784        & 0.955820       \\
Proposed        & \textbf{0.984279}        & 0.955825      \\ \hline

% Random 0.961400  0.957431
\end{tabular}
\end{table}
\section{Limitations and Future Work}\label{sec7}
Identifying fake reviews is challenging because of the scarcity of fake reviews on different online platforms. It is obvious that our dataset is quite imbalanced, thus we used augmentation techniques to compensate. Furthermore, we could only offer a limited amount of fake data due to the insufficient number of skilled annotators that worked on this project. So, compiling and curating additional fake reviews with experienced annotators could be a great future work. The dataset utilized in this study is limited to restaurant reviews because of the difficulties involved in the data collection and curation process. The study can be expanded to find fake reviews in other categories, such as those for electronics, hotels, movies, books, and more. We also encountered resource constraints, which resulted in fewer sequence lengths for several models, such as sahajBERT and BanglaBERT Large. That may account for the poor performance of these LLMs. In the future, these models can be reassessed using larger sequence lengths. Exploration of different large language models (LLMs) for converting the code-mixed reviews can be another prospect of this research.  An experiment to divide up the dataset and detect several types of fake reviews could be an exciting extension of this work. It would be worth exploring the possibility of developing a detection system or model that could effectively address the various categories of misclassification that have been identified. However, with the use of the proposed model on the fake review dataset, online platforms may benefit from establishing a real-time detection system that can track and flag probable fraudulent reviews as they are submitted.

\section{Conclusion}\label{sec8}
This study addresses the significant concern of the rise of fake reviews in Bengali language on various online platforms. Firstly, the Bengali Fake Review Detection (BFRD) dataset containing over 9,000 food-related reviews has been developed and made publicly accessible. The dataset was carefully annotated by expert annotators, making it a valuable resource for identifying fake reviews in Bengali. Secondly, a weighted ensemble model consisting of four pre-trained Bengali language models has been proposed, outperforming other deep learning and transformer models. Thirdly, a unique text conversion pipeline was created to translate non-Bengali words to Bengali and back transliterate Romanized Bengali to Bengali. Fourthly, text augmentation techniques were employed to handle class imbalance and increase the number of fake review instances. The proposed weighted ensemble model consisting of four different pre-trained BERT models produced a 0.9843 weighted F1-score on 13390 reviews, of which 6695 were fake (1339 were genuine fakes, while the remaining 6695 were augmented using nlpaug library and 6695 were non-fake (randomly chosen from 7710 cases). Finally, extensive experimentation was conducted, and both quantitative and qualitative analysis of the results were presented, including explanations for some of the model's predictions using the LIME text explainer framework. These contributions are crucial towards addressing the issue of fake reviews in Bengali language and can benefit consumers, businesses, and the online review industry as a whole.

\section*{Declarations}
\subsection{Credit authorship contribution statement}
% Omar Sharif: Conceptualization, Data curation, Methodology, Writing - original draft. Mohammed Moshiul Hoque: Conceptualization, Methodology, Writing - review & editing, Supervision.

\textbf{G. M. Shahariar}: Conceptualization, Data collection, Methodology, Implementations, Drafting the manuscript. \textbf{Md Tanvir Rouf Shawon}: Conceptualization, Data collection, Methodology, Implementations, Drafting the manuscript. \textbf{Faisal Muhammad Shah}: Conceptualization, Data collection, Methodology, Manuscript review, Supervision. \textbf{Mohammad Shafiul Alam}: Data collection, Supervision. \textbf{Md. Shahriar Mahbub}: Conceptualization, Methodology, Manuscript review, Supervision.

\subsection{Ethical Approval and Consent to participate}
Not applicable.
\subsection{Consent for publication}
Not applicable.
\subsection{Human and Animal Ethics}
Not applicable.
% \subsection{Availability of data and materials}
% The BFRD dataset is available at - \url{https://github.com/shahariar-shibli/Bengali-Fake-Reviews-A-Benchmark-Dataset-and-Detection-System} 
% \subsection{Code availability}
% The implementation can be found at - \url{https://github.com/shahariar-shibli/Bengali-Fake-Reviews-A-Benchmark-Dataset-and-Detection-System}
\subsection{Competing interests}
The authors declare that they have no competing interests.
\subsection{Funding}
This research work is funded by Ahsanullah University of Science and Technology, Dhaka (AUST) Internal Research Grant supported by The Committee For Advanced Studies And Research (CASR) [Project ID: ARP/2021/CSE/01/2]

% \verb+\printcredits+ command is used after appendix sections to list  author credit taxonomy contribution roles tagged using \verb+\credit+  in frontmatter.


\bibliographystyle{cas-model2-names}

\bibliography{cas-refs}

\begin{thebibliography}{64}
\expandafter\ifx\csname natexlab\endcsname\relax\def\natexlab#1{#1}\fi
\providecommand{\url}[1]{\texttt{#1}}
\providecommand{\href}[2]{#2}
\providecommand{\path}[1]{#1}
\providecommand{\DOIprefix}{doi:}
\providecommand{\ArXivprefix}{arXiv:}
\providecommand{\URLprefix}{URL: }
\providecommand{\Pubmedprefix}{pmid:}
\providecommand{\doi}[1]{\href{http://dx.doi.org/#1}{\path{#1}}}
\providecommand{\Pubmed}[1]{\href{pmid:#1}{\path{#1}}}
\providecommand{\bibinfo}[2]{#2}
\ifx\xfnm\relax \def\xfnm[#1]{\unskip,\space#1}\fi
%Type = Article
\bibitem[{Akil et~al.(2022)Akil, Sultana, Bhattacharjee and Shahriyar}]{akil2022banglaparaphrase}
\bibinfo{author}{Akil, A.}, \bibinfo{author}{Sultana, N.}, \bibinfo{author}{Bhattacharjee, A.}, \bibinfo{author}{Shahriyar, R.}, \bibinfo{year}{2022}.
\newblock \bibinfo{title}{Banglaparaphrase: A high-quality bangla paraphrase dataset}.
\newblock \bibinfo{journal}{arXiv preprint arXiv:2210.05109} .
%Type = Inproceedings
\bibitem[{Banerjee et~al.(2015)Banerjee, Chua and Kim}]{banerjee2015using}
\bibinfo{author}{Banerjee, S.}, \bibinfo{author}{Chua, A.Y.}, \bibinfo{author}{Kim, J.J.}, \bibinfo{year}{2015}.
\newblock \bibinfo{title}{Using supervised learning to classify authentic and fake online reviews}, in: \bibinfo{booktitle}{Proceedings of the 9th international conference on ubiquitous information management and communication}, pp. \bibinfo{pages}{1--7}.
%Type = Inproceedings
\bibitem[{Bhattacharjee et~al.(2022a)Bhattacharjee, Hasan, Ahmad, Mubasshir, Islam, Iqbal, Rahman and Shahriyar}]{bhattacharjee-etal-2022-banglabert}
\bibinfo{author}{Bhattacharjee, A.}, \bibinfo{author}{Hasan, T.}, \bibinfo{author}{Ahmad, W.}, \bibinfo{author}{Mubasshir, K.S.}, \bibinfo{author}{Islam, M.S.}, \bibinfo{author}{Iqbal, A.}, \bibinfo{author}{Rahman, M.S.}, \bibinfo{author}{Shahriyar, R.}, \bibinfo{year}{2022}a.
\newblock \bibinfo{title}{{B}angla{BERT}: Language model pretraining and benchmarks for low-resource language understanding evaluation in {B}angla}, in: \bibinfo{booktitle}{Findings of the Association for Computational Linguistics: NAACL 2022}, \bibinfo{publisher}{Association for Computational Linguistics}, \bibinfo{address}{Seattle, United States}. pp. \bibinfo{pages}{1318--1327}.
\newblock \URLprefix \url{https://aclanthology.org/2022.findings-naacl.98}.
%Type = Article
\bibitem[{Bhattacharjee et~al.(2022b)Bhattacharjee, Hasan, Ahmad and Shahriyar}]{bhattacharjee2022banglanlg}
\bibinfo{author}{Bhattacharjee, A.}, \bibinfo{author}{Hasan, T.}, \bibinfo{author}{Ahmad, W.U.}, \bibinfo{author}{Shahriyar, R.}, \bibinfo{year}{2022}b.
\newblock \bibinfo{title}{Banglanlg: Benchmarks and resources for evaluating low-resource natural language generation in bangla}.
\newblock \bibinfo{journal}{CoRR} \bibinfo{volume}{abs/2205.11081}.
\newblock \URLprefix \url{https://arxiv.org/abs/2205.11081}, \href{http://arxiv.org/abs/2205.11081}{\tt arXiv:2205.11081}.
%Type = Article
\bibitem[{Clark et~al.(2020)Clark, Luong, Le and Manning}]{clark2020electra}
\bibinfo{author}{Clark, K.}, \bibinfo{author}{Luong, M.T.}, \bibinfo{author}{Le, Q.V.}, \bibinfo{author}{Manning, C.D.}, \bibinfo{year}{2020}.
\newblock \bibinfo{title}{Electra: Pre-training text encoders as discriminators rather than generators}.
\newblock \bibinfo{journal}{arXiv preprint arXiv:2003.10555} .
%Type = Article
\bibitem[{Cohen(1960)}]{cohen1960coefficient}
\bibinfo{author}{Cohen, J.}, \bibinfo{year}{1960}.
\newblock \bibinfo{title}{A coefficient of agreement for nominal scales}.
\newblock \bibinfo{journal}{Educational and psychological measurement} \bibinfo{volume}{20}, \bibinfo{pages}{37--46}.
%Type = Inproceedings
\bibitem[{Devlin et~al.(2019)Devlin, Chang, Lee and Toutanova}]{devlin-etal-2019-bert}
\bibinfo{author}{Devlin, J.}, \bibinfo{author}{Chang, M.W.}, \bibinfo{author}{Lee, K.}, \bibinfo{author}{Toutanova, K.}, \bibinfo{year}{2019}.
\newblock \bibinfo{title}{{BERT}: Pre-training of deep bidirectional transformers for language understanding}, in: \bibinfo{booktitle}{Proceedings of the 2019 Conference of the North {A}merican Chapter of the Association for Computational Linguistics: Human Language Technologies, Volume 1 (Long and Short Papers)}, \bibinfo{publisher}{Association for Computational Linguistics}, \bibinfo{address}{Minneapolis, Minnesota}. pp. \bibinfo{pages}{4171--4186}.
\newblock \URLprefix \url{https://aclanthology.org/N19-1423}, \DOIprefix\doi{10.18653/v1/N19-1423}.
%Type = Article
\bibitem[{Dhamani et~al.(2019)Dhamani, Azunre, Gleason, Corcoran, Honke, Kramer and Morgan}]{dhamani2019using}
\bibinfo{author}{Dhamani, N.}, \bibinfo{author}{Azunre, P.}, \bibinfo{author}{Gleason, J.L.}, \bibinfo{author}{Corcoran, C.}, \bibinfo{author}{Honke, G.}, \bibinfo{author}{Kramer, S.}, \bibinfo{author}{Morgan, J.}, \bibinfo{year}{2019}.
\newblock \bibinfo{title}{Using deep networks and transfer learning to address disinformation}.
\newblock \bibinfo{journal}{arXiv preprint arXiv:1905.10412} .
%Type = Article
\bibitem[{Duan et~al.(2008)Duan, Gu and Whinston}]{duan2008online}
\bibinfo{author}{Duan, W.}, \bibinfo{author}{Gu, B.}, \bibinfo{author}{Whinston, A.B.}, \bibinfo{year}{2008}.
\newblock \bibinfo{title}{Do online reviews matter?—an empirical investigation of panel data}.
\newblock \bibinfo{journal}{Decision support systems} \bibinfo{volume}{45}, \bibinfo{pages}{1007--1016}.
%Type = Article
\bibitem[{Etaiwi and Naymat(2017)}]{etaiwi2017impact}
\bibinfo{author}{Etaiwi, W.}, \bibinfo{author}{Naymat, G.}, \bibinfo{year}{2017}.
\newblock \bibinfo{title}{The impact of applying different preprocessing steps on review spam detection}.
\newblock \bibinfo{journal}{Procedia computer science} \bibinfo{volume}{113}, \bibinfo{pages}{273--279}.
%Type = Article
\bibitem[{Fleiss(1971)}]{fleiss1971measuring}
\bibinfo{author}{Fleiss, J.L.}, \bibinfo{year}{1971}.
\newblock \bibinfo{title}{Measuring nominal scale agreement among many raters.}
\newblock \bibinfo{journal}{Psychological bulletin} \bibinfo{volume}{76}, \bibinfo{pages}{378}.
%Type = Article
\bibitem[{Graves and Schmidhuber(2005)}]{graves2005framewise}
\bibinfo{author}{Graves, A.}, \bibinfo{author}{Schmidhuber, J.}, \bibinfo{year}{2005}.
\newblock \bibinfo{title}{Framewise phoneme classification with bidirectional lstm and other neural network architectures}.
\newblock \bibinfo{journal}{Neural networks} \bibinfo{volume}{18}, \bibinfo{pages}{602--610}.
%Type = Article
\bibitem[{Gundapu and Mamidi(2021)}]{gundapu2021transformer}
\bibinfo{author}{Gundapu, S.}, \bibinfo{author}{Mamidi, R.}, \bibinfo{year}{2021}.
\newblock \bibinfo{title}{Transformer based automatic covid-19 fake news detection system}.
\newblock \bibinfo{journal}{arXiv preprint arXiv:2101.00180} .
%Type = Article
\bibitem[{Guo et~al.(2023)Guo, Mustafaoglu and Koundal}]{guo2023spam}
\bibinfo{author}{Guo, Y.}, \bibinfo{author}{Mustafaoglu, Z.}, \bibinfo{author}{Koundal, D.}, \bibinfo{year}{2023}.
\newblock \bibinfo{title}{Spam detection using bidirectional transformers and machine learning classifier algorithms}.
\newblock \bibinfo{journal}{Journal of Computational and Cognitive Engineering} \bibinfo{volume}{2}, \bibinfo{pages}{5--9}.
%Type = Inproceedings
\bibitem[{Gupta et~al.(2021)Gupta, Gandhi and Chakravarthi}]{gupta2021leveraging}
\bibinfo{author}{Gupta, P.}, \bibinfo{author}{Gandhi, S.}, \bibinfo{author}{Chakravarthi, B.R.}, \bibinfo{year}{2021}.
\newblock \bibinfo{title}{Leveraging transfer learning techniques-bert, roberta, albert and distilbert for fake review detection}, in: \bibinfo{booktitle}{Forum for Information Retrieval Evaluation}, pp. \bibinfo{pages}{75--82}.
%Type = Inproceedings
\bibitem[{Gutierrez-Espinoza et~al.(2020)Gutierrez-Espinoza, Abri, Namin, Jones and Sears}]{gutierrez2020ensemble}
\bibinfo{author}{Gutierrez-Espinoza, L.}, \bibinfo{author}{Abri, F.}, \bibinfo{author}{Namin, A.S.}, \bibinfo{author}{Jones, K.S.}, \bibinfo{author}{Sears, D.R.}, \bibinfo{year}{2020}.
\newblock \bibinfo{title}{Ensemble learning for detecting fake reviews}, in: \bibinfo{booktitle}{2020 IEEE 44th Annual Computers, Software, and Applications Conference (COMPSAC)}, \bibinfo{organization}{IEEE}. pp. \bibinfo{pages}{1320--1325}.
%Type = Article
\bibitem[{Ha et~al.(2015)Ha, Bae and Son}]{ha2015impact}
\bibinfo{author}{Ha, S.H.}, \bibinfo{author}{Bae, S.}, \bibinfo{author}{Son, L.K.}, \bibinfo{year}{2015}.
\newblock \bibinfo{title}{Impact of online consumer reviews on product sales: Quantitative analysis of the source effect}.
\newblock \bibinfo{journal}{Applied Mathematics and Information Sciences} \bibinfo{volume}{9}, \bibinfo{pages}{373--387}.
%Type = Article
\bibitem[{Hammad and El-Halees(2013)}]{hammad2013approach}
\bibinfo{author}{Hammad, A.A.}, \bibinfo{author}{El-Halees, A.}, \bibinfo{year}{2013}.
\newblock \bibinfo{title}{An approach for detecting spam in arabic opinion reviews}.
\newblock \bibinfo{journal}{The International Arab Journal of Information Technology} \bibinfo{volume}{12}.
%Type = Inproceedings
\bibitem[{Hasan et~al.(2020)Hasan, Bhattacharjee, Samin, Hasan, Basak, Rahman and Shahriyar}]{hasan-etal-2020-low}
\bibinfo{author}{Hasan, T.}, \bibinfo{author}{Bhattacharjee, A.}, \bibinfo{author}{Samin, K.}, \bibinfo{author}{Hasan, M.}, \bibinfo{author}{Basak, M.}, \bibinfo{author}{Rahman, M.S.}, \bibinfo{author}{Shahriyar, R.}, \bibinfo{year}{2020}.
\newblock \bibinfo{title}{Not low-resource anymore: Aligner ensembling, batch filtering, and new datasets for {B}engali-{E}nglish machine translation}, in: \bibinfo{booktitle}{Proceedings of the 2020 Conference on Empirical Methods in Natural Language Processing (EMNLP)}, \bibinfo{publisher}{Association for Computational Linguistics}, \bibinfo{address}{Online}. pp. \bibinfo{pages}{2612--2623}.
\newblock \URLprefix \url{https://www.aclweb.org/anthology/2020.emnlp-main.207}, \DOIprefix\doi{10.18653/v1/2020.emnlp-main.207}.
%Type = Article
\bibitem[{Hern{\'a}ndez-Casta{\~n}eda et~al.(2017)Hern{\'a}ndez-Casta{\~n}eda, Calvo, Gelbukh and Flores}]{hernandez2017cross}
\bibinfo{author}{Hern{\'a}ndez-Casta{\~n}eda, {\'A}.}, \bibinfo{author}{Calvo, H.}, \bibinfo{author}{Gelbukh, A.}, \bibinfo{author}{Flores, J.J.G.}, \bibinfo{year}{2017}.
\newblock \bibinfo{title}{Cross-domain deception detection using support vector networks}.
\newblock \bibinfo{journal}{Soft Computing} \bibinfo{volume}{21}, \bibinfo{pages}{585--595}.
%Type = Article
\bibitem[{Javed et~al.(2021)Javed, Majeed, Mujtaba and Beg}]{javed2021fake}
\bibinfo{author}{Javed, M.S.}, \bibinfo{author}{Majeed, H.}, \bibinfo{author}{Mujtaba, H.}, \bibinfo{author}{Beg, M.O.}, \bibinfo{year}{2021}.
\newblock \bibinfo{title}{Fake reviews classification using deep learning ensemble of shallow convolutions}.
\newblock \bibinfo{journal}{Journal of Computational Social Science} , \bibinfo{pages}{1--20}.
%Type = Inproceedings
\bibitem[{Jindal and Liu(2007)}]{jindal2007review}
\bibinfo{author}{Jindal, N.}, \bibinfo{author}{Liu, B.}, \bibinfo{year}{2007}.
\newblock \bibinfo{title}{Review spam detection}, in: \bibinfo{booktitle}{Proceedings of the 16th international conference on World Wide Web}, pp. \bibinfo{pages}{1189--1190}.
%Type = Inproceedings
\bibitem[{Jindal and Liu(2008)}]{jindal2008opinion}
\bibinfo{author}{Jindal, N.}, \bibinfo{author}{Liu, B.}, \bibinfo{year}{2008}.
\newblock \bibinfo{title}{Opinion spam and analysis}, in: \bibinfo{booktitle}{Proceedings of the 2008 international conference on web search and data mining}, pp. \bibinfo{pages}{219--230}.
%Type = Article
\bibitem[{Khurshid et~al.(2019)Khurshid, Zhu, Xu, Ahmad and Ahmad}]{khurshid2019enactment}
\bibinfo{author}{Khurshid, F.}, \bibinfo{author}{Zhu, Y.}, \bibinfo{author}{Xu, Z.}, \bibinfo{author}{Ahmad, M.}, \bibinfo{author}{Ahmad, M.}, \bibinfo{year}{2019}.
\newblock \bibinfo{title}{Enactment of ensemble learning for review spam detection on selected features}.
\newblock \bibinfo{journal}{International Journal of Computational Intelligence Systems} \bibinfo{volume}{12}, \bibinfo{pages}{387--394}.
%Type = Inproceedings
\bibitem[{Khurshid et~al.(2017)Khurshid, Zhu, Yohannese and Iqbal}]{khurshid2017recital}
\bibinfo{author}{Khurshid, F.}, \bibinfo{author}{Zhu, Y.}, \bibinfo{author}{Yohannese, C.W.}, \bibinfo{author}{Iqbal, M.}, \bibinfo{year}{2017}.
\newblock \bibinfo{title}{Recital of supervised learning on review spam detection: An empirical analysis}, in: \bibinfo{booktitle}{2017 12th International Conference on Intelligent Systems and Knowledge Engineering (ISKE)}, \bibinfo{organization}{IEEE}. pp. \bibinfo{pages}{1--6}.
%Type = Article
\bibitem[{Lan et~al.(2019)Lan, Chen, Goodman, Gimpel, Sharma and Soricut}]{lan2019albert}
\bibinfo{author}{Lan, Z.}, \bibinfo{author}{Chen, M.}, \bibinfo{author}{Goodman, S.}, \bibinfo{author}{Gimpel, K.}, \bibinfo{author}{Sharma, P.}, \bibinfo{author}{Soricut, R.}, \bibinfo{year}{2019}.
\newblock \bibinfo{title}{Albert: A lite bert for self-supervised learning of language representations}.
\newblock \bibinfo{journal}{arXiv preprint arXiv:1909.11942} .
%Type = Article
\bibitem[{LeCun et~al.(1998)LeCun, Bottou, Bengio and Haffner}]{lecun1998gradient}
\bibinfo{author}{LeCun, Y.}, \bibinfo{author}{Bottou, L.}, \bibinfo{author}{Bengio, Y.}, \bibinfo{author}{Haffner, P.}, \bibinfo{year}{1998}.
\newblock \bibinfo{title}{Gradient-based learning applied to document recognition}.
\newblock \bibinfo{journal}{Proceedings of the IEEE} \bibinfo{volume}{86}, \bibinfo{pages}{2278--2324}.
%Type = Inproceedings
\bibitem[{Li et~al.(2015)Li, Ren, Qin and Liu}]{li2015learning}
\bibinfo{author}{Li, L.}, \bibinfo{author}{Ren, W.}, \bibinfo{author}{Qin, B.}, \bibinfo{author}{Liu, T.}, \bibinfo{year}{2015}.
\newblock \bibinfo{title}{Learning document representation for deceptive opinion spam detection}, in: \bibinfo{booktitle}{Chinese Computational Linguistics and Natural Language Processing Based on Naturally Annotated Big Data: 14th China National Conference, CCL 2015 and Third International Symposium, NLP-NABD 2015, Guangzhou, China, November 13-14, 2015, Proceedings 14}, \bibinfo{organization}{Springer}. pp. \bibinfo{pages}{393--404}.
%Type = Article
\bibitem[{Liu et~al.(2020)Liu, Jing and Li}]{liu2020incorporating}
\bibinfo{author}{Liu, W.}, \bibinfo{author}{Jing, W.}, \bibinfo{author}{Li, Y.}, \bibinfo{year}{2020}.
\newblock \bibinfo{title}{Incorporating feature representation into bilstm for deceptive review detection}.
\newblock \bibinfo{journal}{Computing} \bibinfo{volume}{102}, \bibinfo{pages}{701--715}.
%Type = Article
\bibitem[{Liu et~al.(2021)Liu, Lu and Nayak}]{liu2021spam}
\bibinfo{author}{Liu, X.}, \bibinfo{author}{Lu, H.}, \bibinfo{author}{Nayak, A.}, \bibinfo{year}{2021}.
\newblock \bibinfo{title}{A spam transformer model for sms spam detection}.
\newblock \bibinfo{journal}{IEEE Access} \bibinfo{volume}{9}, \bibinfo{pages}{80253--80263}.
%Type = Article
\bibitem[{Lu et~al.(2021)Lu, Li, Wang and Qin}]{lu2021cnn}
\bibinfo{author}{Lu, W.}, \bibinfo{author}{Li, J.}, \bibinfo{author}{Wang, J.}, \bibinfo{author}{Qin, L.}, \bibinfo{year}{2021}.
\newblock \bibinfo{title}{A cnn-bilstm-am method for stock price prediction}.
\newblock \bibinfo{journal}{Neural Computing and Applications} \bibinfo{volume}{33}, \bibinfo{pages}{4741--4753}.
%Type = Incollection
\bibitem[{Luca(2015)}]{luca2015user}
\bibinfo{author}{Luca, M.}, \bibinfo{year}{2015}.
\newblock \bibinfo{title}{User-generated content and social media}, in: \bibinfo{booktitle}{Handbook of media Economics}. \bibinfo{publisher}{Elsevier}. volume~\bibinfo{volume}{1}, pp. \bibinfo{pages}{563--592}.
%Type = Article
\bibitem[{Luca(2016)}]{luca2016reviews}
\bibinfo{author}{Luca, M.}, \bibinfo{year}{2016}.
\newblock \bibinfo{title}{Reviews, reputation, and revenue: The case of yelp. com}.
\newblock \bibinfo{journal}{Com (March 15, 2016). Harvard Business School NOM Unit Working Paper} .
%Type = Article
\bibitem[{Luca and Zervas(2016)}]{luca2016fake}
\bibinfo{author}{Luca, M.}, \bibinfo{author}{Zervas, G.}, \bibinfo{year}{2016}.
\newblock \bibinfo{title}{Fake it till you make it: Reputation, competition, and yelp review fraud}.
\newblock \bibinfo{journal}{Management Science} \bibinfo{volume}{62}, \bibinfo{pages}{3412--3427}.
%Type = Misc
\bibitem[{Ma(2019)}]{ma2019nlpaug}
\bibinfo{author}{Ma, E.}, \bibinfo{year}{2019}.
\newblock \bibinfo{title}{Nlp augmentation}.
\newblock \URLprefix \url{https://github.com/makcedward/nlpaug}.
%Type = Inproceedings
\bibitem[{Mani et~al.(2018)Mani, Kumari, Jain and Kumar}]{mani2018spam}
\bibinfo{author}{Mani, S.}, \bibinfo{author}{Kumari, S.}, \bibinfo{author}{Jain, A.}, \bibinfo{author}{Kumar, P.}, \bibinfo{year}{2018}.
\newblock \bibinfo{title}{Spam review detection using ensemble machine learning}, in: \bibinfo{booktitle}{Machine Learning and Data Mining in Pattern Recognition: 14th International Conference, MLDM 2018, New York, NY, USA, July 15-19, 2018, Proceedings, Part II 14}, \bibinfo{organization}{Springer}. pp. \bibinfo{pages}{198--209}.
%Type = Article
\bibitem[{Mir et~al.(2023)Mir, Khan and Chishti}]{mir2023online}
\bibinfo{author}{Mir, A.Q.}, \bibinfo{author}{Khan, F.Y.}, \bibinfo{author}{Chishti, M.A.}, \bibinfo{year}{2023}.
\newblock \bibinfo{title}{Online fake review detection using supervised machine learning and bert model}.
\newblock \bibinfo{journal}{arXiv preprint arXiv:2301.03225} .
%Type = Article
\bibitem[{Mohawesh et~al.(2021)Mohawesh, Tran, Ollington and Xu}]{mohawesh2021analysis}
\bibinfo{author}{Mohawesh, R.}, \bibinfo{author}{Tran, S.}, \bibinfo{author}{Ollington, R.}, \bibinfo{author}{Xu, S.}, \bibinfo{year}{2021}.
\newblock \bibinfo{title}{Analysis of concept drift in fake reviews detection}.
\newblock \bibinfo{journal}{Expert Systems with Applications} \bibinfo{volume}{169}, \bibinfo{pages}{114318}.
%Type = Inproceedings
\bibitem[{Mukherjee et~al.(2012)Mukherjee, Liu and Glance}]{mukherjee2012spotting}
\bibinfo{author}{Mukherjee, A.}, \bibinfo{author}{Liu, B.}, \bibinfo{author}{Glance, N.}, \bibinfo{year}{2012}.
\newblock \bibinfo{title}{Spotting fake reviewer groups in consumer reviews}, in: \bibinfo{booktitle}{Proceedings of the 21st international conference on World Wide Web}, pp. \bibinfo{pages}{191--200}.
%Type = Inproceedings
\bibitem[{Mukherjee et~al.(2013a)Mukherjee, Venkataraman, Liu and Glance}]{mukherjee2013yelp}
\bibinfo{author}{Mukherjee, A.}, \bibinfo{author}{Venkataraman, V.}, \bibinfo{author}{Liu, B.}, \bibinfo{author}{Glance, N.}, \bibinfo{year}{2013}a.
\newblock \bibinfo{title}{What yelp fake review filter might be doing?}, in: \bibinfo{booktitle}{Proceedings of the international AAAI conference on web and social media}, pp. \bibinfo{pages}{409--418}.
%Type = Article
\bibitem[{Mukherjee et~al.(2013b)Mukherjee, Venkataraman, Liu, Glance et~al.}]{mukherjee2013fake}
\bibinfo{author}{Mukherjee, A.}, \bibinfo{author}{Venkataraman, V.}, \bibinfo{author}{Liu, B.}, \bibinfo{author}{Glance, N.}, et~al., \bibinfo{year}{2013}b.
\newblock \bibinfo{title}{Fake review detection: Classification and analysis of real and pseudo reviews}.
\newblock \bibinfo{journal}{UIC-CS-03-2013. Technical Report} .
%Type = Misc
\bibitem[{Nisen()}]{businessinsiderFakeReviews}
\bibinfo{author}{Nisen, M.}, .
\newblock \bibinfo{title}{{F}ake {R}eviews {A}re {B}ecoming {A}n {E}ven {B}igger {P}roblem {F}or {B}usinesses --- businessinsider.com}.
\newblock \bibinfo{howpublished}{\url{https://www.businessinsider.com/fake-reviews-are-becoming-a-huge-problem-for-businesses-2012-9}}.
\newblock \bibinfo{note}{[Accessed 06-May-2023]}.
%Type = Inproceedings
\bibitem[{Ott et~al.(2011)Ott, Choi, Cardie and Hancock}]{ott-etal-2011-finding}
\bibinfo{author}{Ott, M.}, \bibinfo{author}{Choi, Y.}, \bibinfo{author}{Cardie, C.}, \bibinfo{author}{Hancock, J.T.}, \bibinfo{year}{2011}.
\newblock \bibinfo{title}{Finding deceptive opinion spam by any stretch of the imagination}, in: \bibinfo{booktitle}{Proceedings of the 49th Annual Meeting of the Association for Computational Linguistics: Human Language Technologies}, \bibinfo{publisher}{Association for Computational Linguistics}, \bibinfo{address}{Portland, Oregon, USA}. pp. \bibinfo{pages}{309--319}.
\newblock \URLprefix \url{https://aclanthology.org/P11-1032}.
%Type = Article
\bibitem[{Price et~al.(2020)Price, Gifford-Moore, Fleming, Musker, Roichman, Sylvain, Thain, Dixon and Sorensen}]{price2020six}
\bibinfo{author}{Price, I.}, \bibinfo{author}{Gifford-Moore, J.}, \bibinfo{author}{Fleming, J.}, \bibinfo{author}{Musker, S.}, \bibinfo{author}{Roichman, M.}, \bibinfo{author}{Sylvain, G.}, \bibinfo{author}{Thain, N.}, \bibinfo{author}{Dixon, L.}, \bibinfo{author}{Sorensen, J.}, \bibinfo{year}{2020}.
\newblock \bibinfo{title}{Six attributes of unhealthy conversation}.
\newblock \bibinfo{journal}{arXiv preprint arXiv:2010.07410} .
%Type = Article
\bibitem[{Rao et~al.(2021)Rao, Verma and Bhatia}]{rao2021review}
\bibinfo{author}{Rao, S.}, \bibinfo{author}{Verma, A.K.}, \bibinfo{author}{Bhatia, T.}, \bibinfo{year}{2021}.
\newblock \bibinfo{title}{A review on social spam detection: Challenges, open issues, and future directions}.
\newblock \bibinfo{journal}{Expert Systems with Applications} \bibinfo{volume}{186}, \bibinfo{pages}{115742}.
%Type = Inproceedings
\bibitem[{Ren and Zhang(2016)}]{ren2016deceptive}
\bibinfo{author}{Ren, Y.}, \bibinfo{author}{Zhang, Y.}, \bibinfo{year}{2016}.
\newblock \bibinfo{title}{Deceptive opinion spam detection using neural network}, in: \bibinfo{booktitle}{Proceedings of COLING 2016, the 26th International Conference on Computational Linguistics: Technical Papers}, pp. \bibinfo{pages}{140--150}.
%Type = Article
\bibitem[{Rhanoui et~al.(2019)Rhanoui, Mikram, Yousfi and Barzali}]{rhanoui2019cnn}
\bibinfo{author}{Rhanoui, M.}, \bibinfo{author}{Mikram, M.}, \bibinfo{author}{Yousfi, S.}, \bibinfo{author}{Barzali, S.}, \bibinfo{year}{2019}.
\newblock \bibinfo{title}{A cnn-bilstm model for document-level sentiment analysis}.
\newblock \bibinfo{journal}{Machine Learning and Knowledge Extraction} \bibinfo{volume}{1}, \bibinfo{pages}{832--847}.
%Type = Inproceedings
\bibitem[{Ribeiro et~al.(2016)Ribeiro, Singh and Guestrin}]{ribeiro-etal-2016-trust}
\bibinfo{author}{Ribeiro, M.}, \bibinfo{author}{Singh, S.}, \bibinfo{author}{Guestrin, C.}, \bibinfo{year}{2016}.
\newblock \bibinfo{title}{{``}why should {I} trust you?{''}: Explaining the predictions of any classifier}, in: \bibinfo{booktitle}{Proceedings of the 2016 Conference of the North {A}merican Chapter of the Association for Computational Linguistics: Demonstrations}, \bibinfo{publisher}{Association for Computational Linguistics}, \bibinfo{address}{San Diego, California}. pp. \bibinfo{pages}{97--101}.
\newblock \URLprefix \url{https://aclanthology.org/N16-3020}, \DOIprefix\doi{10.18653/v1/N16-3020}.
%Type = Article
\bibitem[{Rojas-Galeano(2021)}]{rojas2021using}
\bibinfo{author}{Rojas-Galeano, S.}, \bibinfo{year}{2021}.
\newblock \bibinfo{title}{Using bert encoding to tackle the mad-lib attack in sms spam detection}.
\newblock \bibinfo{journal}{arXiv preprint arXiv:2107.06400} .
%Type = Article
\bibitem[{Sahmoud et~al.(2022)Sahmoud, Mikki et~al.}]{sahmoud2022spam}
\bibinfo{author}{Sahmoud, T.}, \bibinfo{author}{Mikki, D.}, et~al., \bibinfo{year}{2022}.
\newblock \bibinfo{title}{Spam detection using bert}.
\newblock \bibinfo{journal}{arXiv preprint arXiv:2206.02443} .
%Type = Article
\bibitem[{Salminen et~al.(2022)Salminen, Kandpal, Kamel, Jung and Jansen}]{salminen2022creating}
\bibinfo{author}{Salminen, J.}, \bibinfo{author}{Kandpal, C.}, \bibinfo{author}{Kamel, A.M.}, \bibinfo{author}{Jung, S.g.}, \bibinfo{author}{Jansen, B.J.}, \bibinfo{year}{2022}.
\newblock \bibinfo{title}{Creating and detecting fake reviews of online products}.
\newblock \bibinfo{journal}{Journal of Retailing and Consumer Services} \bibinfo{volume}{64}, \bibinfo{pages}{102771}.
%Type = Misc
\bibitem[{Sarker(2020)}]{Sagor_2020}
\bibinfo{author}{Sarker, S.}, \bibinfo{year}{2020}.
\newblock \bibinfo{title}{Banglabert: Bengali mask language model for bengali language understanding}.
\newblock \URLprefix \url{https://github.com/sagorbrur/bangla-bert}.
%Type = Inproceedings
\bibitem[{Sedighi et~al.(2017)Sedighi, Ebrahimpour-Komleh and Bagheri}]{sedighi2017rlosd}
\bibinfo{author}{Sedighi, Z.}, \bibinfo{author}{Ebrahimpour-Komleh, H.}, \bibinfo{author}{Bagheri, A.}, \bibinfo{year}{2017}.
\newblock \bibinfo{title}{Rlosd: Representation learning based opinion spam detection}, in: \bibinfo{booktitle}{2017 3rd Iranian Conference on Intelligent Systems and Signal Processing (ICSPIS)}, \bibinfo{organization}{IEEE}. pp. \bibinfo{pages}{74--80}.
%Type = Inproceedings
\bibitem[{Shahariar et~al.(2019)Shahariar, Biswas, Omar, Shah and Binte~Hassan}]{shiblispam}
\bibinfo{author}{Shahariar, G.M.}, \bibinfo{author}{Biswas, S.}, \bibinfo{author}{Omar, F.}, \bibinfo{author}{Shah, F.M.}, \bibinfo{author}{Binte~Hassan, S.}, \bibinfo{year}{2019}.
\newblock \bibinfo{title}{Spam review detection using deep learning}, in: \bibinfo{booktitle}{2019 IEEE 10th Annual Information Technology, Electronics and Mobile Communication Conference (IEMCON)}, pp. \bibinfo{pages}{0027--0033}.
\newblock \DOIprefix\doi{10.1109/IEMCON.2019.8936148}.
%Type = Article
\bibitem[{Shan et~al.(2021)Shan, Zhou and Zhang}]{shan2021conflicts}
\bibinfo{author}{Shan, G.}, \bibinfo{author}{Zhou, L.}, \bibinfo{author}{Zhang, D.}, \bibinfo{year}{2021}.
\newblock \bibinfo{title}{From conflicts and confusion to doubts: Examining review inconsistency for fake review detection}.
\newblock \bibinfo{journal}{Decision Support Systems} \bibinfo{volume}{144}, \bibinfo{pages}{113513}.
%Type = Article
\bibitem[{Sharif and Hoque(2022)}]{sharif2022tackling}
\bibinfo{author}{Sharif, O.}, \bibinfo{author}{Hoque, M.M.}, \bibinfo{year}{2022}.
\newblock \bibinfo{title}{Tackling cyber-aggression: Identification and fine-grained categorization of aggressive texts on social media using weighted ensemble of transformers}.
\newblock \bibinfo{journal}{Neurocomputing} \bibinfo{volume}{490}, \bibinfo{pages}{462--481}.
%Type = Article
\bibitem[{Shifath et~al.(2021)Shifath, Khan, Islam et~al.}]{shifath2021transformer}
\bibinfo{author}{Shifath, S.}, \bibinfo{author}{Khan, M.F.}, \bibinfo{author}{Islam, M.}, et~al., \bibinfo{year}{2021}.
\newblock \bibinfo{title}{A transformer based approach for fighting covid-19 fake news}.
\newblock \bibinfo{journal}{arXiv preprint arXiv:2101.12027} .
%Type = Inproceedings
\bibitem[{Wang et~al.(2018a)Wang, Day, Chen and Liou}]{wang2018lstm}
\bibinfo{author}{Wang, C.C.}, \bibinfo{author}{Day, M.Y.}, \bibinfo{author}{Chen, C.C.}, \bibinfo{author}{Liou, J.W.}, \bibinfo{year}{2018}a.
\newblock \bibinfo{title}{Detecting spamming reviews using long short-term memory recurrent neural network framework}, in: \bibinfo{booktitle}{Proceedings of the 2nd International Conference on E-commerce, E-Business and E-Government}, pp. \bibinfo{pages}{16--20}.
%Type = Inproceedings
\bibitem[{Wang et~al.(2018b)Wang, Liu and Zhao}]{wang2018detecting}
\bibinfo{author}{Wang, X.}, \bibinfo{author}{Liu, K.}, \bibinfo{author}{Zhao, J.}, \bibinfo{year}{2018}b.
\newblock \bibinfo{title}{Detecting deceptive review spam via attention-based neural networks}, in: \bibinfo{booktitle}{Natural Language Processing and Chinese Computing: 6th CCF International Conference, NLPCC 2017, Dalian, China, November 8--12, 2017, Proceedings 6}, \bibinfo{organization}{Springer}. pp. \bibinfo{pages}{866--876}.
%Type = Misc
\bibitem[{Wikipedia()}]{wikipediaListLanguages}
\bibinfo{author}{Wikipedia}, .
\newblock \bibinfo{title}{List of languages by total number of speakers - {W}ikipedia --- en.wikipedia.org}.
\newblock \bibinfo{howpublished}{\url{https://en.wikipedia.org/wiki/List_of_languages_by_total_number_of_speakers}}.
\newblock \bibinfo{note}{[Accessed 07-May-2023]}.
%Type = Article
\bibitem[{Yao et~al.(2021)Yao, Zheng and Jiang}]{yao2021ensemble}
\bibinfo{author}{Yao, J.}, \bibinfo{author}{Zheng, Y.}, \bibinfo{author}{Jiang, H.}, \bibinfo{year}{2021}.
\newblock \bibinfo{title}{An ensemble model for fake online review detection based on data resampling, feature pruning, and parameter optimization}.
\newblock \bibinfo{journal}{IEEE Access} \bibinfo{volume}{9}, \bibinfo{pages}{16914--16927}.
%Type = Article
\bibitem[{Zeng et~al.(2019)Zeng, Lin, Chen, Chen, Lan and Liu}]{zeng2019review}
\bibinfo{author}{Zeng, Z.Y.}, \bibinfo{author}{Lin, J.J.}, \bibinfo{author}{Chen, M.S.}, \bibinfo{author}{Chen, M.H.}, \bibinfo{author}{Lan, Y.Q.}, \bibinfo{author}{Liu, J.L.}, \bibinfo{year}{2019}.
\newblock \bibinfo{title}{A review structure based ensemble model for deceptive review spam}.
\newblock \bibinfo{journal}{Information} \bibinfo{volume}{10}, \bibinfo{pages}{243}.
%Type = Article
\bibitem[{Zhang et~al.(2018)Zhang, Du, Yoshida and Wang}]{zhang2018dri}
\bibinfo{author}{Zhang, W.}, \bibinfo{author}{Du, Y.}, \bibinfo{author}{Yoshida, T.}, \bibinfo{author}{Wang, Q.}, \bibinfo{year}{2018}.
\newblock \bibinfo{title}{Dri-rcnn: An approach to deceptive review identification using recurrent convolutional neural network}.
\newblock \bibinfo{journal}{Information Processing \& Management} \bibinfo{volume}{54}, \bibinfo{pages}{576--592}.
%Type = Article
\bibitem[{Zhao et~al.(2018)Zhao, Xu, Liu, Guo and Yun}]{zhao2018towards}
\bibinfo{author}{Zhao, S.}, \bibinfo{author}{Xu, Z.}, \bibinfo{author}{Liu, L.}, \bibinfo{author}{Guo, M.}, \bibinfo{author}{Yun, J.}, \bibinfo{year}{2018}.
\newblock \bibinfo{title}{Towards accurate deceptive opinions detection based on word order-preserving cnn}.
\newblock \bibinfo{journal}{Mathematical Problems in Engineering} \bibinfo{volume}{2018}.

\end{thebibliography}

\appendix

\section{Some more Instances of Fake Reviews} \label{secA7}
In this part, we present a few more instances of fake reviews for a better understanding. Table \ref{tab:moreannot} shows two more fake samples. The first sample satisfies criteria 3 and 5, while the last sample satisfies criteria 5, which pertains to professional or poetic writing. 

\begin{table}[H]
\caption{Few more annotated fake review instances with corresponding annotation criteria}
\label{tab:moreannot}
\centering
% \resizebox{\columnwidth}{!}
\begin{tabular}{lc}
\hline

\multicolumn{1}{c}{\textbf{Fake reviews}}  & \textbf{Criteria} \\ \hline
\begin{tabular}[c]{@{}l@{}}\parbox[t]{5.4in}{{\bng bn/dhu tuim Eka Hel id{O} Aamay Dak, etamar Takay isk/s pYak khaeba Aaim sararat. sir raet ekn khaeba?
bn/dhu Jkhn cYaelNJ/j edy isk/s pYak eshSh krar ikn/tu es inej{I}  eshSh kret paer na EbNNG Aamar ibl tar{I} ed{O}Ja laeg!} Tour De Cyclist - Uttara {\bng Er isk/s pYak bar/gar sm/per/k Aenk sunam shunich! eta bhablam E{I} rakKs Taek eshSh ker{I} Aais! bar/gar E Aenk ebish pirmaen  pYaiT thakay Er ibhterr idk juis krar jenY Aenk ebish pirmaeN icj EbNNG emeya {I}Uj kra H{I}ech. Jar kareN shukna lageb na bar/gar Ta! puraTa eshSh kra khadk na Hel sm/bhb na! Aa{I}eTm: isk/s pYak bar/gar pRa{I}s: 600/= eriTNNG: 8}.{\bng 5/10. saeth iDRNNGks iHeseb cekaelT es/pak ineyichlam. ETay Hu{I}pD ikRm AYaD ker TRa{I} krebn AbshY{I}. AtYn/t ejas laeg ekhet. Aa{I}eTm: cekaelT es/pak U{I}th Hu{I}pD ikRm pRa{I}s: 100/= + 60/= (Hu{I}pD ikRm Er jenY) eriTNNG: 9}.{\bng 5/10 elaekshn:} House \# 05, Road \# 15, Rabindra Sarani, Sector \# 03, Uttara, Dhaka. {\bng cYaelNJ/j E Aaim ijtichlam ta{I} ibl eta pheTagRaphar idech.}

(Friend, if you are alone, just call me, I will have a six-pack on your tab all night. Sorry, why would you eat at night? When a friend challenges to finish the six-pack burger at Tour De Cyclist - Uttara, but can not finish it themselves, I end up footing the bill! I have heard a lot about the six-pack burger at Tour De Cyclist - Uttara! So, I thought I would come and conquer this beast! The burger is loaded with lots of cheese and mayo to make it juicier. You will not feel dry while eating it! Finishing it all is a must; there is no other way! Item: Six-pack burger Price: 600/= Rating: 8.5/10 I also ordered a chocolate spook as a drink. You must try it with whipped cream added. It's incredibly refreshing to have. Item: Chocolate spook with whipped cream Price: 100/= + 60/= (for whipped cream) Rating: 9.5/10 Location: House \# 05, Road \# 15, Rabinndra Sarani, Sector \# 03, Uttara, Dhaka. Since I won the challenge, the bill was on me, as the photographer has captured.)}\end{tabular}  & 3, 5                 \\ \hline
\begin{tabular}[c]{@{}l@{}}\parbox[t]{5.4in}{{\bng dhanmiN/D kaipey buephr bs Jkhn Dhakar rajdhanii imrpuer tkhn na eJey ik para Jay? blichlam eHTar/sibHiin es{I} bueph khYat} The Buffet stories {\bng Er ktha. taedr dam AnuJayii sair/bhs pRbha{I}eDr bYapar Ta sbar ethek Aalada ker erekhech. Aakr/Shniiy bYapar Hec/ch Era 550 Taka (laNJ/c) EbNNG 600 Taka (iDnar) E la{I}bh kabab, sasilk, kRabs, icekn kabab sabimT ker. blet egel} buffet stories {\bng {I} pRthm Jara imrpurbaisr bhaela bueph kha{O}yar du{h}kh emacn kerech. 
taedr eJsb khabar Aamar kaech bhaela lagar mt ichl
1. ibph erjala 2. icekn U{I}NNGs 3. kr/N sYup 4. icekn iTk/ka kabab 5. ephLimNNG icekn 6. gair/lk nan 7. c{O}imn 8. icekn sashilk 9. cekaelT ekk 10. puiDNNG Eguelaek Aaim mas/T TRa{I} krar saejs/T krb. eDjar/T Aa{I}eTm es eJn pura{I} Ummmm. Jara cekaelT epRim taedr jnY pura IId IId men Heb. cekaelT ekk Er saeth epRm ker{O} Aapin ebRk Aap kret parebn na kkhn {O}. Jara bhaela maenr ikchu buephet ekhet can tara AbshY{I}} buffet stories {\bng E eJey edkhet paern pRa{I}s}- lunch item and Unlimited drinks @550 {\bng Taka eriTNNG-8}.{\bng 99/10 ibeHibhyar-9}.{\bng 5/10} \\(When you are in Dhaka's capital Mirpur and can not go to Dhanmondi for the famous Buffet stories, what can you do? I was talking about that extraordinary buffet place, The Buffet Stories. Their service stands out according to their price. The attraction is, they serve live kababs, saslik, crabs, and chicken kababs for 550Tk (lunch) and 600Tk (dinner). To say, Buffet Stories has been a solace for those who reside in Mirpur and love good buffets. Here are the foods I enjoyed the most: 1. Beef rezala 2. Chicken wings 3. Corn soup 4. Chicken 5. tikka kabab 6. Flaming chicken 7. Garlic naan 8. Chowmein 9. Chicken saslik 10. Chocolate cake Pudding. I suggest you must try these. The dessert items are simply divine, especially for chocolate lovers, it feels like Eid every time. You will not be able to resist breaking your diet with their chocolate cake. Those who want to eat at a good quality buffet should definitely go to Buffet Stories. Price: Lunch item and unlimited drinks @ 550Tk Rating: 8.99/10 Behavior: 9.5/10)} \end{tabular} & 5                 \\ \hline
\end{tabular}
\end{table}

\section{Conversion with LLM -- Llama 2 (70B)} \label{secA8}
Table \ref{tab:converted-textsllama} shows the same instances as in \ref{tab:converted-texts}, but with the converted text using Llama 2 (70B). It's evident that Llama 2 (70B) is unable to generate meaningful conversion when the text is long. In the third case, it was able to promptly translate the sentence because there are only two English words mixed into Bengali, but it failed to perform in the second example because the number of English terms is spread over the text. Llama 2 attempted to convert the restaurant's name but failed, significantly affecting the sense of the statement instead of having 70 billion parameters. In addition, punctuation marks are almost non-existent. In the first scenario, the LLM experienced a hallucination in which various groups of words were repeatedly produced because it was unable to comprehend the transliterated Bengali words.

\begin{table}[h]
\caption{Few example instances before and after applying the text conversion pipeline using Llama 2}
\label{tab:converted-textsllama}
%\resizebox{\textwidth}{!}{
\begin{tabular}{c|c}
\hline
\textbf{Original Text}& \textbf{Converted Text (Llama 2)}         \\ \hline
\begin{tabular}[c]{@{}c@{}}\parbox[t]{3in}{Best pizza offer for Mirpur peoples Buy 1 Get 1 free 1.Mashroom lovers pizza.2.Meat lovers pizza. Choto bhaiyer jsc xam sesh kore gelm n e offer ta kheye ashlam onk tasty chilo n tader bebohar o valo chilo.. mane ak kothay paisa osul Taste:9/10 Price: 330tk Location: The Hub Restaurant (60 feet) — at The Hub Restaurant}\end{tabular}            & \begin{tabular}[c]{@{}c@{}}\parbox[t]{3in}{\bng Aapin imrpur pirbar elakael EkiT iprha Aphar paebn? Aapin E{I} AphariT inet Aamaeta Aapnar Aaechl thaka Aaechl Aamaeta Aapnar Aaechl thaka Aamaeta Aapnar Aaechl thaka Aamaeta Aapnar Aaechl thaka}\end{tabular}                            
\\ \hline
\begin{tabular}[c]{@{}c@{}}\parbox[t]{3in}{Cafe Famous Wari {\bng pRay ebsh keykidn Aaeg igeyichlam.... Eedr bar/gar gula bhaela laeg Aamar. ekan jaygar saeth tulna krb na  teb bhaela laeg Aamar sbsmy icekn bar/gar TRa{I} kra H{I}es sb khaen ephmaes egel{O} E{I}Ta{I} TRa{I} kra Hy Ebar Jas/T EkTu ecNJ/j krlam ibph Ta iney ibph ebkn-350 Taka iDRNNGk/s-40 Taka ker ephRNJ/c phRa{I}s-100 Taka \#elaekshn: {O}yarii} (yellow {\bng Er Uperr tlay) \#Aa{I}eTm:} Beef n bacon \bng \#dam: 350 Taka} \end{tabular} & \begin{tabular}[c]{@{}c@{}}\parbox[t]{3in}{{\bng Aapin E{I} Anusar Aaeshpaesh EkiT khabar Aa{I}eTm AidhSh/Th ker chairh Aamar sb khaen ephmaes egel{O} E{I}Ta{I} TRa{I} kra Hy Ebar Jas/T EkTu eceNJ/j krlam ibph Ta iney ibph \& ebkn-350 Taka iDRNNGk/s-40 Taka ker ephRNJ/c phRa{I}s-100 Taka \#elaekshn: {O}yarii} (yellow {\bng Er Uperr tlay) \#Aa{I}eTm:} Beef n bacon \bng \#dam: 350 Taka}\end{tabular} \\ \hline
\begin{tabular}[c]{@{}c@{}}\parbox[t]{3in}{Berger Xpress \bng E bn/dhuedr saeth bar/gar ekhet egichlam. per inejr bar/gar inej{I} pa{I}lam na}\end{tabular}            & \begin{tabular}[c]{@{}c@{}}\parbox[t]{3in}{\bng bar/gar Ek/sepRs E bn/dhuedr saeth bar/gar ekhet egichlam per inejr bar/gar inej{I} pa{I}lam na}\end{tabular}\\\hline
\end{tabular}
%}
\end{table}

% \textbf{Appendix}
\section{Algorithm of the Proposed Ensemble Model }\label{secA4}
The process of utilizing the weighted ensemble technique is outlined in Algorithm \ref{algo:ensemble}. Firstly, we multiply each class's probability (softmax) by the two corresponding f1 scores, and then we sum them up. Next, we divide the total of the prior f1 scores of the models for each class to normalize the individual scores for each class. Lastly, we find the final forecast by identifying the highest predicted scores in all classes.

\begin{algorithm}[]
\caption{Algorithm for the proposed weighted ensemble model}
\label{algo:ensemble}
\begin{algorithmic}[1]

    \State $m \leftarrow no.\_of\_test\_samples$
    \State $n \leftarrow no.\_of\_models$ 
    \State $c \leftarrow no.\_of\_class$ 
    \State $p \leftarrow \left[m \right]\left[n \right]\left[c \right]$ \Comment{softmax probabilities}
    \State $f1 \leftarrow \left[n\right]\left[c\right]$ \Comment{f1 scores of individual classes}
    % \State $f1_1 \leftarrow \left[n\right]$ \Comment{f1 scores of non-fake class}
    \State $sum = \left[m\right]\left[c\right]$ \Comment{weighted sum of individual classes}
    % \State $sum_1 = \left[ \right]$ \Comment{weighted sum of non-fake class }
    \newline
    \For{$i\epsilon(1,m)$ }
        \For{$j\epsilon(1,n)$}
            \For{$k\epsilon(1,c)$}
                % \If{$p\left[i\right]\left[j\right]$ is the probability of fake class}
                    \State $sum\left[i\right]\left[k\right] += p\left[i\right]\left[j\right]\left[k\right] * f1\left[j\right]\left[k\right]$ 
                % \Else
                    % \State $sum_1\left[i\right] += p\left[j\right]\left[i\right]_{non-fake} * f1_1[j]$
                \State $k+=1$
                % \EndIf
            \EndFor
            \State $j+=1$
        \EndFor
        \State $i+=1$
    \EndFor\\
\State $n\_sum = \left[c\right]$
% \State $w\_sum_1 = 0$
\For{$j\epsilon(1,n)$}
    \For{$k\epsilon(1,c)$}
        \State $n\_sum[k] +=  f1[j][k]$
        % \State $w\_sum_1 +=  f1_1[j]$ 
        \State $k+=1$
    \EndFor
    \State $j+=1$
\EndFor
\State $pred = sum/n\_sum$
% \State $P_{non-fake} = sum_1/w\_sum_1$
\State $Output = argmax(pred)$
% \State $Output = argmax(P_{fake},P_{non-fake})$
\end{algorithmic}
\end{algorithm}

\section{Details of Evaluation Metrics}\label{secA3}
In our study, we used several metrics to evaluate the performance of our trained models. These metrics include Precision, Recall, F1-score, ROC-AUC score and MCC score. In this section, we provide a brief explanation of each metric. It is worth noting that all metrics are described using \textit{fake} as the positive class.\\
\textbf{\underline{Precision (P)}:} Precision measures the proportion of true positives among all positive predictions. (How many predicted fake reviews are actually fake?)\\
\begin{equation}
\text { P }=\frac{tp}{tp + fp}
\end{equation}
\textbf{\underline{Recall (R)}:} Recall measures the proportion of true positives among all actual positives. (How many fake reviews are correctly identified?)\\
\begin{equation}
\text { R }=\frac{tp}{tp + fn}
\end{equation}
\textbf{\underline{F1-score (F1)}:} The F1-score calculates the harmonic mean of precision and recall while taking into account the significance of both the precision and recall values. The weighted F1 score is calculated by taking into account each of the fake and non-fake classes as positive separately and averaging out the results. It takes into account the quantity of samples from each class.\\
\begin{equation}
F1= \frac {2 {\text *} ( { P } {\text *}  { R })}{( { P }+  { R })}
\end{equation}
\textbf{\underline{ROC-AUC score}:} The ROC-AUC score of a model is an indication of its ability to distinguish between positive and negative classes. A score less than 0.5 indicates poor performance of the classifier. On the other hand, if the classifier detects more True positives and True negatives than False negatives and False positives, its score is greater than 0.5. A good ROC-AUC score ranges between 0.7 and 0.8.\\
\begin{equation}
    ROC{\text -}AUC =\frac{1+TPR-FPR}{2}
\end{equation}
The likelihood of a false alarm is known as FPR and true positive rate (TPR) basically denotes the recall score.\\
\textbf{\underline{MCC score}:} The Matthews correlation coefficient (MCC) is a more robust statistical measure, and it only yields a high value when the prediction is accurate in all four scenarios (tp, tn, fp, fn). Even if one of the classes is heavily over-represented or under-represented, a value close to 1 indicates that both classes have been correctly predicted..\\
\begin{equation}
\centering
M C C=\frac{t n {\text *} t p {\text-}f n {\text *} f p}{\sqrt{(t p+f p)(t p+f n)(t n+f p)(t n+f n)}}
\end{equation}
\section{Hyperparameter Settings}\label{secA2} The configuration of hyperparameters used in different models can be found in Table \ref{tab:hyper}.  It is worth noting that due to memory constraints, the input sequence length for all models, except \textit{sahajBERT} and \textit{BanglaBERT Large} was limited to 512. However, the input sequence length was 256 for these two models.

\begin{table}[H]
\centering
\caption{Hyper-parameter settings used for deep learning and transformer models on all three approaches (Approach-1, 2, 3). Here, value 0 in \textit{`No. of Aug'} column denotes no augmentation}
\label{tab:hyper}
\begin{tabular}{ccccc}
\hline
\multirow{3}{*}{\textbf{Model}}                      & \multirow{3}{*}{\textbf{\begin{tabular}[c]{@{}c@{}}No. of  Aug\end{tabular}}} & \multirow{3}{*}{\textbf{Epoch}} & \multirow{3}{*}{\textbf{Batch Size}} & \multirow{3}{*}{\textbf{\begin{tabular}[c]{@{}c@{}}Learning\\ Rate\end{tabular}}}  \\
             &             & &      & \\
                     & &      & &     \\ \hline
CNN & 0-4           & 15             & 128                 & 5e\textasciicircum{}4                         \\  
             \hline
BiLSTM                             & 0-4           & 15             & 256                 & 5e\textasciicircum{}4  \\ 
             \hline
\multirow{5}{*}{\begin{tabular}[c]{@{}c@{}}CNN \\ BiLSTM\end{tabular}}                      & 0           & 20& \multirow{5}{*}{128}                 & \multirow{5}{*}{5e\textasciicircum{}4}  \\ 
             & 1           & 15&      & \\ 
             & 2           & 15&      & \\ 
             & 3           & 15&      & \\ 
             & 4           & 15&      & \\ \hline
\multirow{5}{*}{\begin{tabular}[c]{@{}c@{}}CNN \\ BiLSTM \\ with \\ Attention\end{tabular}} & 0           & 20& \multirow{5}{*}{256}                 & \multirow{5}{*}{5e\textasciicircum{}4}  \\ 
             & 1           & 15&      & \\ 
             & 2           & 10&      & \\ 
             & 3           & 10&      & \\ 
             & 4           & 15&      & \\ \hline
\multirow{5}{*}{\begin{tabular}[c]{@{}c@{}}Bangla \\ BERT\end{tabular}}                      & 0           & 15& \multirow{5}{*}{4}                   & \multirow{5}{*}{2e\textasciicircum{}-7} \\ 
             & 1           & 7&      & \\ 
             & 2           & 6&      & \\ 
             & 3           & 6&      & \\ 
             & 4           & 6&      & \\ \hline
\multirow{2}{*}{\begin{tabular}[c]{@{}c@{}}Bangla BERT \\ Genertor\end{tabular}}          & \multirow{2}{*}{0-4}           & \multirow{2}{*}{10}             & \multirow{2}{*}{8}                   & \multirow{2}{*}{5e\textasciicircum{}-5} \\ 
             &           & &      & \\ 
             \hline
\multirow{5}{*}{\begin{tabular}[c]{@{}c@{}}Bangla \\ BERT\\ Base\end{tabular}}               & 0           & 10& \multirow{5}{*}{8}                   & \multirow{5}{*}{5e\textasciicircum{}-5} \\ 
             & 1           & 8&      & \\ 
             & 2           & 5&      & \\ 
             & 3           & 5&      & \\ 
             & 4           & 5&      & \\ \hline
\multirow{2}{*}{\begin{tabular}[c]{@{}c@{}}sahaj \\ BERT\end{tabular}}                       & \multirow{2}{*}{0-4}           & \multirow{2}{*}{20}             & \multirow{2}{*}{4}                   & \multirow{2}{*}{2e\textasciicircum{}-7}                         \\ 
             &          & &      & \\  \hline
\multirow{2}{*}{\begin{tabular}[c]{@{}c@{}}Bangla BERT\\ Large\end{tabular}}              & \multirow{2}{*}{0-4}           & \multirow{2}{*}{20}             & \multirow{2}{*}{4}                   & \multirow{2}{*}{2e\textasciicircum{}-7}                     \\ 
             & & &      & \\ \hline
\end{tabular}
\end{table}

\section{Performance of Models for Approach 2 and 3 }\label{secA1}
This section shows the detailed findings of approaches 2 and 3 for both stand-alone and ensemble models. In Tables \ref{tab:appensemblenlpaug} and \ref{tab:appensemblesagar}, the ensemble techniques provide the outcomes of employing 0 to 3 augmented fake data per review for Approach-2 and Approach-3, respectively.

\subsection{Deep Learning and Transformer Models:} 
The detailed results of the standalone models for approach-2 and approach-3 are shown in Table \ref{tab:tradnlpaug} and \ref{tab:tradsagar} respectively.  The results were previously presented in Fig. \ref{fig:compchart} only for weighted-F1 scores in terms of four augmented samples along with the fake one for both the approaches.\\

\noindent\textbf{• \underline{Approach-2:}}
% \subsubsection{Performance (Augmentation with nlpaug)}
The outcomes obtained utilizing the \textit{nlpaug} augmentation technique are shown in Table \ref{tab:tradnlpaug}. The first two DL methods: CNN and BiLSTM, display striking WF1 values of 0.974 and 0.969, respectively with MCC scores of 0.948 and 0.939 for the case of 4 augmented samples of fake data.
\noindent We also observe consistent improvement in performance for both models across all metrics. It is obvious that CNN under performed in the instances of two augmentations while maintaining continuity in the other scenarios. The CNN and BiLSTM hybrid models exhibit similar performance to the standalone models. The best WF1 and MCC scores (0.978 and 0.957) for four distinct augmentations were provided by the model built with CNN-BiLSTM with attention layer. It should be noted that the precision and recall values for both the fake and non-fake classes indicate that the predictions for the two classes are not always the same for this model. As a result, due to the model's bias against a specific class, it cannot be said that this hybrid model is the most effective one. The performance of transformers is not as good as it was in the no augmentation scenario. Transformers exhibit outcomes that are similar to those of DL approaches in the presence of augmentation. \textit{BanglaBERT} (BB) in terms of WF1 score (0.981) for four different augmentations performes best. ROC-AUC (0.981) and MCC (0.961) scores also supports the claim. BBG and BBB demonstrate comparable results with a WF1 score of 0.963 which are encouraging. Furthermore, the capacity of the models to accurately classify fake reviews is demonstrated by the continuous improvement in performance with increased numbers of augmented samples for each real data. SB is the lowest performing model as always having an MCC of only 0.713 for four augmentation. It could be due to the short sequence length (256) and low number of parameters (18M). In terms of fake class, the result of BBL is intriguing because it has the maximum precision of 0.984. On the other hand, the recall score (0.808) is quite low indicating the model's bias towards the fake classes. This might be due to the short sequence length. 
 
\begin{table}[H]
\caption{Performance comparison among individual models on \textit{nlpaug} generated augmentations (Approach - 2)}
\label{tab:tradnlpaug}
%\resizebox{\textwidth}{!}{
\begin{tabular}{cc|ccc|ccc|ccc}
\hline
\textbf{}       & \textbf{}                           & \multicolumn{3}{c|}{\textbf{Fake}}        & \multicolumn{3}{c|}{\textbf{Non-Fake}}     & \textbf{}    & \textbf{}          & \textbf{}    \\ \hline
\textbf{Model}  & \multicolumn{1}{c|}{\begin{tabular}[c]{@{}c@{}}{\rotatebox[origin=c]{90}{ \textbf{Aug} }}\end{tabular}} & \multicolumn{1}{c}{\textbf{P}} & \multicolumn{1}{c}{\textbf{R}} & \textbf{F1} & \multicolumn{1}{c}{\textbf{P}} & \multicolumn{1}{c}{\textbf{R}} & \textbf{F1} & \multicolumn{1}{l}{\textbf{WF1}}  & \multicolumn{1}{l}{\begin{tabular}[c]{@{}c@{}}\textbf{ROC}\\ \textbf{-AUC}\end{tabular}} & \textbf{MCC}  \\ \hline

\multirow{4}{*}{CNN} & 1  & \multicolumn{1}{c}{0.940}      & \multicolumn{1}{c}{0.884}      & 0.912       & \multicolumn{1}{c}{0.891}      & \multicolumn{1}{c}{0.944}      & 0.917       & 0.914           & 0.955            & 0.830        \\  
     & 2  & \multicolumn{1}{c}{0.872}      & \multicolumn{1}{c}{0.963}      & 0.915       & \multicolumn{1}{c}{0.958}      & \multicolumn{1}{c}{0.858}      & 0.906       & 0.910                & 0.971            & 0.825        \\  
     & 3  & \multicolumn{1}{c}{0.975}      & \multicolumn{1}{c}{0.953}      & 0.964       & \multicolumn{1}{c}{0.954}      & \multicolumn{1}{c}{0.976}      & 0.965       & 0.965               & 0.991            & 0.929        \\  
     & 4  & \multicolumn{1}{c}{0.962}      & \multicolumn{1}{c}{0.987}      & 0.974       & \multicolumn{1}{c}{0.986}      & \multicolumn{1}{c}{0.961}      & 0.974       & \textbf{0.974}               & 0.997            & \textbf{0.948}        \\ \hline
\multirow{4}{*}{BiLSTM}& 1  & \multicolumn{1}{c}{0.891}      & \multicolumn{1}{c}{0.910}      & 0.900       & \multicolumn{1}{c}{0.908}      & \multicolumn{1}{c}{0.888}      & 0.898       & 0.899             & 0.950            & 0.799        \\  
     & 2  & \multicolumn{1}{c}{0.939}      & \multicolumn{1}{c}{0.913}      & 0.926       & \multicolumn{1}{c}{0.915}      & \multicolumn{1}{c}{0.940}      & 0.928       & 0.927            & 0.963            & 0.854        \\  
     & 3  & \multicolumn{1}{c}{0.944}      & \multicolumn{1}{c}{0.974}      & 0.959       & \multicolumn{1}{c}{0.973}      & \multicolumn{1}{c}{0.942}      & 0.957       & 0.958               & 0.990            & 0.917        \\  
     & 4  & \multicolumn{1}{c}{0.974}      & \multicolumn{1}{c}{0.964}      & 0.969       & \multicolumn{1}{c}{0.965}      & \multicolumn{1}{c}{0.975}      & 0.970       & \textbf{0.969}              & 0.995            & \textbf{0.939 }       \\ \hline
\multirow{4}{*}{\begin{tabular}[c]{@{}c@{}}CNN   \\ BiLSTM\end{tabular}}                   & 1  & \multicolumn{1}{c}{0.912}      & \multicolumn{1}{c}{0.892}      & 0.902       & \multicolumn{1}{c}{0.894}      & \multicolumn{1}{c}{0.914}      & 0.904       & 0.903              & 0.949            & 0.806        \\  
     & 2  & \multicolumn{1}{c}{0.938}      & \multicolumn{1}{c}{0.905}      & 0.922       & \multicolumn{1}{c}{0.909}      & \multicolumn{1}{c}{0.940}      & 0.924       & 0.923               & 0.971            & 0.846        \\  
     & 3  & \multicolumn{1}{c}{0.955}      & \multicolumn{1}{c}{0.955}      & 0.955       & \multicolumn{1}{c}{0.955}      & \multicolumn{1}{c}{0.955}      & 0.955       & 0.955              & 0.985            & 0.910        \\  
     & 4  & \multicolumn{1}{c}{0.970}      & \multicolumn{1}{c}{0.979}      & 0.975       & \multicolumn{1}{c}{0.979}      & \multicolumn{1}{c}{0.970}      & 0.975       & 0.975             & 0.996            & 0.949        \\ \hline
\multirow{4}{*}{\begin{tabular}[c]{@{}c@{}}CNN\\ BiLSTM\\ with\\ Attention\end{tabular}} & 1  & \multicolumn{1}{c}{0.926}      & \multicolumn{1}{c}{0.884}      & 0.905       & \multicolumn{1}{c}{0.889}      & \multicolumn{1}{c}{0.929}      & 0.909       & 0.907              & 0.954            & 0.814        \\  
     & 2  & \multicolumn{1}{c}{0.961}      & \multicolumn{1}{c}{0.920}      & 0.940       & \multicolumn{1}{c}{0.924}      & \multicolumn{1}{c}{0.963}      & 0.943       & 0.942                & 0.978            & 0.884        \\  
     & 3  & \multicolumn{1}{c}{0.972}      & \multicolumn{1}{c}{0.916}      & 0.943       & \multicolumn{1}{c}{0.921}      & \multicolumn{1}{c}{0.974}      & 0.947       & 0.945               & 0.985            & 0.891        \\  
     & 4  & \multicolumn{1}{c}{0.968}      & \multicolumn{1}{c}{0.990}      & 0.979       & \multicolumn{1}{c}{0.989}      & \multicolumn{1}{c}{0.967}      & 0.978       & \textbf{0.978}               & 0.998            & \textbf{0.957 }       \\ \hline
\multirow{4}{*}{\begin{tabular}[c]{@{}c@{}}Bangla\\BERT\\(BB)\end{tabular}}                           & 1  & \multicolumn{1}{c}{0.951}      & \multicolumn{1}{c}{0.933}      & 0.942       & \multicolumn{1}{c}{0.934}      & \multicolumn{1}{c}{0.951}      & 0.943       & 0.942          & 0.942            & 0.884        \\  
     & 2  & \multicolumn{1}{c}{0.953}      & \multicolumn{1}{c}{0.960}      & 0.957       & \multicolumn{1}{c}{0.960}      & \multicolumn{1}{c}{0.953}      & 0.956       & 0.956             & 0.956            & 0.913        \\  
     & 3  & \multicolumn{1}{c}{0.972}      & \multicolumn{1}{c}{0.978}      & 0.975       & \multicolumn{1}{c}{0.977}      & \multicolumn{1}{c}{0.972}      & 0.975       & 0.975              & 0.975            & 0.950        \\  
     & 4  & \multicolumn{1}{c}{0.976}      & \multicolumn{1}{c}{0.985}      & 0.981       & \multicolumn{1}{c}{0.985}      & \multicolumn{1}{c}{0.976}      & 0.980       & \textbf{0.981}              & \textbf{0.981}            & \textbf{0.961 }       \\ \hline
\multirow{4}{*}{\begin{tabular}[c]{@{}c@{}}Bangla\\BERT \\ Generator\\(BBG)\end{tabular}}            & 1  & \multicolumn{1}{c}{0.868}      & \multicolumn{1}{c}{0.910}      & 0.889       & \multicolumn{1}{c}{0.906}      & \multicolumn{1}{c}{0.862}      & 0.883       & 0.886           & 0.886            & 0.773        \\  
     & 2  & \multicolumn{1}{c}{0.906}      & \multicolumn{1}{c}{0.933}      & 0.919       & \multicolumn{1}{c}{0.931}      & \multicolumn{1}{c}{0.903}      & 0.917       & 0.918               & 0.918            & 0.836        \\  
     & 3  & \multicolumn{1}{c}{0.920}      & \multicolumn{1}{c}{0.970}      & 0.945       & \multicolumn{1}{c}{0.968}      & \multicolumn{1}{c}{0.916}      & 0.942       & 0.943              & 0.943            & 0.887        \\  
     & 4  & \multicolumn{1}{c}{0.944}      & \multicolumn{1}{c}{0.984}      & 0.963       & \multicolumn{1}{c}{0.983}      & \multicolumn{1}{c}{0.942}      & 0.962       & \textbf{0.963}              & 0.963            & 0.926        \\ \hline
\multirow{4}{*}{\begin{tabular}[c]{@{}c@{}}Bangla \\ BERT\\ Base\\ (BBB)\end{tabular}}              & 1  & \multicolumn{1}{c}{0.888}      & \multicolumn{1}{c}{0.892}      & 0.890       & \multicolumn{1}{c}{0.891}      & \multicolumn{1}{c}{0.888}      & 0.890       & 0.890              & 0.890            & 0.780        \\  
     & 2  & \multicolumn{1}{c}{0.911}      & \multicolumn{1}{c}{0.940}      & 0.925       & \multicolumn{1}{c}{0.938}      & \multicolumn{1}{c}{0.908}      & 0.923       & 0.924               & 0.924            & 0.849        \\  
     & 3  & \multicolumn{1}{c}{0.922}      & \multicolumn{1}{c}{0.966}      & 0.944       & \multicolumn{1}{c}{0.965}      & \multicolumn{1}{c}{0.918}      & 0.941       & 0.942              & 0.942            & 0.885        \\  
     & 4  & \multicolumn{1}{c}{0.954}      & \multicolumn{1}{c}{0.973}      & 0.964       & \multicolumn{1}{c}{0.973}      & \multicolumn{1}{c}{0.954}      & 0.963       & \textbf{0.963}               & 0.963            & 0.927        \\ \hline
\multirow{4}{*}{\begin{tabular}[c]{@{}c@{}}sahaj\\BERT\\ (SB)\end{tabular}}                             & 1  & \multicolumn{1}{c}{0.726}      & \multicolumn{1}{c}{0.750}      & 0.738       & \multicolumn{1}{c}{0.741}      & \multicolumn{1}{c}{0.716}      & 0.729       & 0.733               & 0.733            & 0.467        \\  
     & 2  & \multicolumn{1}{c}{0.819}      & \multicolumn{1}{c}{0.823}      & 0.821       & \multicolumn{1}{c}{0.823}      & \multicolumn{1}{c}{0.818}      & 0.820       & 0.821               & 0.821            & 0.642        \\  
     & 3  & \multicolumn{1}{c}{0.825}      & \multicolumn{1}{c}{0.847}      & 0.836       & \multicolumn{1}{c}{0.843}      & \multicolumn{1}{c}{0.821}      & 0.832       & 0.834             & 0.834            & 0.668        \\  
     & 4  & \multicolumn{1}{c}{0.840}      & \multicolumn{1}{c}{0.880}      & 0.859       & \multicolumn{1}{c}{0.874}      & \multicolumn{1}{c}{0.833}      & 0.853       & 0.856             & 0.856            & \textbf{0.713}        \\ \hline
\multirow{4}{*}{\begin{tabular}[c]{@{}c@{}}Bangla\\BERT\\ Large\\ (BBL)\end{tabular}}                 & 1  & \multicolumn{1}{c}{0.939}      & \multicolumn{1}{c}{0.743}      & 0.829       & \multicolumn{1}{c}{0.787}      & \multicolumn{1}{c}{0.951}      & 0.861       & 0.845                & 0.847            & 0.710        \\  
     & 2  & \multicolumn{1}{c}{0.966}      & \multicolumn{1}{c}{0.781}      & 0.864       & \multicolumn{1}{c}{0.816}      & \multicolumn{1}{c}{0.973}      & 0.888       & 0.876               & 0.877            & 0.768        \\  
     & 3  & \multicolumn{1}{c}{0.957}      & \multicolumn{1}{c}{0.869}      & 0.911       & \multicolumn{1}{c}{0.880}      & \multicolumn{1}{c}{0.961}      & 0.919       & 0.915            & 0.915            & 0.834        \\  
     & 4  & \multicolumn{1}{c}{\textbf{0.984}}      & \multicolumn{1}{c}{\textbf{0.808}}      & 0.887       & \multicolumn{1}{c}{0.838}      & \multicolumn{1}{c}{0.987}      & 0.906       & 0.897                & 0.897            & 0.808        \\ \hline
\end{tabular}
%}
\end{table}

\begin{table}[H]
\caption{Performance comparison among individual models on \textit{bnaug} generated augmentations (Approach - 3)}
\label{tab:tradsagar}
%\resizebox{\textwidth}{!}{
\begin{tabular}{cc|ccc|ccc|ccc}
\hline
\textbf{}       & \textbf{}                           & \multicolumn{3}{c|}{\textbf{Fake}}        & \multicolumn{3}{c|}{\textbf{Non-Fake}}     & \textbf{}    & \textbf{}            & \textbf{}    \\ \hline
\textbf{Model}  & \multicolumn{1}{c|}{\begin{tabular}[c]{@{}c@{}}{\rotatebox[origin=c]{90}{ \textbf{Aug} }}\end{tabular}} & \multicolumn{1}{c}{\textbf{P}} & \multicolumn{1}{c}{\textbf{R}} & \textbf{F1} & \multicolumn{1}{c}{\textbf{P}} & \multicolumn{1}{c}{\textbf{R}} & \textbf{F1} & \multicolumn{1}{l}{\textbf{WF1}}  & \multicolumn{1}{l}{\begin{tabular}[c]{@{}c@{}}\textbf{ROC}\\ \textbf{-AUC}\end{tabular}} & \textbf{MCC}  \\ \hline
\multirow{4}{*}{CNN}                     & 1  & \multicolumn{1}{c}{0.927}      & \multicolumn{1}{c}{0.806}      & 0.862       & \multicolumn{1}{c}{0.828}      & \multicolumn{1}{c}{0.937}      & 0.879       & 0.871               & 0.944            & 0.749        \\ 
   & 2  & \multicolumn{1}{c}{0.977}      & \multicolumn{1}{c}{0.851}      & 0.910       & \multicolumn{1}{c}{0.868}      & \multicolumn{1}{c}{0.980}      & 0.921       & 0.915               & 0.951            & 0.838        \\ 
   & 3  & \multicolumn{1}{c}{0.950}      & \multicolumn{1}{c}{0.896}      & 0.922       & \multicolumn{1}{c}{0.901}      & \multicolumn{1}{c}{0.953}      & 0.927       & 0.924                & 0.982            & 0.850        \\ 
   & 4  & \multicolumn{1}{c}{0.888}      & \multicolumn{1}{c}{0.963}      & 0.924       & \multicolumn{1}{c}{0.959}      & \multicolumn{1}{c}{0.879}      & 0.917       & \textbf{0.921 }               & 0.981            & 0.845        \\ \hline
\multirow{4}{*}{BiLSTM}                 & 1  & \multicolumn{1}{c}{0.885}      & \multicolumn{1}{c}{0.802}      & 0.841       & \multicolumn{1}{c}{0.819}      & \multicolumn{1}{c}{0.896}      & 0.856       & 0.849            & 0.935            & 0.701        \\ 
   & 2  & \multicolumn{1}{c}{0.930}      & \multicolumn{1}{c}{0.893}      & 0.911       & \multicolumn{1}{c}{0.897}      & \multicolumn{1}{c}{0.933}      & 0.915       & 0.913              & 0.973            & 0.827        \\ 
   & 3  & \multicolumn{1}{c}{0.961}      & \multicolumn{1}{c}{0.866}      & 0.911       & \multicolumn{1}{c}{0.878}      & \multicolumn{1}{c}{0.965}      & 0.919       & 0.915              & 0.976            & 0.834        \\ 
   & 4  & \multicolumn{1}{c}{0.957}      & \multicolumn{1}{c}{0.894}      & 0.924       & \multicolumn{1}{c}{0.900}      & \multicolumn{1}{c}{0.960}      & 0.929       & \textbf{0.927}                & 0.970            & 0.855        \\ \hline
\multirow{4}{*}{\begin{tabular}[c]{@{}c@{}}CNN   \\ BiLSTM\end{tabular}}             & 1  & \multicolumn{1}{c}{0.905}      & \multicolumn{1}{c}{0.780}      & 0.838       & \multicolumn{1}{c}{0.807}      & \multicolumn{1}{c}{0.918}      & 0.859       & 0.848            & 0.934            & 0.705        \\ 
   & 2  & \multicolumn{1}{c}{0.916}      & \multicolumn{1}{c}{0.863}      & 0.889       & \multicolumn{1}{c}{0.871}      & \multicolumn{1}{c}{0.920}      & 0.895       & 0.892             & 0.963            & 0.785        \\ 
   & 3  & \multicolumn{1}{c}{0.911}      & \multicolumn{1}{c}{0.916}      & 0.913       & \multicolumn{1}{c}{0.916}      & \multicolumn{1}{c}{0.910}      & 0.913       & 0.913                & 0.975            & 0.827        \\ 
   & 4  & \multicolumn{1}{c}{0.948}      & \multicolumn{1}{c}{0.931}      & 0.940       & \multicolumn{1}{c}{0.932}      & \multicolumn{1}{c}{0.949}      & 0.941       & \textbf{0.940}                & 0.982            & \textbf{0.881 }       \\ \hline
\multirow{4}{*}{\begin{tabular}[c]{@{}c@{}}CNN\\ BiLSTM\\ with\\ Attention\end{tabular}} & 1  & \multicolumn{1}{c}{0.887}      & \multicolumn{1}{c}{0.817}      & 0.850       & \multicolumn{1}{c}{0.830}      & \multicolumn{1}{c}{0.896}      & 0.862       & 0.856              & 0.941            & 0.715        \\ 
   & 2  & \multicolumn{1}{c}{0.950}      & \multicolumn{1}{c}{0.893}      & 0.921       & \multicolumn{1}{c}{0.899}      & \multicolumn{1}{c}{0.953}      & 0.925       & 0.923             & 0.975            & 0.847        \\ 
   & 3  & \multicolumn{1}{c}{0.879}      & \multicolumn{1}{c}{0.959}      & 0.917       & \multicolumn{1}{c}{0.955}      & \multicolumn{1}{c}{0.868}      & 0.909       & 0.913                & 0.973            & 0.830        \\ 
   & 4  & \multicolumn{1}{c}{0.953}      & \multicolumn{1}{c}{0.909}      & 0.930       & \multicolumn{1}{c}{0.913}      & \multicolumn{1}{c}{0.955}      & 0.934       & 0.932        & \textbf{0.979}            & 0.865        \\ \hline
\multirow{4}{*}{\begin{tabular}[c]{@{}c@{}}Bangla\\BERT\\(BB)\end{tabular}}                   & 1  & \multicolumn{1}{c}{0.917}      & \multicolumn{1}{c}{0.910}      & 0.914       & \multicolumn{1}{c}{0.911}      & \multicolumn{1}{c}{0.918}      & 0.914       & 0.914              & 0.914            & 0.828        \\ 
   & 2  & \multicolumn{1}{c}{0.947}      & \multicolumn{1}{c}{0.925}      & 0.936       & \multicolumn{1}{c}{0.927}      & \multicolumn{1}{c}{0.948}      & 0.937       & 0.937             & 0.937            & 0.873        \\ 
   & 3  & \multicolumn{1}{c}{0.962}      & \multicolumn{1}{c}{0.942}      & 0.952       & \multicolumn{1}{c}{0.943}      & \multicolumn{1}{c}{0.963}      & 0.953       & 0.952             & 0.952            & 0.905        \\ 
   & 4  & \multicolumn{1}{c}{0.955}      & \multicolumn{1}{c}{0.930}      & 0.942       & \multicolumn{1}{c}{0.932}      & \multicolumn{1}{c}{0.957}      & 0.944       & \textbf{0.943}             & 0.943            & \textbf{0.887}        \\ \hline
\multirow{4}{*}{\begin{tabular}[c]{@{}c@{}}Bangla\\BERT\\Generator\\(BBG)\end{tabular}}          & 1  & \multicolumn{1}{c}{0.895}      & \multicolumn{1}{c}{0.867}      & 0.881       & \multicolumn{1}{c}{0.871}      & \multicolumn{1}{c}{0.898}      & 0.884       & 0.883              & 0.883            & 0.766        \\ 
   & 2  & \multicolumn{1}{c}{0.941}      & \multicolumn{1}{c}{0.910}      & 0.925       & \multicolumn{1}{c}{0.913}      & \multicolumn{1}{c}{0.943}      & 0.927       & 0.926               & 0.926            & 0.853        \\ 
   & 3  & \multicolumn{1}{c}{0.954}      & \multicolumn{1}{c}{0.925}      & 0.939       & \multicolumn{1}{c}{0.928}      & \multicolumn{1}{c}{0.955}      & 0.941       & 0.940               & 0.940            & 0.881        \\ 
   & 4  & \multicolumn{1}{c}{0.955}      & \multicolumn{1}{c}{0.938}      & 0.947       & \multicolumn{1}{c}{0.940}      & \multicolumn{1}{c}{0.957}      & 0.948       & \textbf{0.947}              & 0.947            & \textbf{0.895 }       \\ \hline
\multirow{4}{*}{\begin{tabular}[c]{@{}c@{}}Bangla \\ BERT\\ Base\\(BBB)\end{tabular}}                   & 1  & \multicolumn{1}{c}{0.912}      & \multicolumn{1}{c}{0.884}      & 0.898       & \multicolumn{1}{c}{0.888}      & \multicolumn{1}{c}{0.914}      & 0.901       & 0.899                & 0.899            & 0.799        \\ 
   & 2  & \multicolumn{1}{c}{0.932}      & \multicolumn{1}{c}{0.923}      & 0.928       & \multicolumn{1}{c}{0.924}      & \multicolumn{1}{c}{0.933}      & 0.928       & 0.928                & 0.928            & 0.856        \\ 
   & 3  & \multicolumn{1}{c}{0.926}      & \multicolumn{1}{c}{0.909}      & 0.917       & \multicolumn{1}{c}{0.910}      & \multicolumn{1}{c}{0.927}      & 0.919       & 0.918               & 0.918            & 0.836        \\ 
   & 4  & \multicolumn{1}{c}{0.951}      & \multicolumn{1}{c}{0.928}      & 0.939       & \multicolumn{1}{c}{0.930}      & \multicolumn{1}{c}{0.952}      & 0.941       & \textbf{0.940}                & 0.940            & 0.880        \\ \hline
\multirow{4}{*}{\begin{tabular}[c]{@{}c@{}}sahaj \\ BERT\\(SB)\end{tabular}}               & 1  & \multicolumn{1}{c}{0.829}      & \multicolumn{1}{c}{0.799}      & 0.814       & \multicolumn{1}{c}{0.806}      & \multicolumn{1}{c}{0.836}      & 0.821       & 0.817                & 0.817            & 0.635        \\ 
   & 2  & \multicolumn{1}{c}{0.894}      & \multicolumn{1}{c}{0.841}      & 0.867       & \multicolumn{1}{c}{0.850}      & \multicolumn{1}{c}{0.900}      & 0.874       & 0.871              & 0.871            & 0.743        \\ 
   & 3  & \multicolumn{1}{c}{0.898}      & \multicolumn{1}{c}{0.851}      & 0.874       & \multicolumn{1}{c}{0.858}      & \multicolumn{1}{c}{0.903}      & 0.880       & 0.877              & 0.877            & 0.755        \\ 
   & 4  & \multicolumn{1}{c}{0.916}      & \multicolumn{1}{c}{0.882}      & 0.898       & \multicolumn{1}{c}{0.886}      & \multicolumn{1}{c}{0.919}      & 0.902       & \textbf{0.900}               & 0.900            & 0.801        \\ \hline
\multirow{4}{*}{\begin{tabular}[c]{@{}c@{}}Bangla \\ BERT\\ Large\\(BBL)\end{tabular}}            & 1  & \multicolumn{1}{c}{0.932}      & \multicolumn{1}{c}{0.813}      & 0.869       & \multicolumn{1}{c}{0.834}      & \multicolumn{1}{c}{0.940}      & 0.884       & 0.876             & 0.877            & 0.760        \\ 
   & 2  & \multicolumn{1}{c}{0.963}      & \multicolumn{1}{c}{0.841}      & 0.898       & \multicolumn{1}{c}{0.859}      & \multicolumn{1}{c}{0.968}      & 0.910       & 0.904             & 0.904            & 0.815        \\ 
   & 3  & \multicolumn{1}{c}{0.978}      & \multicolumn{1}{c}{0.826}      & 0.896       & \multicolumn{1}{c}{0.850}      & \multicolumn{1}{c}{0.981}      & 0.911       & 0.903             & 0.904            & 0.818        \\ 
   & 4  & \multicolumn{1}{c}{0.978}      & \multicolumn{1}{c}{0.876}      & 0.924       & \multicolumn{1}{c}{0.888}      & \multicolumn{1}{c}{0.981}      & 0.932       & \textbf{0.928}             & 0.928            & 0.861        \\ \hline
\end{tabular}
%}
\end{table}

\noindent\textbf{• \underline{Approach-3:}}
% \subsubsection{Performance (Augmentation with our proposed technique)}
Results from augmentation utilizing \textit{bnaug} technique did not perform as well as those from \textit{nlpaug} which can be seen in Table \ref{tab:tradsagar}. Performances have decreased by 2\%-4\% from the previous models of approach-2. The WF1 values obtained by CNN and BiLSTM, 0.921 and 0.927, are extremely close. With a WF1 of 0.94 and MCC of 0.881, which are 2.6\% and 3.6\% higher than CNN and BiLSTM respectively, the hybrid model with CNN-BiLSTM somewhat outperformed the individual models. In case of four distinct augmented samples, the hybrid model with attention layer displays a ROC-AUC score of 0.979 which is almost similar to other DL models. The results make it very evident that most DL algorithms exhibit some bias toward a particular class, highlighting their inconsistency in regards to recognizing erroneous reviews. Similar to other experiments using \textit{nlpaug}, BB and BBG here yield the best results. BBG and BB achieved the highest WF1 of 0.947 and 0.943 and MCC of 0.895 and 0.887. The ability of these models are supported by their ROC-AUC scores. The WF1 values of BBB, SB, and BBL are similarly satisfactory i.e. 0.94, 0.90, and 0.928 respectively. The result of BBB is almost identical to BB and BBG demonstrating the model's ability to identify fake reviews when the similarities between the augmented samples are low. SB and BBL demonstrate good performance despite having a short sequence length; their respective WF1 values are 4.4\% and 3.1\% higher than those of earlier experiments with \textit{nlpaug}.

\subsection{Ensemble Approaches:}
Here, we present some more experimental findings for both ensemble technique approaches. The results were previously presented in Tables \ref{tab:ensemblenlpaug} for the approach using the \textit{nlpaug} augmentation technique (Approach-2) and \ref{tab:ensemblesagar} for the \textit{bnaug} augmentation technique (Approach-3) only for the case of four augmentation samples along with the original fake reviews. We display the additional results for 1 to 3 augmented samples in addition to the original fake review. The experimental findings for the case of 1-3 augmented samples using approach-2 are shown in Table \ref{tab:appensemblenlpaug} while the results for approach-3 are shown in Table \ref{tab:appensemblesagar}.

\begin{table}[H]
\caption{Performance comparison between the two ensemble approaches (average and weighted) on 1 to 3 augmented fake reviews generated using \textit{nlpaug} (Approach - 2)}
\label{tab:appensemblenlpaug}
\begin{tabular}{ccc|ccc|ccc|ccc}
\hline
     & \multicolumn{2}{c|}{}   & \multicolumn{3}{c|}{\textbf{Fake}}                   & \multicolumn{3}{c|}{\textbf{Non-Fake}}                & \multicolumn{3}{c}{}                           \\ \hline
     & \multicolumn{1}{c}{\textbf{Methods}} & \multicolumn{1}{c|}{\begin{tabular}[c]{@{}c@{}}{\rotatebox[origin=c]{90}{\textbf{Aug}}}\end{tabular}} & \multicolumn{1}{c}{\textbf{P}}    & \multicolumn{1}{l}{\textbf{R}}    & \multicolumn{1}{l|}{\textbf{F1}}   & \multicolumn{1}{l}{\textbf{P}}    & \multicolumn{1}{l}{\textbf{R}}    & \multicolumn{1}{l|}{\textbf{F1}}   & \multicolumn{1}{l}{\textbf{WF1}}  & \multicolumn{1}{l}{\begin{tabular}[c]{@{}c@{}}\textbf{ROC}\\ \textbf{-AUC}\end{tabular}} & \textbf{MCC}  \\ \hline
\multirow{24}{*}{\rotatebox[origin=c]{90}{AVG-Ensemble Models}}      & EN1    & 1& 0.96 & 0.93 & 0.95 & 0.93 & 0.96 & 0.95 & 0.9477  & 0.95         & 0.90 \\  
     &         & 2& 0.97 & 0.95 & 0.96 & 0.95 & 0.97 & 0.96 & 0.9614  & 0.96         & 0.92 \\  
     &         & 3& 0.98 & 0.98 & 0.98 & 0.98 & 0.98 & 0.98 & 0.9785  & 0.98         & 0.96 \\  
     % &         & 4& 0.98 & 0.99 & 0.98 & 0.98 & 0.98 & 0.98 & 0.98  & 0.98         & 0.97 \\
     \cline{2-12} 
     & EN2     & 1& 0.96 & 0.91 & 0.94 & 0.92 & 0.97 & 0.94 & 0.9403  & 0.94         & 0.88 \\  
     &         & 2& 0.97 & 0.95 & 0.96 & 0.95 & 0.97 & 0.96 & 0.9614  & 0.96         & 0.92 \\  
     &         & 3& 0.97 & 0.98 & 0.97 & 0.98 & 0.97 & 0.97 & 0.9748  & 0.97         & 0.95 \\  
     % &         & 4& 0.98 & 0.98 & 0.98 & 0.98 & 0.98 & 0.98 & 0.98  & 0.98         & 0.96 \\ 
     \cline{2-12} 
     & EN3     & 1& 0.95 & 0.92 & 0.94 & 0.92 & 0.96 & 0.94 & 0.9365  & 0.94         & 0.87 \\  
     &         & 2& 0.96 & 0.94 & 0.95 & 0.94 & 0.97 & 0.95 & 0.9502  & 0.95         & 0.90 \\  
     &         & 3& 0.97 & 0.96 & 0.97 & 0.96 & 0.97 & 0.97 & 0.9664  & 0.97         & 0.93 \\  
     % &         & 4& 0.98 & 0.97 & 0.97 & 0.97 & 0.98 & 0.97 & 0.97  & 0.97         & 0.95 \\ 
     \cline{2-12} 
     & EN4     & 1& 0.92 & 0.94 & 0.93 & 0.94 & 0.92 & 0.93 & 0.9272  & 0.93         & 0.85 \\  
     &         & 2& 0.95 & 0.96 & 0.96 & 0.96 & 0.95 & 0.96 & 0.9565  & 0.96         & 0.91 \\  
     &         & 3& 0.95 & 0.98 & 0.97 & 0.98 & 0.95 & 0.96 & 0.9655  & 0.97         & 0.93 \\  
     % &         & 4& 0.97 & 0.99 & 0.98 & 0.99 & 0.97 & 0.98 & 0.98  & 0.98         & 0.95 \\ 
     \cline{2-12} 
     & EN5    & 1& 0.91 & 0.90 & 0.91 & 0.90 & 0.91 & 0.91 & 0.9086  & 0.91         & 0.82 \\  
     &         & 2& 0.96 & 0.93 & 0.95 & 0.93 & 0.96 & 0.95 & 0.9465  & 0.95         & 0.89 \\  
     &         & 3& 0.95 & 0.96 & 0.96 & 0.96 & 0.95 & 0.96 & 0.9571  & 0.96         & 0.91 \\  
     % &         & 4& 0.97 & 0.97 & 0.97 & 0.97 & 0.97 & 0.97 & 0.97  & 0.97         & 0.94 \\
     \cline{2-12} 
     & EN6 & 1& 0.94 & 0.92 & 0.93 & 0.92 & 0.94 & 0.93 & 0.9328  & 0.93         & 0.87 \\  
     &         & 2& 0.96 & 0.95 & 0.95 & 0.95 & 0.96 & 0.96 & 0.9552 & 0.96         & 0.91 \\  
     &         & 3& 0.96 & 0.99 & 0.97 & 0.98 & 0.96 & 0.97 & 0.9729  & 0.97         & 0.95 \\  
     % &         & 4& 0.97 & 0.98 & 0.98 & 0.98 & 0.97 & 0.98 & 0.98  & 0.98         & 0.95 \\ 
     \cline{1-12} 
     % & \multicolumn{11}{l}{}        \\ 
\multirow{24}{*}{\rotatebox[origin=c]{90}{Weighted-Ensemble Models}} & EN1    & 1& 0.96 & 0.93 & 0.94 & 0.93 & 0.96 & 0.95 & 0.9459  & 0.95         & 0.89 \\  
     &         & 2& 0.97 & 0.96 & 0.96 & 0.96 & 0.97 & 0.96 & 0.9614  & 0.96         & 0.92 \\  
     &         & 3& 0.98 & 0.98 & 0.98 & 0.98 & 0.98 & 0.98 & 0.9795  & 0.98         & 0.96 \\  
     % &         & 4& 0.97 & 0.99 & 0.98 & 0.99 & 0.97 & 0.98 & 0.98  & 0.98         & 0.96 \\ 
     \cline{2-12} 
     & EN2     & 1& 0.96 & 0.91 & 0.94 & 0.92 & 0.96 & 0.94 & 0.9384  & 0.94         & 0.88 \\  
     &         & 2& 0.97 & 0.95 & 0.96 & 0.95 & 0.97 & 0.96 & 0.9627  & 0.96         & 0.93 \\  
     &         & 3& 0.97 & 0.98 & 0.98 & 0.98 & 0.97 & 0.98 & 0.9757  & 0.98         & 0.95 \\  
     % &         & 4& 0.98 & 0.98 & 0.98 & 0.98 & 0.98 & 0.98 & 0.98  & 0.98         & 0.96 \\ 
     \cline{2-12} 
     & EN3     & 1& 0.95 & 0.92 & 0.94 & 0.92 & 0.96 & 0.94 & 0.9365  & 0.94         & 0.87 \\  
     &         & 2& 0.96 & 0.94 & 0.95 & 0.94 & 0.97 & 0.95 & 0.9527  & 0.95         & 0.91 \\  
     &         & 3& 0.97 & 0.98 & 0.98 & 0.98 & 0.97 & 0.98 & 0.9785  & 0.98         & 0.96 \\  
     % &         & 4& 0.98 & 0.99 & 0.98 & 0.98 & 0.98 & 0.98 & 0.98  & 0.98         & 0.96 \\ 
     \cline{2-12} 
     & EN4     & 1& 0.92 & 0.94 & 0.93 & 0.94 & 0.92 & 0.93 & 0.9272  & 0.93         & 0.85 \\  
     &         & 2& 0.95 & 0.96 & 0.96 & 0.96 & 0.95 & 0.96 & 0.9565  & 0.96         & 0.91 \\  
     &         & 3& 0.96 & 0.98 & 0.97 & 0.98 & 0.96 & 0.97 & 0.9673  & 0.97         & 0.93 \\  
     % &         & 4& 0.97 & 0.99 & 0.98 & 0.99 & 0.97 & 0.98 & 0.98  & 0.98         & 0.96 \\ 
     \cline{2-12} 
     & EN5    & 1& 0.91 & 0.90 & 0.91 & 0.90 & 0.91 & 0.91 & 0.9086  & 0.91         & 0.82 \\  
     &         & 2& 0.96 & 0.93 & 0.94 & 0.93 & 0.96 & 0.95 & 0.9453  & 0.95         & 0.89 \\  
     &         & 3& 0.95 & 0.98 & 0.97 & 0.98 & 0.95 & 0.97 & 0.9664  & 0.97         & 0.93 \\  
     % &         & 4& 0.97 & 0.98 & 0.98 & 0.98 & 0.97 & 0.98 & 0.98  & 0.98         & 0.96 \\ 
     \cline{2-12} 
     & EN6 & 1& 0.95 & 0.92 & 0.94 & 0.92 & 0.95 & 0.94 & 0.9366  & 0.94         & 0.87 \\  
     &         & 2& 0.96 & 0.95 & 0.95 & 0.95 & 0.96 & 0.96 & 0.9552  & 0.96         & 0.91 \\  
     &         & 3& 0.96 & 0.99 & 0.97 & 0.98 & 0.96 & 0.97 & 0.9729  & 0.97         & 0.95 \\  
     % &         & 4& 0.97 & 0.98 & 0.98 & 0.98 & 0.97 & 0.98 & 0.98  & 0.98         & 0.96 \\ 
     \cline{1-12}
\end{tabular}
\end{table}

\begin{table}[H]
\caption{Performance comparison between the two ensemble approaches (average and weighted) on 1 to 3 augmented fake reviews generated using \textit{bnaug} (Approach - 3)}
\label{tab:appensemblesagar}
\begin{tabular}{ccc|ccc|ccc|ccc}
\hline
     & \multicolumn{2}{c|}{}   & \multicolumn{3}{c|}{\textbf{Fake}}                   & \multicolumn{3}{c|}{\textbf{Non-Fake}}                & \multicolumn{3}{c}{}                           \\ \hline
     & \multicolumn{1}{c}{\textbf{Methods}} & \multicolumn{1}{c|}{\begin{tabular}[c]{@{}c@{}}{\rotatebox[origin=c]{90}{\textbf{Aug}}}\end{tabular}} & \multicolumn{1}{c}{\textbf{P}}    & \multicolumn{1}{l}{\textbf{R}}    & \multicolumn{1}{l|}{\textbf{F1}}   & \multicolumn{1}{l}{\textbf{P}}    & \multicolumn{1}{l}{\textbf{R}}    & \multicolumn{1}{l|}{\textbf{F1}}   & \multicolumn{1}{l}{\textbf{WF1}}  & \multicolumn{1}{l}{\begin{tabular}[c]{@{}c@{}}\textbf{ROC}\\ \textbf{-AUC}\end{tabular}} & \textbf{MCC}  \\ \hline
\multirow{24}{*}{\rotatebox[origin=c]{90}{AVG-Ensemble Models}}      & EN1     & 1& 0.92 & 0.90 & 0.91 & 0.90 & 0.93 & 0.91 & 0.9123  & 0.91         & 0.82 \\  
     &         & 2& 0.96 & 0.93 & 0.94 & 0.93 & 0.96 & 0.94 & 0.9415  & 0.94         & 0.88 \\  
     &         & 3& 0.97 & 0.93 & 0.95 & 0.93 & 0.97 & 0.95 & 0.9515  & 0.95         & 0.90 \\  
     % &         & 4& 0.97 & 0.94 & 0.95 & 0.94 & 0.97 & 0.95 & 0.95  & 0.95         & 0.91 \\ 
     \cline{2-12} 
     & EN2     & 1& 0.93 & 0.89 & 0.91 & 0.90 & 0.93 & 0.91 & 0.9123  & 0.91         & 0.83 \\  
     &         & 2& 0.97 & 0.91 & 0.94 & 0.92 & 0.97 & 0.94 & 0.9427  & 0.94         & 0.89 \\  
     &         & 3& 0.97 & 0.91 & 0.94 & 0.92 & 0.97 & 0.94 & 0.9430  & 0.94         & 0.89 \\  
     % &         & 4& 0.97 & 0.93 & 0.95 & 0.93 & 0.97 & 0.95 & 0.95  & 0.95         & 0.90 \\ 
     \cline{2-12}
     & EN3     & 1& 0.93 & 0.88 & 0.90 & 0.89 & 0.93 & 0.91 & 0.9048  & 0.90         & 0.81 \\  
     &         & 2& 0.97 & 0.91 & 0.94 & 0.92 & 0.97 & 0.94 & 0.9390  & 0.94         & 0.88 \\  
     &         & 3& 0.97 & 0.92 & 0.95 & 0.92 & 0.98 & 0.95 & 0.9477  & 0.95         & 0.90 \\  
     % &         & 4& 0.97 & 0.93 & 0.95 & 0.93 & 0.97 & 0.95 & 0.95  & 0.95         & 0.90 \\ 
     \cline{2-12} 
     & EN4     & 1& 0.91 & 0.89 & 0.90 & 0.89 & 0.91 & 0.90 & 0.9030  & 0.90         & 0.81 \\  
     &         & 2& 0.96 & 0.93 & 0.95 & 0.93 & 0.96 & 0.95 & 0.9465  & 0.95         & 0.89 \\  
     &         & 3& 0.97 & 0.93 & 0.95 & 0.93 & 0.97 & 0.95 & 0.9487  & 0.95         & 0.90 \\  
     % &         & 4& 0.96 & 0.95 & 0.96 & 0.95 & 0.97 & 0.96 & 0.96  & 0.96         & 0.91 \\ 
     \cline{2-12} 
     & EN5    & 1& 0.91 & 0.87 & 0.89 & 0.88 & 0.91 & 0.90 & 0.8936  & 0.89         & 0.79 \\  
     &         & 2& 0.97 & 0.92 & 0.94 & 0.92 & 0.97 & 0.94 & 0.9427  & 0.94         & 0.89 \\  
     &         & 3& 0.97 & 0.91 & 0.94 & 0.92 & 0.97 & 0.94 & 0.9412  & 0.94         & 0.88 \\  
     % &         & 4& 0.97 & 0.93 & 0.95 & 0.93 & 0.97 & 0.95 & 0.95  & 0.95         & 0.90 \\ 
     \cline{2-12} 
     & EN6 & 1& 0.91 & 0.89 & 0.90 & 0.89 & 0.91 & 0.90 & 0.9030  & 0.90         & 0.81 \\  
     &         & 2& 0.96 & 0.93 & 0.95 & 0.93 & 0.96 & 0.95 & 0.9465  & 0.95         & 0.89 \\  
     &         & 3& 0.97 & 0.93 & 0.95 & 0.93 & 0.97 & 0.95 & 0.9487  & 0.95         & 0.90 \\  
     % &         & 4& 0.96 & 0.95 & 0.96 & 0.95 & 0.97 & 0.96 & 0.96 & 0.96         & 0.91 \\ 
     \cline{1-12}
     % & \multicolumn{11}{l}{}        \\ 
\multirow{24}{*}{\rotatebox[origin=c]{90}{Weighted-Ensemble Models}} & EN1   & 1& 0.92 & 0.90 & 0.91 & 0.90 & 0.93 & 0.91 & 0.9123  & 0.91         & 0.82 \\  
     &         & 2& 0.96 & 0.93 & 0.94 & 0.93 & 0.96 & 0.94 & 0.9415  & 0.94         & 0.88 \\  
     &         & 3& 0.97 & 0.93 & 0.95 & 0.93 & 0.97 & 0.95 & 0.9515  & 0.95         & 0.90 \\  
     % &         & 4& 0.97 & 0.94 & 0.95 & 0.94 & 0.97 & 0.95 & 0.95  & 0.95         & 0.91 \\ 
     \cline{2-12} 
     & EN2     & 1& 0.93 & 0.89 & 0.91 & 0.89 & 0.94 & 0.91 & 0.9123  & 0.91         & 0.83 \\  
     &         & 2& 0.97 & 0.90 & 0.94 & 0.91 & 0.97 & 0.94 & 0.9377  & 0.94         & 0.88 \\  
     &         & 3& 0.98 & 0.92 & 0.95 & 0.93 & 0.98 & 0.95 & 0.9524  & 0.95         & 0.91 \\  
     % &         & 4& 0.97 & 0.93 & 0.95 & 0.93 & 0.97 & 0.95 & 0.95  & 0.95         & 0.90 \\ 
     \cline{2-12} 
     & EN3     & 1& 0.92 & 0.88 & 0.90 & 0.88 & 0.93 & 0.90 & 0.9011 & 0.90         & 0.80 \\  
     &         & 2& 0.97 & 0.91 & 0.94 & 0.92 & 0.97 & 0.94 & 0.9390  & 0.94         & 0.88 \\  
     &         & 3& 0.98 & 0.92 & 0.95 & 0.93 & 0.98 & 0.95 & 0.9496 & 0.95         & 0.90 \\  
     % &         & 4& 0.97 & 0.93 & 0.95 & 0.93 & 0.97 & 0.95 & 0.95  & 0.95         & 0.90 \\ 
     \cline{2-12} 
     & EN4     & 1& 0.92 & 0.90 & 0.91 & 0.90 & 0.92 & 0.91 & 0.9086 & 0.91         & 0.82 \\  
     &         & 2& 0.96 & 0.93 & 0.94 & 0.93 & 0.96 & 0.95 & 0.9453  & 0.95         & 0.89 \\  
     &         & 3& 0.97 & 0.93 & 0.95 & 0.93 & 0.97 & 0.95 & 0.9496  & 0.95         & 0.90 \\  
     % &         & 4& 0.96 & 0.94 & 0.95 & 0.94 & 0.97 & 0.95 & 0.95  & 0.95         & 0.91 \\ 
     \cline{2-12} 
     & EN5    & 1& 0.91 & 0.88 & 0.89 & 0.88 & 0.91 & 0.90 & 0.8955  & 0.90         & 0.79 \\  
     &         & 2& 0.97 & 0.91 & 0.94 & 0.92 & 0.97 & 0.94 & 0.9427 & 0.94         & 0.89 \\  
     &         & 3& 0.97 & 0.91 & 0.94 & 0.92 & 0.98 & 0.94 & 0.9430 & 0.94         & 0.89 \\  
     % &         & 4& 0.98 & 0.92 & 0.95 & 0.93 & 0.98 & 0.95 & 0.95  & 0.95         & 0.91 \\ 
     \cline{2-12} 
     & EN6 & 1& 0.93 & 0.88 & 0.90 & 0.89 & 0.93 & 0.91 & 0.9067  & 0.91         & 0.81 \\  
     &         & 2& 0.97 & 0.91 & 0.94 & 0.92 & 0.97 & 0.94 & 0.9415 & 0.94         & 0.88 \\  
     &         & 3& 0.98 & 0.92 & 0.95 & 0.92 & 0.98 & 0.95 & 0.9568 & 0.95         & 0.90 \\  
     % &         & 4& 0.97 & 0.93 & 0.95 & 0.94 & 0.97 & 0.96 & 0.95  & 0.95         & 0.91 \\ 
     \cline{1-12} 
\end{tabular}
\end{table}

\section{Performance Analysis on Some Unseen Test Reviews}\label{secA5}
Here, some more samples are provided which are taken from recent social media posts and tested on our selected models. The first instance from Table \ref{tab:samples2} is a non-fake review, and both models accurately predicted it. However, in the fourth sample, the reviewer is somewhat overstating the case, but approach-2 was unable to identify this as the review is so nearly a non-fake review as well. We must admit that the reviewer deserves credit for this.

Some examples of misclassified data utilizing LIME can be found in Table \ref{tab:LIME2}. In the first sample, there is some discrepancy between the models since the words {\bng kha{I}lam, ekhlam} and {\bng elaekshn} are colored differently in each model, which has an impact on the prediction. The models using approach-3 are unable to accurately predict the samples, however, the model of approach-2 does. In the second sample, words like {\bng baNGailyana, E{I}khaen, rkm}, and {\bng bhaela} are surprisingly deceiving the model in approach-3, dominating the prediction. In case of approach-3, the variances in data augmentation might be the cause. The result of approach-2 is always high as the augmentations are slightly biased as only 15\% words are replaced whereas other words remain same for a fake sample.
\begin{table}[H]
\centering
\caption{Generalization performance on some more unseen test reviews taken from recent social media posts by the proposed weighted ensemble model. (0 stands for \textit{fake} and 1 stands for \textit{non-fake})}
      \label{tab:samples2}
\begin{tabular}{cccc}
\hline
\textbf{Review}& \begin{tabular}[c]{@{}c@{}}\textbf{True}\\ \textbf{Label} \end{tabular} &  \begin{tabular}[c]{@{}c@{}} \textbf{Approach}\\ \textbf{-2} \end{tabular} & \begin{tabular}[c]{@{}c@{}} \textbf{Approach}\\ \textbf{-3} \end{tabular} \\ \hline
% \begin{tabular}[c]{@{}c@{}}\bng Taka ideJ E{I} ibirJain ek khaJ bha{I} Aar ETaek ibirJain bel?\\ \bng kar kaech ekmn laeg E{I} ebabar  ibirJain?\\ (Who eats this biryani with money and calls it biryani?\\Who likes this Boba's biryani?) \end{tabular}& 1             & 1               & 0              \\ \hline
% \begin{tabular}[c]{@{}c@{}}\bng Aapin ik eDsar/T pagl epis/TR ekk edkhel inejek thamaet\\ \bng paern na Aar thamaet Heb na . sada Hl khabar bhir/t eTibl Er\\ \bng bueph et paec/chn AanilimeTD epis/TR ekk ta{O} Aabar 4 dhrenr\\ \bng Aaech ibibhn/n ephLbhaerr mus eDanaT EbNNG Aaera Aenk Aenk\\ \bng Aa{I}eTm . E{I}sb eDsar/T Aa{I}eTemr pashapaish bueph et rJeech\\  \bng Aaera Aenk Aa{I}eTm eJmn phRa{I}D ra{I}s phRa{I}D icekn 3 dhrenr\\  \bng kabab baTar EbNNG garilk ebRD 3dhrenr sYup EbNNG Aaera Aenk\\  \bng Aa{I}eTm. 9 10s/TueDn/T Aphaer 750Takar buephet paec/chn 10\% \\ \bng chaD shudhu matR laNJ/c Er jnY\\ (Are you crazy about desserts? Can't stop yourself when\\ you see a pastry cake? You don't have to stop anymore.\\ You can get unlimited pastry cakes in the buffet of the\\ White hall with a full table, there are also 4 types of\\ mousse donuts of different flavors and many more items.\\ In addition to these dessert items, the buffet has many other\\ items such as fried rice, fried chicken, 3 types of kebabs,\\ butter and garlic bread, 3 types of soup and many more\\ items. 9/10 student offer 750 taka buffet get 10\%\\ discount. Just for lunch)  \end{tabular} & 0             & 1               & 0              \\ \hline 
\begin{tabular}[c]{@{}c@{}}\parbox[t]{4.3in}{{\bng ebaba ibiryanii Er sb ethek es/pshal Aa{I}eTm Hela grur cap epala{O} Ja Aamar sb ethek ebish pchn/d taedr E{I} cap epala{O} Ta Aenk ibkhYat ikhlgNNa{O} Er medhY Aaim pRay smy taedr E{I} khaen kha{I} Aaera Aenk Aa{I}eTm Aaech  Ja Aenk mja sitY ktha grur cap epala{O} 180 Taka Ek ips Aenk brh sa{I}j Er ibph thaek Aaim Ek saeth 2 Ta ineyichlam ta{I} Aenk men Hytaech maNNGs guela Aenk sphT ichela Aar epala{O} Er saeth maNNGs Er kimWenshn Ta Aenk juis lagtaichela eTs/T 9/10 elaekshn ikhlgNNa{O} itlpaparha 16 namWar eraD Er mathay}\\ (The most special item of Boba Biryani is beef chap polao which I like the most. Their chap polao is very famous in Khilgaon. There is a very big size of beef, I took 2 at a time, so it seems that the meat was very soft and the combination of the fragrant rice with the chicken was very juicy. Test 9/10. Location Khilgaon Tilpapara at the point of road number 16)} \end{tabular}   & 1             & 1               & 1              \\ \hline 
\begin{tabular}[c]{@{}c@{}}\parbox[t]{4.3in}{{\bng shHerr bhyabhH shb/d duuShN ethek ebirey pRan khuel EkTu  ephRsh inshWas iney ikchu muuHur/t kaTalam imrpuerr kYaephet. Aac/cha cel eJet Heb imrpuerr E{I} ruphTp epLes. Ek kthay esoun/dr/JY Aar  shHerr bYs/tta echerh palaenar mt. prhn/t ibekel AaD/Dabajii krar jnY{I} keykjn bn/dhu imel igeyichlam Eeta sun/dr ruphibhU edkhar jnY. tar per {I}phtair Ta ker Aaslam. taedr mYanu Ta edkhlam, pRa{I}s irjenbl bhaeb{I} kerech. rash km thaek, sair/bhs bhaela edy. {O}epn esMaikNNG epLs.}\\ (I spent a few moments in the cafe of Mirpur to get out of the terrible noise pollution of the city and take a breath of fresh air. Well, you have to go to this rooftop place in Mirpur. In a word, beauty and escape from the hustle and bustle of the city. I along with a few of my friends went together to see such a beautiful roof view for chatting in the afternoon. I came after doing my Iftar. I saw their menu, the price is reasonable. The rush is less, and the service is good. Open smoking place.)} \end{tabular}& 0             & 1               & 0   \\ \hline     
\end{tabular}
\end{table}

\begin{table}[H]
\caption{Feature importance explanation of generated by LIME based on the best performing individual model (\textit{BanglaBERT}). Here, the true level of the data is given on top of each review}
      \label{tab:LIME2}
\begin{tabular}{ccc} %#Solution : Take images of data with English annotations
\hline
\textbf{Reviews}  & \textbf{LIME Outputs} & \textbf{Predictions} \\ \hline
{\begin{tabular}[c]{@{}c@{}} \textbf{non-fake}\\\parbox[t]{3in}{{\bng ik kha{I}lam er bha{I} pRit kamerh shudhu maNNGs Aar icj edeshr sbecey Ha{I}pD ipj/ja Da{I}inNNG laUeNJ/jr imTbl iDp iDsh ipj/ja ekhlam222T kYashbYak Aphaer ekan ktha Hebna jas/T na kha{I}el{I} ims pura{I} elaekshn{h} 81 ernikn is/TRT {O}yarii sWp/nr ApijeT}\\ (What have I eaten! Only meat and cheese per bite. Most hyped pizza in the country. I have eaten Dining Lounge Meatball Deep Dish Pizza 222 taka. Cashback offer. Don't talk. Just don't miss. Location: 81 Rankin Street, Opposite to Wari Swapna)}\end{tabular}}     &        \begin{tabular}[c]{@{}c@{}}Approach-2\\\raisebox{-0.5\totalheight}{\includegraphics[width=0.35\textwidth, height=20mm]{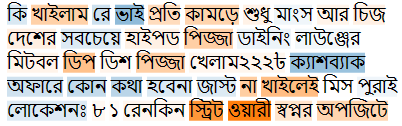}} \\Approach-3\\\raisebox{-0.5\totalheight}{\includegraphics[width=0.35\textwidth, height=20mm]{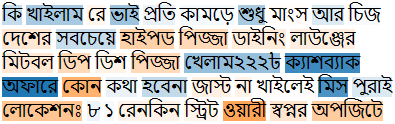}} \end{tabular}   &    {\begin{tabular}[l]{@{}c@{}} \textbf{non-fake}\\ \\ \\\\ \\\textbf{fake} \end{tabular}}       \\ \hline
{\begin{tabular}[c]{@{}c@{}} \textbf{fake}\\\parbox[t]{3in}{{\bng 150 Takay bueph Aphar ibeshWr sbecey km Takay esra bueph baNGailyana ebhaj baNGailyana ebhaejr AlemaSh/T sb phuD{I} esra. E{I}khaen clech matR 150 Takay bueph Aphar. 11 rkm bhr/ta AnilimeTD bhat. ebhaj Heb puera epT puer. sbguela bhr/ta{I} ejas. kakrhar eTs/T {O} khub bhaela. teb ebs/T ichl grur kala bhunaTa. Haesr maNNGsTa sphT ichl. {O}bharAl khuib bhaela. Aa{I}eTm nam dam 1. bhat {O} bhr/ta 150T 2. grur kala bhuna 150T 3. kakrhar edaepyaeja 150T 4. Has bhuna 200 eriTNNG 9/10  Abs/than pish/cm pan/thpth msijedr ibpriit paesh.} \\(Buffet offer at 150 taka, the best buffet at the lowest price in the world. Buffet offer is running here for only 150 taka. 11 types of bharta and unlimited rice. The feast will be full. All fillings are great. Test of crab is also very good. But the best was the black roast beef. The meat of the duck was soft. Overall very good. Item Name Price 1. Rice and Bharta 150 tak 2. Beef black roast 150 taka 3. Crab fry 150 taka 4. Duck Curry 200. Rating 9/10. Location Opposite to west Panthpath Mosque.)}\end{tabular}}     &        \begin{tabular}[c]{@{}c@{}}Approach-2\\\raisebox{-0.5\totalheight}{\includegraphics[width=0.35\textwidth, height=37mm]{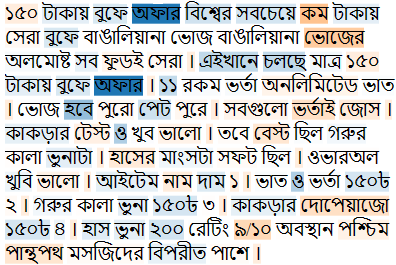}} \\Approach-3\\\raisebox{-0.5\totalheight}{\includegraphics[width=0.35\textwidth, height=37mm]{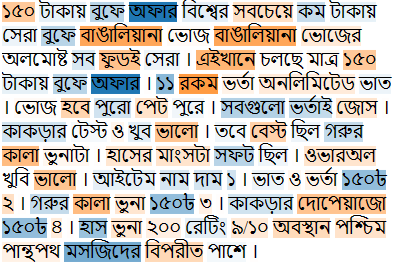}} \end{tabular}   &    {\begin{tabular}[l]{@{}c@{}} \textbf{fake}\\ \\ \\\\ \\ \\ \\ \\ \\\textbf{non-fake} \end{tabular}}       \\ \hline
\end{tabular}
\end{table}


% Author biography without author photo.
% Author biography. Author biography. Author biography.
% Author biography. Author biography. Author biography.
% Author biography. Author biography. Author biography.
% Author biography. Author biography. Author biography.
% Author biography. Author biography. Author biography.
% Author biography. Author biography. Author biography.
% Author biography. Author biography. Author biography.
% Author biography. Author biography. Author biography.
% Author biography. Author biography. Author biography.

% \bio{figs/pic1}
% Author biography with author photo.
% Author biography. Author biography. Author biography.
% Author biography. Author biography. Author biography.
% Author biography. Author biography. Author biography.
% Author biography. Author biography. Author biography.
% Author biography. Author biography. Author biography.
% Author biography. Author biography. Author biography.
% Author biography. Author biography. Author biography.
% Author biography. Author biography. Author biography.
% Author biography. Author biography. Author biography.

% \bio{figs/pic1}
% Author biography with author photo.
% Author biography. Author biography. Author biography.
% Author biography. Author biography. Author biography.
% Author biography. Author biography. Author biography.
% Author biography. Author biography. Author biography.

\end{document}